\PassOptionsToPackage{table}{xcolor}
\documentclass{article}
\usepackage{iclr2027_conference,times}
\usepackage{amsmath,amsfonts,bm}

\def\eqref#1{equation~\ref{#1}}
\def\1{\bm{1}}

\DeclareMathAlphabet{\mathsfit}{\encodingdefault}{\sfdefault}{m}{sl}
\SetMathAlphabet{\mathsfit}{bold}{\encodingdefault}{\sfdefault}{bx}{n}

\usepackage{amsmath,amssymb}
\usepackage{booktabs}
\usepackage{graphicx}
\usepackage{xcolor}
\usepackage{float}
\usepackage{hyperref}
\usepackage{url}
\usepackage{enumitem}
\usepackage{microtype}
\usepackage{pgfplots}
\usepgfplotslibrary{groupplots}
\usepackage{tabularx}

\definecolor{wardblue}{HTML}{2C6E9B}
\definecolor{wardgray}{HTML}{5C6770}
\hypersetup{colorlinks=true,citecolor=wardblue,linkcolor=wardblue,urlcolor=wardblue}

\usepackage{amssymb}
\usepackage{amsmath} 
\usepackage{multirow}

\usepackage[normalem]{ulem}
\useunder{\uline}{\ul}{}

\usepackage[skins,breakable]{tcolorbox}

\usepackage{makecell}
\pgfplotsset{compat=1.18}

\usepackage{tikz}

\pgfplotsset{compat=1.18}

\definecolor{modelsol}{HTML}{CFCFCF}
\definecolor{modelterra}{HTML}{F3E6A2}
\definecolor{modelclaude}{HTML}{F5D1D8}
\definecolor{modelgeminipro}{HTML}{D8C3D5}
\definecolor{modelgeminiflash}{HTML}{D9A36A}
\definecolor{modelqwen}{HTML}{D6EBDD}
\definecolor{modelkimi}{HTML}{A9D9A0}
\definecolor{modelglm}{HTML}{4F81BD}
\definecolor{wardblue}{HTML}{4C78A8}
\definecolor{wardgreen}{HTML}{59A14F}
\definecolor{wardorange}{HTML}{F28E2B}
\definecolor{wardred}{HTML}{E15759}
\definecolor{wardgray}{HTML}{9D9D9D}
\definecolor{wardlight}{HTML}{F5F7FA}

\usepackage{colortbl}

\usepackage{wrapfig}

\definecolor{wardblue}{HTML}{4E86B4}
\definecolor{wardlightblue}{HTML}{F3F7FA}
\definecolor{wardrowblue}{HTML}{EAF2F9}
\definecolor{wardbestblue}{HTML}{C8DDF0}

\usepackage{subcaption}

\pgfplotsset{compat=1.18}

\definecolor{wardblue}{HTML}{4C78A8}
\definecolor{wardgreen}{HTML}{59A14F}
\definecolor{wardorange}{HTML}{F28E2B}
\definecolor{wardgray}{HTML}{BAB0AC}
\definecolor{wardred}{HTML}{E15759}
\providecommand{\wardbest}[1]{%
  \cellcolor{wardbestblue}\textbf{#1}%
}

\usepackage{needspace}

\definecolor{wardrow}{RGB}{242,247,252}
\definecolor{wardbest}{RGB}{188,215,239}

\newcommand{\best}[1]{\cellcolor{wardbest}\textbf{#1}}

\newcommand{\greencheck}{{\color{green!70!black}$\boldsymbol{\checkmark}$}}
\newcommand{\redcross}{{\color{red!70!black}$\boldsymbol{\times}$}}

\definecolor{wardNavy}{HTML}{263B50}
\definecolor{wardBlue}{HTML}{4E86B4}
\definecolor{wardTeal}{HTML}{5A9E9A}
\definecolor{wardGold}{HTML}{D6A84B}
\definecolor{wardCoral}{HTML}{D7795B}
\definecolor{wardGray}{HTML}{AEB8C2}
\definecolor{wardGrid}{HTML}{DDE3E8}
\definecolor{wardRow}{HTML}{F3F7FA}
\definecolor{wardBest}{HTML}{E1F0F3}

\definecolor{actionyellow}{HTML}{F9DC83}
\definecolor{actionpink}{HTML}{EFA5D0}
\definecolor{actionblue}{HTML}{82ADD2}
\definecolor{actionpurple}{HTML}{8C88C8}

\definecolor{solblue}{HTML}{8EC5E5}
\definecolor{terrapeach}{HTML}{FFCB98}
\definecolor{flashrose}{HTML}{F5A3A3}
\definecolor{kimimint}{HTML}{93D6D1}
\providecommand{\wardbest}[1]{\cellcolor{wardBest}\textbf{#1}}

\pgfplotsset{
  wardaxis/.style={
    axis line style={draw=wardNavy!65,line width=0.45pt},
    tick style={draw=wardNavy!55,line width=0.4pt},
    tick label style={font=\scriptsize,text=wardNavy},
    label style={font=\scriptsize,text=wardNavy},
    title style={font=\small\bfseries,text=wardNavy,yshift=1pt},
    xmajorgrids=true,
    grid style={draw=wardGrid,line width=0.35pt},
    ytick style={draw=none},
    axis y line*=left,
    axis x line*=bottom,
    scaled ticks=false,
    clip=false
  }
}

\pgfplotsset{compat=1.18}

\definecolor{actionyellow}{HTML}{F9DC83}
\definecolor{actionpink}{HTML}{EFA5D0}
\definecolor{actionblue}{HTML}{82ADD2}
\definecolor{actionpurple}{HTML}{8C88C8}

\definecolor{solblue}{HTML}{8EC5E5}
\definecolor{terrapeach}{HTML}{FFCB98}
\definecolor{flashrose}{HTML}{F5A3A3}
\definecolor{kimimint}{HTML}{93D6D1}

\newcommand{\wardtakeaway}[1]{%
\begin{tcolorbox}[
    colback=blue!3,
    colframe=blue!25,
    boxrule=0.5pt,
    arc=2pt,
    left=5pt,
    right=5pt,
    top=3pt,
    bottom=3pt,
    before skip=4pt,
    after skip=4pt
]
#1
\end{tcolorbox}%
}
\usepackage{xcolor}
\usepackage{listings}
\usepackage{tcolorbox}
\tcbuselibrary{skins,breakable,listings}

\definecolor{wardCaseInk}{HTML}{293441}
\definecolor{wardCasePurple}{HTML}{74588F}
\definecolor{wardCaseRed}{HTML}{A55359}
\definecolor{wardCaseGreen}{HTML}{26796D}

\definecolor{wardCaseSystemBg}{HTML}{F5F6F8}
\definecolor{wardCaseSystemHead}{HTML}{E6E9EE}
\definecolor{wardCaseSystemLine}{HTML}{CFD5DE}

\definecolor{wardCaseUserBg}{HTML}{F0F6FA}
\definecolor{wardCaseUserHead}{HTML}{DCEBF4}
\definecolor{wardCaseUserLine}{HTML}{BFD3E2}

\definecolor{wardCaseModelBg}{HTML}{F5F1FA}
\definecolor{wardCaseModelHead}{HTML}{E9E0F3}
\definecolor{wardCaseModelLine}{HTML}{CDBDDE}

\definecolor{wardCaseToolBg}{HTML}{EFF7F7}
\definecolor{wardCaseToolHead}{HTML}{DDEDEE}
\definecolor{wardCaseToolLine}{HTML}{BDDADC}

\tcbset{
  wardcase system/.style={
    colback=wardCaseSystemBg,
    colbacktitle=wardCaseSystemHead,
    colframe=wardCaseSystemLine
  },
  wardcase user/.style={
    colback=wardCaseUserBg,
    colbacktitle=wardCaseUserHead,
    colframe=wardCaseUserLine
  },
  wardcase model/.style={
    colback=wardCaseModelBg,
    colbacktitle=wardCaseModelHead,
    colframe=wardCaseModelLine,
    coltext=wardCasePurple
  },
  wardcase tool/.style={
    colback=wardCaseToolBg,
    colbacktitle=wardCaseToolHead,
    colframe=wardCaseToolLine
  }
}

\newtcblisting{wardcasebox}[2]{
  enhanced,
  breakable,
  listing only,
  listing engine=listings,
  title={#2},
  fonttitle=\small\bfseries,
  coltitle=wardCaseInk,
  coltext=wardCaseInk,
  boxrule=0.45pt,
  arc=1pt,
  left=2mm,
  right=2mm,
  top=1.5mm,
  bottom=1.5mm,
  before skip=6pt,
  after skip=6pt,
  listing options={
    language={},
    basicstyle=\ttfamily\footnotesize,
    columns=fullflexible,
    keepspaces=true,
    breaklines=true,
    breakatwhitespace=false,
    breakindent=1em,
    showstringspaces=false,
    showspaces=false,
    showtabs=false,
    numbers=none,
    tabsize=2,
    aboveskip=0pt,
    belowskip=0pt
  },
  #1
}
\title{Seeing Is Not Acting: WARD Diagnoses the Evidence-to-State-to-Action Gap in Longitudinal Clinical Agents}

\author{\parbox{0.93\textwidth}{\centering
Chao Ding\textsuperscript{1,*,\textdagger}, Yidong Jiang\textsuperscript{2,*}, Junzhi Ning\textsuperscript{3,*},
Juncheng Wu\textsuperscript{4}, Yuhan Wang\textsuperscript{4}\\ Renjie Lu\textsuperscript{1}
Minjia Yuan\textsuperscript{5}, Gaopeng Wang\textsuperscript{3}, Liyang Wang\textsuperscript{6}\\
Pengcheng Chen\textsuperscript{7}, Dong Yin\textsuperscript{1}, Meiping Wang\textsuperscript{3},
Jinru Ding\textsuperscript{3}, Yuyin Zhou\textsuperscript{4,\textdaggerdbl}\\[4pt]
{\normalfont\scriptsize
\textsuperscript{1} Shanghai University of Traditional Chinese Medicine and Pharmacology
\textsuperscript{2} Carnegie Mellon University\\
\textsuperscript{3} Shanghai Artificial Intelligence Laboratory
\textsuperscript{4} University of California, Santa Cruz
\textsuperscript{5} Shanghai East Hospital
\textsuperscript{6} Tsinghua University\\
\textsuperscript{7} University of Washington}
}}

\iclrfinalcopy

\begin{document}
\maketitle
\begingroup
\renewcommand{\thefootnote}{\fnsymbol{footnote}}
\footnotetext[1]{Equal Contribution.}
\footnotetext[2]{Chao Ding is the project leader.}
\footnotetext[3]{Yuyin Zhou is the corresponding author.}
\endgroup

\begin{abstract}
Large language models for longitudinal clinical care must revise their assessments and actions as patient information accumulates. However, benchmarks centered on isolated responses or final task outcomes provide limited insight into intermediate failures in evidence use, state updating, and action selection. We introduce WARD, a benchmark for evaluating how clinical agents translate accumulating evidence into patient-state judgments and timely, safe actions. WARD comprises 4,560 clinical tasks across seven clinical output domains, organized into 240 checkpoints from 40 longitudinal clinical trajectories spanning 95 hospitalizations. Its Longitudinal Track evaluates these tasks under controlled cumulative evidence. Its Interactive Track evaluates end-to-end, tool-mediated clinical management across 271 verifiable evidence items and 141 reference transitions. Across 12 models with scorable checkpoint outputs, evidence recognition consistently outpaces downstream state and action decisions, with linked-evidence nonresponse reaching 74.47--90.90\%. Models also struggle to retain and update information over time, while in the Interactive Track only 11.46\% of executed actions advance a reference transition. These results reveal a persistent evidence-to-state-to-action gap and highlight the need for process-level evaluation of longitudinal clinical agents.
\par\noindent\textbf{Code and data:} \url{https://github.com/Dingsuper-creator/WARD}.
\end{abstract}

\section{Introduction}

Large language models (LLM) are increasingly deployed across the clinical pipeline, from patient-facing symptom triage to clinician-facing decision aids that predict deterioration and inform coverage or discharge decisions \citep{ross2023unitedhealth,wong2021epic}. In January 2026, OpenAI launched a clinical reasoning platform built on GPT-5 that within months reached major U.S. health systems, giving clinicians direct access to a model designed to synthesize guidelines and patient information for real-time decisions \citep{openai2026healthcare}. These systems are increasingly operating with reduced human oversight over multi-step, longitudinal patient encounters rather than isolated queries.

However, current LLMs still struggle with longitudinal clinical reasoning, failing to integrate new evidence into their assessments as a case unfolds. In a study of oncology records, GPT-4 produced reasoning errors in 23.1\% of case interpretations, with anchoring bias and omission errors directly linked to potentially harmful clinical recommendations \citep{kenaston2026reasoning}. These failures have already caused concrete harm in deployment. A class-action lawsuit against UnitedHealth alleges that its algorithmic length-of-stay tool, nH Predict, systematically cleared patients for discharge before they met clinical criteria, overriding physicians' ongoing assessments \citep{ross2023unitedhealth}. These shortcomings underscore the need to evaluate such capabilities before real-world deployment.

\begin{figure}
    \centering
    \includegraphics[width=\linewidth]{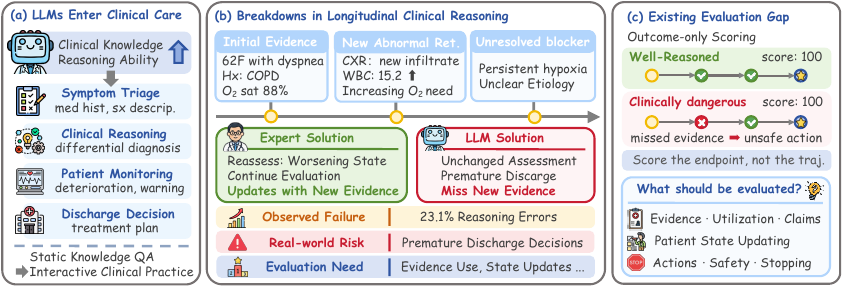}
\caption{Motivation for the WARD. LLMs increasingly operate in interactive clinical settings but fail to update with evolving evidence, while outcome-only evaluation can obscure unsafe reasoning.}
    \label{fig:intro}
\end{figure}

Nevertheless, existing benchmarks do not adequately evaluate model performance on long-horizon clinical reasoning as shown in Figure~\ref{fig:intro}. First, static benchmarks such as MedQA and MultiMedQA query a model once over a complete case and score a single terminal answer, leaving a large gap between benchmark performance and real clinical settings \citep{jin2021medqa,singhal2023clinical}. Second, recent interactive benchmarks move closer to real clinical workflows, introducing sequential dialogue, EHR tool use, and multi-turn database queries \citep{schmidgall2026agentclinic,jiang2025medagentbench,lee2026ehrchatqa}, but their scoring remains tied almost entirely to final outcomes, overlooking whether the model's intermediate decisions remain clinically appropriate as evidence accumulates. As a result, clinically dangerous trajectories can receive the same score as well-reasoned ones.

To address these gaps, we propose WARD, a benchmark designed to evaluate whether longitudinal clinical agents can translate accumulating evidence into correct patient state judgments and timely, safe actions. It comprises 4,560 clinical tasks across seven clinical output domains, organized into 240 checkpoints from 40 real-world longitudinal clinical trajectories. Based on the same set of clinical trajectories, WARD provides 2 evaluation tracks. 
The Longitudinal Track elicits one structured response independently at each checkpoint, providing only the cumulative record up to that time, and examines whether the model updates state and adjusts actions upon new evidence. The Interactive Track places the agent in a tool-mediated environment to complete full case trajectories with one tool call per turn, evaluating evidence acquisition, prerequisite resolution, sustained progress, and safety under a finite action budget.
Our contributions are as follows:

\begin{itemize}
   \item We construct WARD from real inpatient trajectories, with clinician-identified decision points and clinically grounded links between evolving evidence, decision targets, and longitudinal care progress. These trajectories support longitudinal evaluation across sequential clinical checkpoints and interactive evaluation through tool-mediated full-case execution.

   \item We design a process-level evaluation framework covering 7 clinical output domains, including evidence use and patient state, with 5 interactive diagnostics like prerequisite resolution and sustained progress to localize failures along the evidence-to-state-to-action chain.  
\item Experiments reveal 3 key limitations: models often fail to translate recognized evidence into clinical responses, struggle to update and retain correct judgments over time, and rarely turn locally valid actions into sustained trajectory progress.

\end{itemize}

\section{Related Work}
 
Early benchmarks such as MedQA \citep{jin2021medqa} and MultiMedQA \citep{singhal2023clinical} primarily assess medical knowledge and clinical reasoning through examination questions and static clinical cases. Subsequent benchmarks shifted toward EHR-based tasks that better reflect clinical practice. EHRBench \citep{xie2026ehrbench} constructs diagnosis, treatment, and prognosis questions from real-world records; EHRSHOT \citep{wornow2023ehrshot} evaluates clinical prediction from longitudinal patient histories; and TempoMed-Bench \citep{guan2026temporal} examines whether models can recover medical knowledge valid at a particular point in time.
More recent benchmarks extend clinical evaluation to interactive agents. AgentClinic \citep{schmidgall2026agentclinic} places models in simulated clinical encounters where they interact with patients and tools to reach a diagnosis; MedAgentBench \citep{jiang2025medagentbench} provides a FHIR-compatible virtual EHR environment for completing realistic workflows through tool use; and EHRChatQA \citep{lee2026ehrchatqa} evaluates adaptive querying and action selection through multi-turn clarification and database interaction. WARD evaluates evidence-aligned state and action decisions across clinical checkpoints, while its interactive track further tests prerequisite resolution and sustained trajectory progress. See Appendix~\ref{app:i-comparison} for a detailed comparison.

\section{WARD Benchmark Construction}
\subsection{Clinical Trajectory and Checkpoint Construction}
% raw records
% clinician reconstruction
% case selection
% 40 cases
% checkpoint placement
% 240 checkpoints
% cumulative evidence release

WARD was developed from a deidentified institutional EHR snapshot containing 56,833 inpatient encounters across 138 clinical departments. Clinicians screened this snapshot for decision-dense cases with interpretable changes in patient state and clinically consequential dependencies among actions. The selected records were reconstructed into 40 longitudinal clinical trajectories spanning 95 hospitalizations, preserving the documented events and their temporal order. Checkpoints were anchored to clinically meaningful decision transitions rather than uniform time intervals. WARD distinguishes clinical time, when an event occurred, from available time, when its result became accessible for clinical decision making. At checkpoint \(t\), the model receives only the cumulative record available by that point; all subsequently available evidence remains hidden. Detailed selection criteria and the case taxonomy are provided in Appendix~\ref{app:case_taxonomy}.

GPT-4o \citep{openai2024gpt4o} generated provisional, non-authoritative annotations for all 240 checkpoints. Each checkpoint was processed independently within its temporal visibility boundary, without access to future evidence, downstream outcomes, existing reference labels, or clinician answers. This process yielded 1,019 clinical claims, 4,880 normalized patient-state elements, and 823 candidate actions for subsequent clinical review.A panel of 20 physicians reviewed every checkpoint against the temporally eligible clinical record through multiple rounds of quality control. Within each case, at least 40\% of checkpoints received an independent second review. Critical omissions, potentially harmful actions, major safety constraints, and stopping decisions involving unresolved blockers underwent additional review regardless of sampling status. Disagreements were adjudicated by a third physician without input from the annotation model. Only annotations approved or adjudicated by clinicians entered the frozen reference set. The pre-annotation prompts, validation rules, review protocol, and adjudication procedures are documented in Appendix~\ref{app:annotation-protocol}.

% Evidence -> Utilization -> State -> Action
% Claims / Safety / Stopping
% Gold annotation
% state delta / critical actions / blockers
\subsection{Task Formulation}

\begin{figure}
    \centering
    \includegraphics[width=\linewidth]{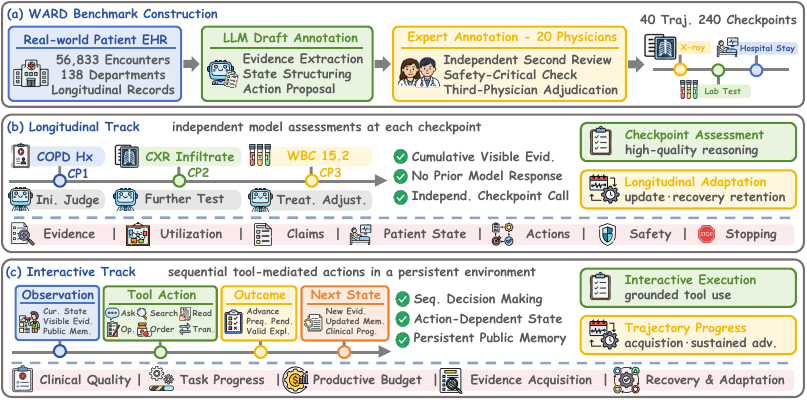}
    \caption{Overview of WARD benchmark construction and evaluation. WARD builds clinician-validated longitudinal trajectories from real-world EHRs and evaluates clinical agents through Longitudinal and Interactive Tracks, assessing checkpoint-level clinical judgment, temporal adaptation, and sequential tool-mediated execution.}
    \label{fig:frame}
\end{figure}

As shown in Figure \ref{fig:frame}, at each checkpoint, WARD evaluates model performance through 7 structured clinical output domains:
\textbf{Evidence} identifies the public record items most directly relevant to the current decision.
\textbf{Utilization} records explicit links from visible evidence to clinical claims, state elements, and actions.
\textbf{Claims} represent clinically meaningful assertions that are supported, contradicted, or remain unresolved under the cumulative record.
\textbf{Patient State} integrates these assertions into the current clinical status, including active and resolved problems, unresolved prerequisites, and treatment readiness.
\textbf{Actions} specify clinically appropriate next steps based on the resulting patient state.
\textbf{Safety} and \textbf{Stopping} assess clinically important constraints, unresolved blockers, and whether investigation, treatment, observation, or termination remains appropriate.

\subsection{Data Statistics}
WARD contains 4,560 clinical tasks, organized into 240 clinically defined checkpoints from 40 longitudinal trajectories covering 95 hospitalizations. Each trajectory contains 4--10 checkpoints. The cohort has a median age of 64 years (IQR, 56.8--70.0; range, 30--76) and is 60\% female and 40\% male. The Interactive Track evaluates end-to-end, tool-mediated clinical management over 271 verifiable evidence items and 141 reference transitions. Interactive trajectories contain a median of 27 evidence items and 12 transitions, with maximum of 41 and 27, respectively. Detailed cohort and case-level statistics are provided in Appendix~\ref{app:case-catalog}.

\section{WARD Evaluation Framework}
\subsection{Longitudinal Evaluation}
% checkpoint-level independent evaluation
% visible context
% model outputs
% what it evaluates
\label{sec:longitudinal-evaluation}
The Longitudinal Track evaluates both the model's structured clinical judgment at individual checkpoints and how these judgments change over time within the same case. At each predefined checkpoint, the model receives the cumulative public record available up to that point, without access to future information, reference answers, or its outputs from earlier checkpoints. The model must identify the evidence relevant to the current decision, explain how that evidence supports its clinical reasoning, represent the current patient state, propose appropriate next actions, and determine whether investigation, treatment, waiting, or stopping is appropriate. WARD then compares checkpoint predictions in temporal order within each case to assess whether the model updates its judgments when new evidence appears, later recovers previously missed information, avoids regression on previously correct information, and adjusts stopping decisions as prerequisites change. Task definition and data visibility rules are detailed in Appendix~\ref{app:visibility-audit} and~\ref{app:b-longitudinal-track} respectively.

\subsection{Interactive Evaluation}
% trajectory-level tool-mediated environment
% tools / feedback / memory
% finite budget
% what it evaluates
\label{sec:interactive-evaluation}

The Interactive Track evaluates whether the model can translate the evidence, state, and action reasoning assessed in the Longitudinal Track into sequential clinical actions that advance the patient trajectory. The model operates in a continuously updated clinical interaction environment, receiving at each turn the current stage, visible and newly released evidence, the previous action outcome, and public memory, and then issuing one clinical tool call. Available actions include patient communication, record search and retrieval, diagnostic or procedural actions, treatment, and care transition. The environment resolves each action against hidden clinical transitions and prerequisites and returns the resulting outcome. Actions that satisfy the current prerequisites advance the patient state and release the corresponding evidence, while prior actions, feedback, and memory persist across turns. The model must use this information to choose subsequent actions and progress through the clinical trajectory within a finite action budget. Tool and environment details are provided in Appendix~\ref{app:interactive-environment}.

\section{Experimental Setup}
\label{sec:experimental-setup}

\subsection{Models and Inference}
\label{sec:models-inference}

We attempted 15 LLMs in the Longitudinal Track. The general-purpose set includes GPT-5.6-SOL and GPT-5.6-Terra \citep{openai2026gpt56}, GPT-5.1 \citep{openai2025gpt51}, Claude Sonnet 5 \citep{anthropic2026sonnet5}, Gemini 3.1 Pro Preview and Gemini 3.6 Flash \citep{google2026gemini31pro,google2026gemini36flash}, Qwen3.7-Max \citep{qwen2026qwen37}, Qwen3-32B and Qwen3-14B \citep{yang2025qwen3}, Kimi-K3 \citep{kimiteam2026k3}, and GLM-5.2 \citep{zai2026glm52}. The medical-domain set includes Baichuan-M3 \citep{baichuanm3team2026}, MedGemma-27B and MedGemma-4B\citep{sellergren2025medgemma}, and Med-PaLM~2 \citep{singhal2025medpalm2}. The Interactive Track evaluates GPT-5.6-SOL, GPT-5.6-Terra, Gemini-3.6-Flash and Kimi-K3 on the prespecified subset. All models use the same setting with temperature = 0. The generation length is 32,768 tokens for checkpoint evaluation and 640 tokens per interactive call. Interactive models issue one tool call per turn with a fixed trajectory-level action budget. Interactive trajectories use an action budget of 40. Full inference configurations are in Appendix~\ref{app:inference-configurations}.

\subsection{Evaluation Protocol}
\label{sec:evaluation}

\subsubsection{Checkpoint Metrics}
\label{sec:checkpoint-metrics}

At each checkpoint, structured fields are matched deterministically, while clinical text fields are first evaluated by normalized exact and registered-alias matching, with unresolved prediction--gold pairs passed to a frozen semantic Judge. Eligible pairs are then resolved through one-to-one matching to determine TP, FP, and FN. We report 7 F1 metrics: Evidence, Utilization, Claims, Patient State, Actions, Safety, and Stopping blockers, together with Stopping-decision accuracy and Critical-action recall. Full matching rules and metric definitions are provided in Appendix~\ref{app:metric-definitions}.

For model \(m\), trajectory \(c\), and dimension \(d\), TP, FP, and FN are pooled across all checkpoints \(T(c)\) before computing trajectory-level F1:
\begin{equation}
F1_{mcd}
=
\frac{2\sum_{t\in T(c)}\mathrm{TP}_{mctd}}
{2\sum_{t\in T(c)}\mathrm{TP}_{mctd}
+\sum_{t\in T(c)}\mathrm{FP}_{mctd}
+\sum_{t\in T(c)}\mathrm{FN}_{mctd}}.
\end{equation}

Model-level scores are then macro-averaged across trajectories: all 40 trajectories in the main analysis and the 10 prespecified Interactive trajectories in Table~\ref{tab:cross-track-performance}. Invalid checkpoint outputs are scored as empty predictions in the primary checkpoint analysis but treated as unscorable in transition-based longitudinal analyses; valid-output-only sensitivity analyses are also reported.

\subsubsection{Longitudinal and Interactive Metrics}
\label{sec:process-metrics}

Longitudinal track compares independently elicited predictions across ordered checkpoints, focusing on response to new evidence, later recovery of missed information, and regression from previously correct states. We report 6 diagnostics: State-delta immediate response, linked-evidence nonresponse, Stopping-change response, Evidence subsequent match, Evidence regression, and Stopping-decision regression. Cases with 0 denominators are excluded from the corresponding metric. Because checkpoint predictions are elicited independently, missed-to-matched transitions are described as subsequent matches or recovery rather than autonomous self-correction.

Interactive track assesses clinical quality, task progress, efficiency, repetition, and recovery. Metrics include clinician-rated overall clinical quality, milestone coverage, productive budget yield, reference-evidence recall, no-new-evidence repeat rate, longest nonproductive-run fraction, post-evidence advance rate, and pending recovery.

\subsection{Evaluation Reliability}
\label{sec:evaluation-reliability}

To reduce evaluator-specific bias and assess sensitivity to judge choice, we use DeepSeek-V4-Pro \citep{deepseekai2026deepseekv4} as the primary semantic evaluator and Qwen3.7-Max \citep{qwen2026qwen37} as an independent sensitivity evaluator, forming a dual-judge setup consistent with established LLM-as-a-Judge practice \citep{zheng2023judging}. Both judges apply the same prompt, response schema, and five-class verdict set and were calibrated on 800 manually labeled semantic pairs. DeepSeek (Judge-B), the primary evaluator, achieved 86.52\% accuracy and Cohen's $\kappa=0.80$ (95\% CI: 0.73--0.87) against the clinician-annotated reference, while Qwen (Judge-A), used for sensitivity analysis, achieved 82.98\% accuracy. These statistics quantify agreement with the clinician-labeled calibration reference.
For a matched baseline, three physicians independently completed the same
50-checkpoint subset under the models' visibility boundary, instructions,
and output schema. Their submissions were scored against frozen references
using the identical pipeline. Mean clinician F1 exceeded the model
mean by 9.70, 13.44, and 19.22 percentage points for Evidence Utilization,
Patient State, and Next Actions, respectively. Scores and adjusted paired
case-resampling intervals are reported in Appendix~\ref{app:human-baseline}.

For the Interactive Track, the primary 1--7 overall clinical quality score achieved a quadratic-weighted Cohen's $\kappa=0.82$ (95\% CI: 0.80--0.87) and an absolute-agreement ICC of 0.88. Confidence intervals are estimated by case-level bootstrap resampling that preserves model pairing, using 20,000 replicates for checkpoint and longitudinal analyses and 10,000 for interactive analyses. Detailed clinician rubrics, independent ratings, and adjudication procedures are provided in Appendix~\ref{app:evaluator-validation}.

\section{Results}

WARD reveals a widening gap between recognizing clinical evidence and acting on it effectively. Relevant evidence does not consistently translate into correct patient-state judgments or next actions; these judgments are not always maintained over time; and locally reasonable actions do not necessarily yield sustained trajectory progress. We therefore analyze this gap through 3 views: the evidence-to-decision gap, longitudinal adaptation and retention, and interactive execution and sustained progress.

\subsection{Evidence Recognition and Downstream Decision Gaps}
\wardtakeaway{
\textit{Evidence recognition substantially outpaces downstream state and action performance.}
}

\begin{table*}
\centering
\scriptsize
\caption{
\textbf{Primary checkpoint-track performance across models.}
All values are case-macro percentages. Evidence, Utilization, Patient State, and Actions are reported as F1; Stopping Accuracy is categorical. Darker cells indicate the best per-column point estimate. \redcross marks models with no scorer-valid outputs; †     indicates MedGemma-27B's 52 schema-invalid checkpoints, scored as empty. 
}
\label{tab:main-checkpoint}

\setlength{\tabcolsep}{2pt}

\begin{tabular}{l cc c ccc}
\toprule

& \multicolumn{2}{c}{\textbf{Evidence Processing}}
& \multicolumn{1}{c}{\textbf{State}}
& \multicolumn{3}{c}{\textbf{Clinical Decisions}} \\

\cmidrule(lr){2-3}
\cmidrule(lr){4-4}
\cmidrule(lr){5-7}

\textbf{Model}
& \makecell[c]{\textbf{Evidence}\\\textbf{Recognition}}
& \makecell[c]{\textbf{Evidence}\\\textbf{Utilization}}
& \makecell[c]{\textbf{Patient}\\\textbf{State}}
& \makecell[c]{\textbf{Next}\\\textbf{Actions}}
& \makecell[c]{\textbf{Critical-Action}\\\textbf{Recall}}
& \textbf{Stopping Acc.} \\

\midrule

\rowcolor{wardrow}
GPT-5.6 SOL
& 55.85
& \best{18.87}
& \best{27.93}
& 20.95
& \best{29.15}
& 63.66 \\

GPT-5.6 Terra
& 45.34
& 14.42
& 23.55
& \best{24.96}
& 26.19
& 57.48 \\

\rowcolor{wardrow}
GPT-5.1
& 63.19
& 15.12
& 19.18
& 13.80
& 17.58
& 59.64 \\

Claude Sonnet 5
& 47.32
& 11.54
& 20.48
& 13.07
& 14.19
& 53.96 \\

\rowcolor{wardrow}
Gemini 3.1 Pro
& 42.33
& 11.36
& 14.90
& 11.02
& 10.71
& 67.37 \\

Gemini 3.6 Flash
& 37.27
& 10.40
& 16.25
& 13.45
& 12.06
& 70.10 \\

\rowcolor{wardrow}
Qwen 3.7 Max
& 49.51
& 10.93
& 14.04
& 11.38
& 11.36
& 71.76 \\

Qwen3-32B
& 31.50
& 5.84
& 10.72
& 4.92
& 4.41
& 36.49 \\

\rowcolor{wardrow}
Qwen3-14B
& 17.94
& 2.01
& 4.19
& 1.18
& 0.60
& 17.79 \\

Kimi K3
& \best{63.40}
& 17.36
& 23.27
& 11.77
& 19.63
& \best{72.09} \\

\rowcolor{wardrow}
GLM-5.2
& 59.39
& 15.07
& 19.97
& 11.84
& 17.08
& 59.49 \\

Baichuan-M3
& \redcross
& \redcross
& \redcross
& \redcross
& \redcross
& \redcross \\

\rowcolor{wardrow}
MedGemma-27B$^\dagger$
& 51.70
& 7.30
& 10.52
& 3.00
& 2.60
& 47.09 \\

MedGemma-4B
& \redcross
& \redcross
& \redcross
& \redcross
& \redcross
& \redcross \\

\rowcolor{wardrow}
Med-PaLM 2
& \redcross
& \redcross
& \redcross
& \redcross
& \redcross
& \redcross \\

\bottomrule
\end{tabular}
\end{table*}

Models perform substantially better at evidence recognition than at patient-state inference and action selection, and this gap is consistent across models. As shown in Table~\ref{tab:main-checkpoint}, across the 12 scorable models, the median Evidence Recognition F1 is 48.42\%, compared with 17.72\% for Patient State and 11.81\% for Next Actions, corresponding to median score gaps of 27.68 and 34.58 percentage points. Evidence-level model rankings also become less aligned with downstream state and action performance: Evidence Recognition correlates strongly with Evidence Utilization (Spearman $\rho=0.80$), but more weakly with Patient State ($\rho=0.53$) and Next Actions ($\rho=0.37$).

\begin{figure}[]
    \centering
    \includegraphics[width=\linewidth]{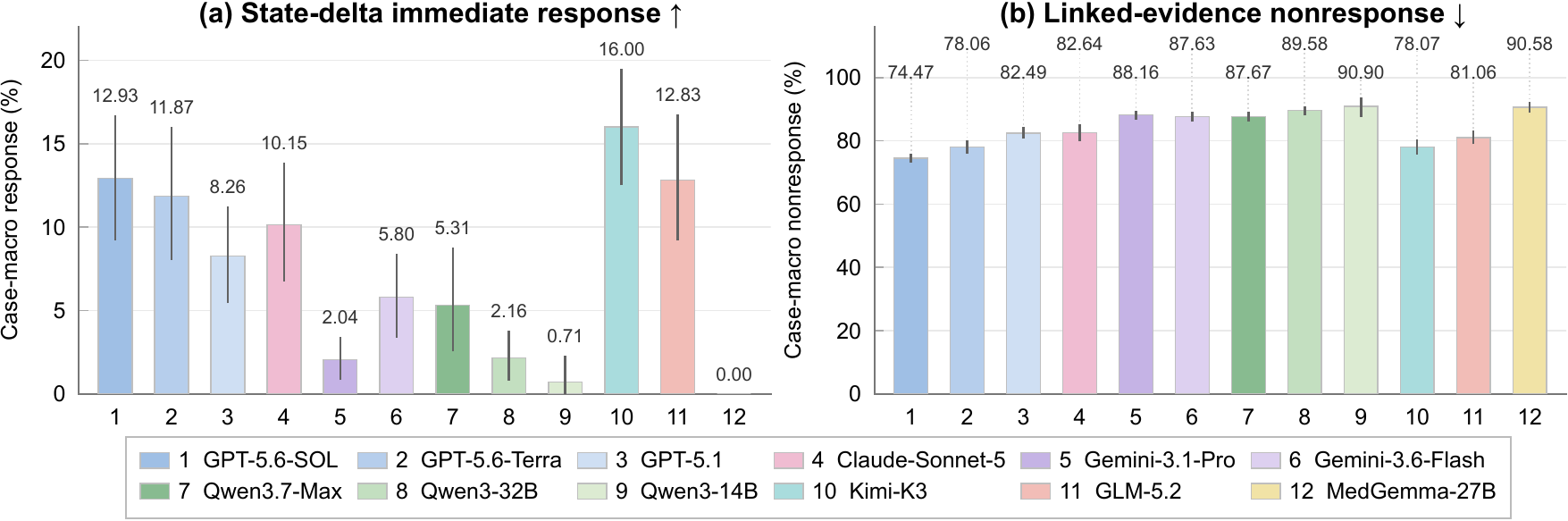}
    \caption{\textbf{Conditional downstream response.}
Bars show case-macro percentages and whiskers show 95\% case-bootstrap CIs.
(a) State-delta immediate response; (b) linked-evidence nonresponse.}
    \label{fig:conditional-response}
\end{figure}

This disconnect persists even when the relevant evidence has already been recognized. Linked-evidence nonresponse remains high at 74.47\%--90.90\%, while state-delta immediate response is only 0--16.00\% (Figure~\ref{fig:conditional-response}). This suggests that models can identify and associate local evidence but struggle to incorporate it into an evolving clinical state and update the dependent state variables and decision conditions. As decisions increasingly depend on multiple interrelated evidence items and prerequisites, these incomplete local updates can compound, preventing evidence-level capability from consistently carrying through to state and action decisions.

\subsection{Longitudinal Adaptation and Retention}

\wardtakeaway{
\textit{Models struggle to both recover and retain correct clinical judgments over time.}
}
As shown in Table \ref{tab:longitudinal-summary}, models struggle to both recover previously missed information and retain evidence that was already identified correctly in longitudinal tasks. The median Later Match is only 22.23\%, while the median Evidence Regression reaches 37.97\%. Kimi-K3 and GLM-5.2 combine relatively high recovery with low regression, whereas MedGemma-27B achieves the highest Later Match (50.59\%) but still exhibits 43.80\% Evidence Regression. Models appear more capable of locally correcting omissions at the current checkpoint than of continuously preserving and updating the previously established clinical state, allowing previously correct information to be lost again as new evidence arrives. Therefore, later correction does not ensure stable long-term state maintenance.

\begin{table}[]
\centering
\scriptsize
\caption{\textbf{Longitudinal adaptation and stability.} All values are case-macro percentages. Recovery/adaptation metrics are higher-is-better; regression metrics are lower-is-better. \textsuperscript{*}No stopping regression occurred among 25 eligible Qwen3-32B clinical trajectories. $^\dagger$ MedGemma-27B has 52 unscorable checkpoints. Eligible cases are 29/40, 35/40, 40/40, and 29/40, respectively.}
\label{tab:longitudinal-summary}
\begin{tabular}{l c c c c}
\toprule

&
\multicolumn{2}{c}{\textbf{Evidence Dynamics}}
&
\multicolumn{2}{c}{\textbf{Stopping Dynamics}}
\\

\cmidrule(lr){2-3}
\cmidrule(lr){4-5}

\textbf{Model}
&
\textbf{Later Match} $\uparrow$
&
\textbf{Evidence Regression} $\downarrow$
&
\textbf{Change Response} $\uparrow$
&
\textbf{Stopping Regression} $\downarrow$
\\
\midrule

\rowcolor{wardrowblue}
GPT-5.6 SOL
& 25.07
& 33.30
& 62.71
& 13.10
\\

GPT-5.6 Terra
& 5.61
& 40.97
& 56.88
& 18.10
\\

\rowcolor{wardrowblue}
GPT-5.1
& 26.69
& \wardbest{22.12}
& 57.50
& 19.38
\\

Claude Sonnet 5
& 32.89
& 38.08
& 51.25
& 19.72
\\

\rowcolor{wardrowblue}
Gemini 3.1 Pro
& 18.64
& 52.09
& 69.58
& 39.18
\\

Gemini 3.6 Flash
& 13.98
& 64.91
& 70.63
& 37.22
\\

\rowcolor{wardrowblue}
Qwen 3.7 Max
& 11.68
& 36.24
& \wardbest{74.58}
& 32.11
\\

Qwen3-32B
& 14.86
& 50.67
& 20.51
& \wardbest{0.00\textsuperscript{*}}
\\

\rowcolor{wardrowblue}
Qwen3-14B
& 19.39
& 37.85
& 48.44
& 35.00
\\

Kimi K3
& 42.41
& 30.56
& 74.38
& 21.93
\\

\rowcolor{wardrowblue}
GLM-5.2
& 41.56
& 26.37
& 58.33
& 22.14\\MedGemma-27B$^\dagger$& \wardbest{50.59}& 43.80& 46.88& 19.20
\\

\bottomrule
\end{tabular}
\end{table}

This instability also extends to stopping decisions, where effective performance requires adapting to new information without disrupting previously correct judgments. Change Response and Stopping Regression are positively correlated with Spearman $\rho=0.53$. Low regression alone therefore does not imply stable stopping behavior. Qwen3-32B achieves 0\% regression but has a Change Response of only 20.51\%, indicating limited adjustment to new information. In contrast, Qwen3.7-Max and Kimi-K3 achieve Change Responses of 74.58\% and 74.38\%, respectively, but still exhibit regression rates of 32.11\% and 21.93\%, showing that frequent updates can disrupt prior judgments when consistency across conditions is not maintained. GPT-5.6-SOL provides a more balanced case, combining a 62.71\% Change Response with only 13.10\% Stopping Regression, suggesting that effective stopping behavior depends on consistent rather than simply frequent updates.

\subsection{Interactive Execution and Sustained Progress}

\wardtakeaway{
\textit{Locally appropriate decisions do not consistently yield sustained interactive progress.}
}

\begin{figure}
    \centering
    \includegraphics[width=\linewidth]{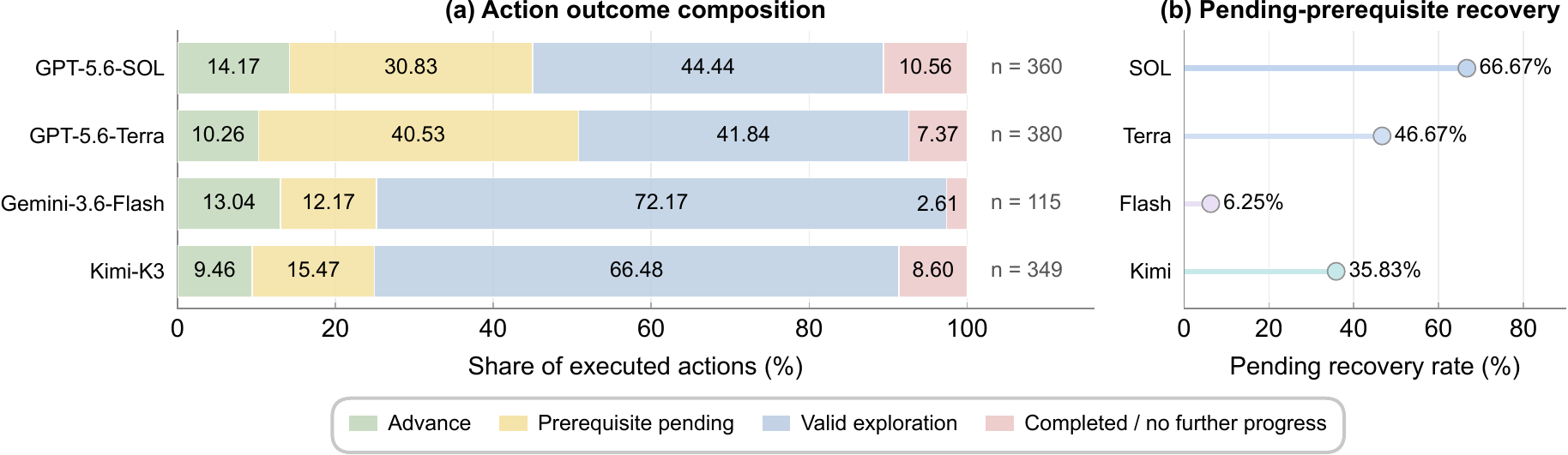}
    \caption{
\textbf{Interactive action outcomes and recovery dynamics.}
(a) Distribution of executed action outcomes, excluding non-executed calls and unused budget; \(n\) denotes executed actions.
(b) Fraction of prerequisite-pending transitions later completed within the same trajectory.
}
    \label{fig:interactive-runner-analysis}
\end{figure}

The primary bottleneck in interactive execution is converting consecutive actions into sustained reference-path progress (Figure \ref{fig:interactive-runner-analysis}). After aggregating the four models weighted by executed actions, only 11.46\% of actions advance a reference transition, 52.66\% correspond to valid exploration, 27.66\% fall into prerequisite pending, and the remaining 8.23\% are completed / no further progress. At the trajectory level, 76.92\% of nonterminal trajectories exhibit repeated or prolonged stagnation, and 56.41\% ultimately end with unresolved prerequisites. Many actions therefore fail to accumulate into sustained task progress and instead remain in exploration or unresolved dependency conditions.

\begin{table}
\centering
\scriptsize
\setlength{\tabcolsep}{2.0pt}

\caption{
\textbf{Checkpoint and interactive performance on the same prespecified clinical trajectories.}
Next Actions F1 and all Interactive Track metrics use the same trajectories. Clinical Quality uses a 1--7 scale; all others are in \%. Productive Budget Yield includes unused budget in its denominator.}
\label{tab:cross-track-performance}

\begin{tabular}{@{}lccccc@{}}
\toprule
&
\multicolumn{1}{c}{\textbf{Checkpoint}}
&
\multicolumn{4}{c}{\textbf{Interactive}}
\\
\cmidrule(lr){2-2}
\cmidrule(lr){3-6}

\textbf{Model}
&
\textbf{Next Actions F1}
&
\textbf{Clinical Quality}
&
\textbf{Milestone Coverage}
&
\textbf{Productive Budget Yield}
&
\textbf{Evidence Recall}
\\
\midrule

\rowcolor{wardrowblue}
GPT-5.6-SOL
& 22.41
& \wardbest{4.40}
& \wardbest{38.13}
& \wardbest{13.75}
& \wardbest{41.98}
\\

GPT-5.6-Terra
& \wardbest{26.18}
& 2.80
& 27.91
& 10.75
& 31.32
\\

\rowcolor{wardrowblue}
Gemini-3.6-Flash
& 17.33
& 2.30
& 11.36
& 4.50
& 12.57
\\

Kimi-K3
& 16.69
& 3.10
& 26.62
& 9.25
& 28.33
\\

\bottomrule
\end{tabular}
\end{table}

A major source of this stagnation is unresolved prerequisite blocking, from which models recover inconsistently. As shown in Figure \ref{fig:interactive-runner-analysis}, the median pending recovery rate across the four models is only 41.25\%, ranging from 6.25\% to 66.67\%, indicating that most pending transitions are not reliably resolved in subsequent steps and that recovery ability varies substantially across models. Although GPT-5.6-SOL achieves the highest recovery rate, in one trajectory it resolves an early prerequisite failure and establishes the primary diagnosis, yet the remaining 31 executed actions still make no further reference-path progress. Even after a local prerequisite is resolved, the model may still fall back into stagnation during subsequent execution.

However, these execution failures cannot be simply resolved by stronger local action selection or a larger action budget. In Table \ref{tab:cross-track-performance}, GPT-5.6-Terra achieves the highest Next Actions F1 at 26.18\%, whereas GPT-5.6-SOL scores 22.41\% but ranks first on all four interactive metrics. In addition, the reference policy completes all 10 trajectories within the original 40-action budget, whereas increasing the model budget to 60 actions yields no additional terminal completions. Long-horizon execution requires models to continuously maintain the current clinical state, satisfied and unsatisfied prerequisites, and accumulated progress across multiple steps; if these information states are not updated consistently, subsequent actions may still deviate from the conditions that actually need to be advanced, even when individual actions are reasonable or more action budget is available.

\section{Case Study}
\begin{figure}
    \centering
    \includegraphics[width=\linewidth]{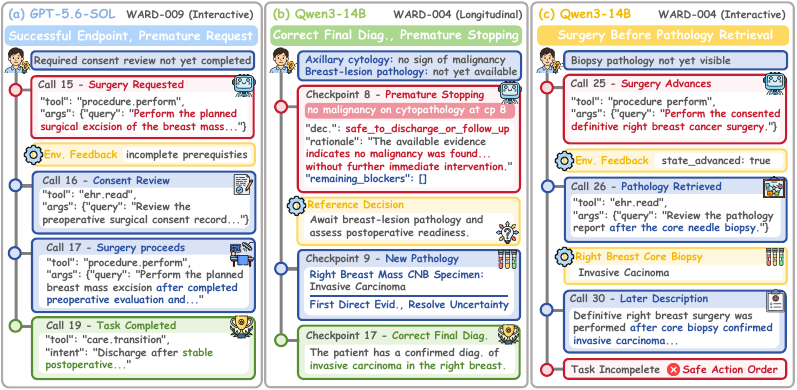}
\caption{
\textbf{Representative WARD failure cases revealing errors.}\textit{checkpoint }is a predefined clinical assessment time point.
(a) A successful Interactive trajectory contains a premature surgical request before consent review.
(b) A correct final diagnosis follows premature stopping while pathology remains pending.
(c) Surgery is executed before the required pathology retrieval, resulting in an unsafe action order and incomplete task execution.}
    \label{fig:case}
\end{figure}
Fig.~\ref{fig:case} presents three representative process-level failure cases in WARD. Even when a model ultimately completes the task or gives the correct diagnosis, it may still make critical errors in intermediate decisions, so final outcomes alone cannot fully reflect the true capabilities of longitudinal clinical agents. These models do not fail to recognize clinical information; rather, they struggle to track unresolved evidence, preconditions, and temporal dependencies among actions, and thus prematurely advance the task state when local evidence appears sufficient.

Cases (a) and (b) illustrate this premature advancement in interactive execution and longitudinal judgment, respectively. In case (a), GPT-5.6-SOL ultimately completes the task, but requests surgery before the required consent review is completed. In case (b), Qwen3-14B eventually identifies the correct breast cancer diagnosis, but judges that the patient can enter discharge or follow-up and sets remaining\_blockers to empty while the breast-lesion pathology has not yet returned, whereas the reference still requires waiting for pathology.

Case (c) exposes a more severe trajectory-level inconsistency. GPT-5.6-Terra performs definitive surgery before the biopsy pathology is visible, and reads the pathology report only afterward, obtaining the invasive carcinoma diagnosis. It then states that definitive surgery followed core biopsy confirmation of carcinoma, contradicting the actual tool-call order. This reflects both unsafe action ordering and a failure to reliably maintain the history of executed states. Together with the quantitative results from WARD, these cases indicate that a main difficulty for current models lies in transforming locally reasonable evidence recognition and single-step decisions into sustained, dependency-explicit, and temporally consistent longitudinal task advancement.
\section{Conclusion}

We introduce WARD, a process-level benchmark for longitudinal clinical agents comprising 4,560 clinical tasks organized into 240 clinical decision checkpoints across 40 longitudinal clinical trajectories. Based on these trajectories, WARD provides two evaluation tracks: the Longitudinal Track assesses clinical judgments under accumulating evidence and how they evolve over time, while the Interactive Track further evaluates whether models can advance complete clinical trajectories through sequential actions in a tool-mediated environment. Experiments reveal a substantial gap along the evidence-to-state-to-action chain: the median Evidence Recognition F1 is 48.42\%, whereas Patient State and Next Actions reach only 17.72\% and 11.81\%, respectively, and linked-evidence nonresponse remains as high as 74.47--90.90\% even when the relevant evidence has been recognized. This instability also persists across longitudinal reasoning and interactive execution: the median Later Match is only 22.23\%, Evidence Regression reaches 37.97\%, and in the Interactive Track only 11.46\% of actions advance a reference transition, while 76.92\% of nonterminal trajectories exhibit repeated or prolonged stagnation. Therefore, the central challenge for longitudinal clinical agents is not only recognizing local evidence or selecting reasonable single-step actions, but also consistently maintaining clinical state, unresolved prerequisites, and accumulated progress across multiple steps.

\section*{AI Use Statement}

Generative AI was used within the research pipeline: GPT-4o generated non-authoritative pre-annotations during benchmark construction, and separate language models served as semantic judges, as described in Sections~3 and~5. All pre-annotations were reviewed by clinicians, and the authors validated the resulting labels, scores, analyses, and citations. ChatGPT was also used for methodological feedback, literature discovery, result interpretation, translation, and manuscript drafting and editing; the authors take full responsibility for the final content.

\section*{Ethics Statement}

This retrospective study used deidentified EHR data and was approved by the responsible institutional review board. The requirement for individual informed consent was waived. Only deidentified case representations were used for benchmark construction and model evaluation, and no model output informed the care of any patient. WARD is intended solely for research evaluation and does not establish that an evaluated model is safe for autonomous clinical use.

\section*{Reproducibility Statement}

WARD's construction, annotation protocol, evaluation procedures, and statistical analyses are documented in Sections~3--5 and the appendices. We provide the prompts, schemas, model and judge configurations, and evaluation code required to reconstruct the evaluation pipeline. 

\bibliography{references}
\bibliographystyle{iclr2027_conference}

\clearpage
\appendix
\clearpage
\setcounter{section}{0}

\begin{center}
{\LARGE\bfseries Appendix}
\end{center}

\vspace{0.8em}
\noindent{\large\bfseries Contents}\par
\vspace{0.25em}
\hrule
\vspace{0.6em}

\begingroup
\small
\setlength{\parindent}{0pt}
\newcommand{\appnavsection}[3]{%
  \par\addvspace{0.45\baselineskip}%
  \noindent\hyperref[#1]{\textbf{#2\quad #3}}\nobreak\dotfill\nobreak\textbf{\pageref{#1}}\par}
\newcommand{\appnavsubsection}[3]{%
  \noindent\hspace*{1.5em}\hyperref[#1]{#2\quad #3}\nobreak\dotfill\nobreak\pageref{#1}\par}
  \appnavsection{app:i-comparison}{A}{Benchmark Comparison}
\appnavsection{app:a-construction}{B}{Data and Annotation}
\appnavsubsection{app:a-source}{B.1}{Data Sources and Case Selection}
\appnavsubsection{app:a-trajectory}{B.2}{Trajectories and Checkpoints}
\appnavsubsection{app:a-annotation}{B.3}{Reference Annotation and Review}
\appnavsubsection{app:a-cohort}{B.4}{Cohort and Case Catalog}
\appnavsubsection{app:a-interactive-subset}{B.5}{Interactive Subset}
\appnavsection{app:b-task-specifications}{C}{Tasks and Information Access}
\appnavsubsection{app:b-longitudinal-track}{C.1}{Longitudinal Inputs and Outputs}
\appnavsubsection{app:b-interactive-track}{C.2}{Interactive Observations and Tools}
\appnavsubsection{app:b-transitions}{C.3}{Prerequisites, Transitions, and Termination}
\appnavsubsection{app:b-visibility}{C.4}{Temporal Visibility and Label Access}
\appnavsection{app:c-evaluation}{D}{Metrics and Scoring}
\appnavsubsection{app:c-matching}{D.1}{Matching Rules}
\appnavsubsection{app:c-checkpoint-metrics}{D.2}{Checkpoint Metrics}
\appnavsubsection{app:c-longitudinal}{D.3}{Longitudinal Diagnostics}
\appnavsubsection{app:c-interactive}{D.4}{Interactive Metrics}
\appnavsubsection{app:c-invalid-output}{D.5}{Invalid Outputs and Eligibility}
\appnavsubsection{app:c-aggregation}{D.6}{Aggregation and Confidence Intervals}
\appnavsection{app:d-models}{E}{Models and Prompts}
\appnavsubsection{app:d-inference}{E.1}{Models and Inference Settings}
\appnavsubsection{app:d-preannotation}{E.2}{Pre-annotation Prompts}
\appnavsubsection{app:d-target-prompts}{E.3}{Target-model Prompts}
\appnavsubsection{app:d-judge-prompts}{E.4}{Judge Prompts and Verdicts}
\appnavsubsection{app:d-execution}{E.5}{Execution and Error Handling}
\appnavsection{app:e-reliability}{F}{Reliability and Clinician Baselines}
\appnavsubsection{app:e-judge-calibration}{F.1}{Judge Calibration}
\appnavsubsection{app:e-agreement}{F.2}{Judge Agreement and Sensitivity}
\appnavsubsection{app:e-clinician-checkpoint}{F.3}{Clinician Checkpoint Baseline}
\appnavsubsection{app:e-interactive-ratings}{F.4}{Interactive Ratings and Agreement}
\appnavsection{app:f-results}{G}{Full Results}
\appnavsubsection{app:f-checkpoint}{G.1}{Checkpoint Results and Output Validity}
\appnavsubsection{app:f-longitudinal}{G.2}{Longitudinal Results and Eligibility}
\appnavsubsection{app:f-stopping}{G.3}{Stopping Decisions and Blockers}
\appnavsubsection{app:f-interactive}{G.4}{Interactive Results and Failures}
\appnavsubsection{app:f-valid-output}{G.5}{Valid-output Sensitivity}
\appnavsection{app:g-environment}{H}{Environment Validation}
\appnavsubsection{app:g-reachability}{H.1}{Reference-policy Reachability}
\appnavsubsection{app:g-budget}{H.2}{Action-budget Sensitivity}
\appnavsubsection{app:g-nonadvancing}{H.3}{Reasonable Non-advancing Actions}
\appnavsubsection{app:g-resolver}{H.4}{Resolver Errors}
\appnavsection{app:h-cases}{I}{Case Studies}
\appnavsubsection{app:h-linked-evidence}{I.1}{Longitudinal: Evidence and State}
\appnavsubsection{app:h-prerequisite}{I.2}{Interactive: Progress and Stagnation}
\endgroup

\clearpage
\section{Detailed Realted Benchmark Comparison}
\label{app:i-comparison}
To position WARD relative to prior clinical benchmarks, we compare them along 7 dimensions covering medical knowledge, EHR use, longitudinal and temporal reasoning, interaction, tool use, and process-level evaluation in Table \ref{tab:related-work-comparison}. Existing benchmarks typically emphasize only a subset of these capabilities, ranging from static medical QA to EHR prediction and interactive tool use. WARD integrates these dimensions within a unified longitudinal clinical-agent evaluation framework.
\begin{table}[htbp]
\centering
\scriptsize
\setlength{\tabcolsep}{2.2pt}
\begin{tabular}{lccccccc}
\toprule
\textbf{Benchmark} &
\textbf{MedKnow.} &
\textbf{EHR} &
\textbf{Long.} &
\textbf{Temporal} &
\textbf{Interact.} &
\textbf{ToolUse} &
\textbf{ProcessEval} \\
\midrule

\multicolumn{8}{c}{\cellcolor[HTML]{E6E6E6}Static Medical Benchmarks} \\
\midrule
PubMedQA \citep{jin2019pubmedqa}
& \greencheck & \redcross & \redcross & \redcross & \redcross & \redcross & \redcross \\

MedQA \citep{jin2021medqa}
& \greencheck & \redcross & \redcross & \redcross & \redcross & \redcross & \redcross \\

MedMCQA \citep{pal2022medmcqa}
& \greencheck & \redcross & \redcross & \redcross & \redcross & \redcross & \redcross \\

MultiMedQA \citep{singhal2023clinical}
& \greencheck & \redcross & \redcross & \redcross & \redcross & \redcross & \redcross \\

\midrule
\multicolumn{8}{c}{\cellcolor[HTML]{E6E6E6}EHR and Temporal Benchmarks} \\
\midrule
EHRBench \citep{xie2026ehrbench}
& \greencheck & \greencheck & \redcross & \redcross & \redcross & \redcross & \redcross \\

EHRSHOT \citep{wornow2023ehrshot}
& \greencheck & \greencheck & \greencheck & \greencheck & \redcross & \redcross & \redcross \\

TempoMed-Bench \citep{guan2026temporal}
& \greencheck & \redcross & \redcross & \greencheck & \redcross & \redcross & \redcross \\

\midrule
\multicolumn{8}{c}{\cellcolor[HTML]{E6E6E6}Interactive Clinical Agent Benchmarks} \\
\midrule
AgentClinic \citep{schmidgall2026agentclinic}
& \greencheck & \redcross & \redcross & \redcross & \greencheck & \greencheck & \redcross \\

MedAgentBench \citep{jiang2025medagentbench}
& \greencheck & \greencheck & \redcross & \redcross & \greencheck & \greencheck & \redcross \\

HealthAgentBench \citep{liu2026healthagentbench}
& \greencheck & \greencheck & \redcross & \redcross & \greencheck & \greencheck & \redcross \\

EHRChatQA \citep{lee2026ehrchatqa}
& \greencheck & \greencheck & \redcross & \redcross & \greencheck & \greencheck & \redcross \\

\midrule
\rowcolor[HTML]{D9E1F2}
WARD
& \greencheck & \greencheck & \greencheck & \greencheck & \greencheck & \greencheck & \greencheck \\

\bottomrule
\end{tabular}

\caption{Comparison of WARD with clinical benchmarks across medical knowledge (MedKnow.), EHR-based tasks (EHR), longitudinal history (Long.), temporal reasoning (Temporal), multi-turn interaction (Interact.), tool use (ToolUse), and intermediate process evaluation (ProcessEval).}
\label{tab:related-work-comparison}
\end{table}
\section{Data and Annotation}
\label{app:a-construction}
\subsection{Data Sources and Case Selection}
\label{app:a-source}
WARD was derived from a deidentified institutional EHR snapshot containing
56,833 inpatient encounters from 138 clinical departments. Construction
separated three quantities: source encounters, clinically screened cases, and cases reconstructed
for evaluation. Clinicians screened for \emph{decision-dense} records with
interpretable changes in patient state and clinically consequential action
dependencies. Retained records had decision points where new evidence could
change the assessment or treatment readiness. It could also change the next
action, a safety constraint, or whether stopping was appropriate.

The final benchmark contains 40 reconstructed longitudinal clinical trajectories selected purposively for
evaluation. This cohort was not designed to estimate disease prevalence in the source
institution. Intermediate candidate counts and reason-specific exclusion
counts are unavailable in the current project summary.
\begin{figure}
    \centering
    \includegraphics[width=\linewidth]{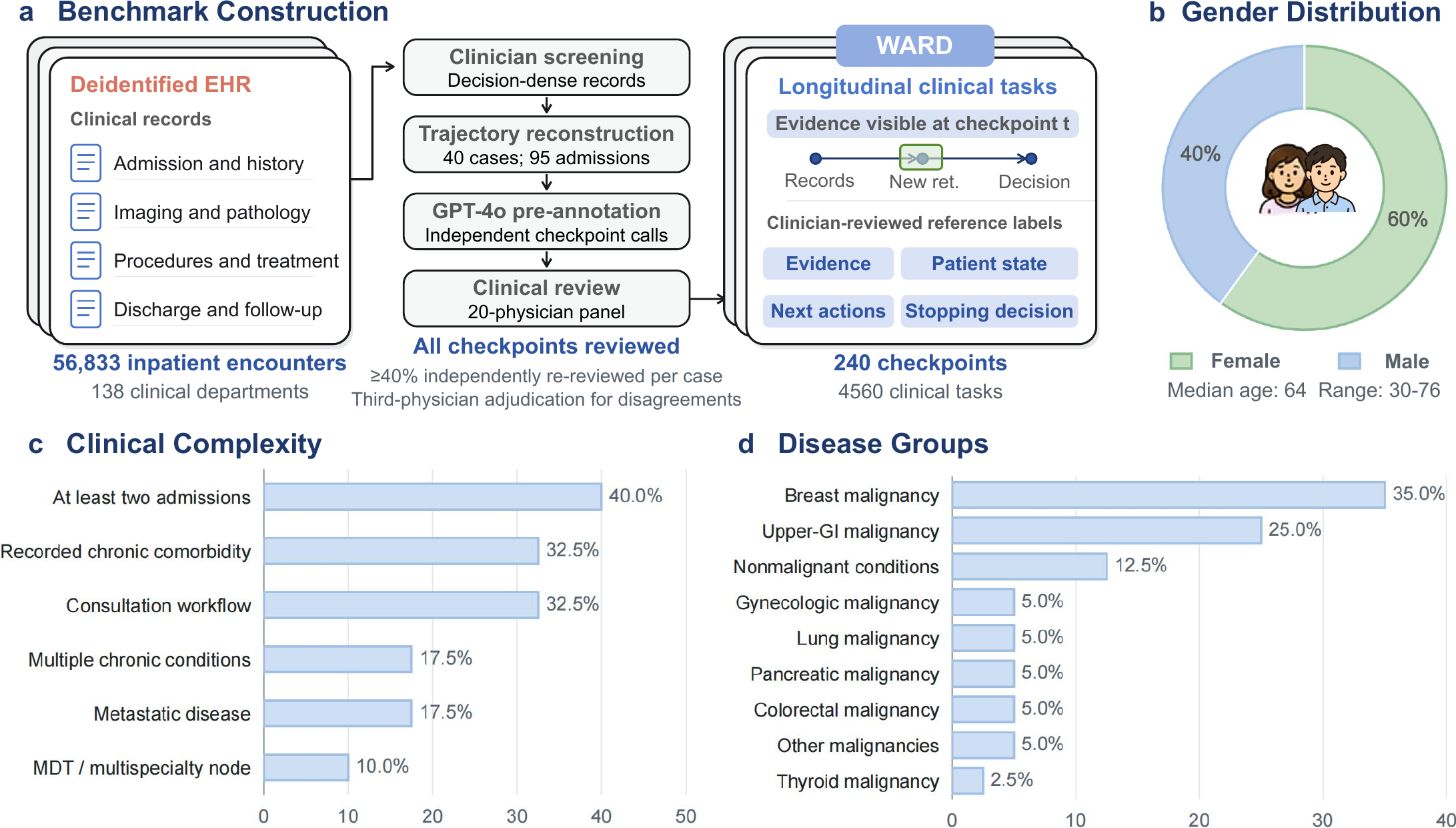}
    \caption*{
        \textbf{Figure A1: Overview of WARD.}
        (a) Benchmark construction and clinical annotation.
        (b) Gender distribution.
        (c) Clinical complexity features, which may overlap.
        (d) Mutually exclusive disease groups.
        Cohort statistics describe all 40 cases.
    }
\end{figure}

\subsection{Trajectories and Checkpoints}
\label{app:a-trajectory}

Records linked to each retained case were reconstructed as one longitudinal clinical trajectory in documented event order. The cohort contains 20 single-index inpatient scenarios and 20 longitudinal oncology scenarios. The latter represent 75 admissions; with the 20 single-index admissions, the benchmark represents 95 hospitalization episodes. Repeated admissions within a trajectory remain in sequence.

Checkpoints mark clinically meaningful decision transitions at nonuniform
intervals. These include new imaging, pathology, or laboratory results and
changes in diagnosis or staging. Other anchors are procedure or treatment
readiness and discharge or follow-up planning. This produced 240 checkpoint tasks across
40 trajectories, with 4--10 checkpoints per trajectory (mean, 6).

WARD separately records \emph{clinical time}, when an event occurred, and
\emph{available time}, when the information became accessible for decision
making. At checkpoint $t$, the public input contains only the cumulative record
available by that boundary. Later evidence, downstream outcomes, and reference
annotations remain hidden. Each checkpoint defines the information available
at a clinical decision point.

\subsection{Reference Annotation and Review}
\label{app:a-annotation}
\label{app:annotation-protocol}

GPT-4o generated provisional annotations for all 240 checkpoints. Each checkpoint was processed independently using only temporally available information. Future evidence, downstream outcomes, reference labels, and clinician answers were hidden. This stage reported 1,019 clinical claims, 4,880 normalized patient-state elements, and 823 candidate actions.

A panel of 20 physicians reviewed every checkpoint against the temporally
eligible record and could accept, revise, or reject each element.
Within every case, at least 40\% of checkpoints underwent an independent second
review. Critical omissions, potentially harmful actions, major safety
constraints, and stopping decisions with unresolved blockers received
additional review regardless of sampling. A third physician adjudicated
disagreements without input from the pre-annotation model. Only
clinician-approved or adjudicated items entered the reference set.

The reference contains relevant evidence and its utilization links, clinical
claims, patient-state elements, and prospective actions. It also records safety
constraints and stopping decisions with blockers. Canonical English expressions
and acceptable aliases are stored where appropriate, along with clinically
important confusions. These support matching of equivalent wording to one-to-one
reference entities. Actions whose omission is
judged clinically consequential are marked critical, with the rationale retained.

\begin{table}
\centering
\scriptsize
\setlength{\tabcolsep}{5pt}
\renewcommand{\arraystretch}{1.08}
\caption*{\textbf{Table A1: Provisional annotation output and clinician-reviewed Gold.}}
\begin{tabular}{@{}lrr@{}}
\toprule
\textbf{Annotation unit} &
\textbf{Provisional pre-annotation} &
\textbf{Clinician-reviewed Gold} \\
\midrule
Evidence records & --- & 730 \\
Clinical claims & 1,019 & 1,019 \\
Patient-state elements & 4,880 & 4,880 \\
Actions & 823 & 823 \\
Action prerequisites & --- & 1,448 \\
Safety entries & --- & 1,813 \\
Stopping rationales & --- & 240 \\
Stopping blockers & --- & 770 \\
Milestone candidates & --- & 200 \\
Milestone criteria & --- & 805 \\
\bottomrule
\end{tabular}

\vspace{2pt}
\begin{minipage}{0.92\linewidth}
\footnotesize
\textit{Note.} \texttt{---} marks a provisional total unavailable in the current
project summary; it does not mean zero. Clinical review could revise content
without changing item counts, so equal totals do not imply unchanged text. The 805 milestone
entries are completion, partial-completion, or safety-invalidation criteria
attached to 200 milestone candidates. Aggregate accept, revise, and reject counts
are unavailable. These Gold totals describe the Longitudinal checkpoint package.
Interactive Track evidence and transition counts are reported separately in Section A.4.
\end{minipage}
\end{table}

\subsection{Cohort and Case Catalog}
\label{app:a-case-catalog}
\label{app:a-cohort}
\label{app:case-catalog}
\label{app:case_taxonomy}

The 40-case cohort spans surgical and medical-oncology pathways with substantial
longitudinal and coordination burden. Table A2 reports descriptive,
non-exclusive features whose percentages should not be summed. Table A3
provides the deidentified case catalog. Checkpoints are complete Longitudinal Track tasks, each comprising field-level clinical tasks grouped into seven clinical output domains.
Evidence records and reference transitions describe the recorded Interactive
Track representation.

\begin{table}
\centering
\scriptsize
\setlength{\tabcolsep}{4.2pt}
\renewcommand{\arraystretch}{1.07}
\caption*{\textbf{Table A2: Cohort characteristics and clinical complexity.}}
\begin{tabular}{@{}lrlr@{}}
\toprule
\textbf{Characteristic} & \textbf{Value} &
\textbf{Characteristic} & \textbf{Value} \\
\midrule
Longitudinal trajectories & 40 & Hospitalizations & 95 \\
Checkpoint tasks & 240 & Checkpoints per trajectory & 4--10 (mean 6) \\
Field-level clinical tasks & 4,560 & Clinical output domains & 7 \\
Female & 24 (60.0\%) & Male & 16 (40.0\%) \\
Age, years & 64 median & Age range, years & 30--76 \\
Age $\geq 65$ years & 20 (50.0\%) & Age $\geq 70$ years & 11 (27.5\%) \\
\midrule
Surgical/procedural pathway & 20 (50.0\%) & Systemic anticancer therapy & 20 (50.0\%) \\
Breast malignancy & 14 (35.0\%) & Upper-GI malignancy & 10 (25.0\%) \\
Gynecologic malignancy & 2 (5.0\%) & Lung malignancy & 2 (5.0\%) \\
Pancreatic malignancy & 2 (5.0\%) & Colorectal malignancy & 2 (5.0\%) \\
Other malignancies & 2 (5.0\%) & Thyroid malignancy & 1 (2.5\%) \\
Nonmalignant conditions & 5 (12.5\%) & Metastatic disease & 7 (17.5\%) \\
Explicit chronic comorbidity & 13 (32.5\%) & Multiple chronic conditions & 7 (17.5\%) \\
At least two admissions & 16 (40.0\%) & At least four admissions & 10 (25.0\%) \\
At least six admissions & 5 (12.5\%) & Maximum admissions in one case & 10 \\
Explicit consultation workflow & 13 (32.5\%) & Explicit MDT/multispecialty node & 4 (10.0\%) \\
Imaging evidence & 31 (77.5\%) & Pathology evidence & 15 (37.5\%) \\
Laboratory evidence & 21 (52.5\%) & High evidence density ($\geq15$) & 21 (52.5\%) \\
Long action graph ($\geq12$) & 17 (42.5\%) & Cases with malignancy & 35 (87.5\%) \\
\bottomrule
\end{tabular}

\vspace{2pt}
\begin{minipage}{0.94\linewidth}
\footnotesize
\textit{Note.} Disease-focus counts form a mutually exclusive grouping; pathway, complexity, evidence, and coordination features are non-exclusive. Median age is 64 years (IQR, 56.8--70.0), and the mean is 61.15.
\end{minipage}
\end{table}

\begin{table}
\centering
\scriptsize
\setlength{\tabcolsep}{3.2pt}
\renewcommand{\arraystretch}{1.04}
\caption*{\textbf{Table A3: Deidentified case catalog (WARD-001--WARD-020).}}
\begin{tabularx}{\linewidth}{@{}l c X X r r r r@{}}
\toprule
\textbf{Case} & \textbf{Int.} & \textbf{Decision challenge} &
\textbf{Evidence / complexity} & \textbf{Adm.} & \textbf{CP} & \textbf{E} & \textbf{T} \\
\midrule
WARD-001 & $\checkmark$ & Multisite workup to breast-conserving surgery & Imaging/pathology; DM & 1 & 10 & 25 & 12 \\
WARD-002 & $\checkmark$ & Nodal confirmation and arrhythmia-risk planning & Pathology; consultation; AF/HTN/DM & 1 & 7 & 31 & 19 \\
WARD-003 & $\checkmark$ & Discordant FNA, nodal mapping, and thyroidectomy & Cytology/imaging; consultation & 1 & 5 & 23 & 13 \\
WARD-004 & $\checkmark$ & Cancer planning with an incidental lung nodule & Pathology/chest imaging; thoracic consult & 1 & 7 & 29 & 18 \\
WARD-005 & $\checkmark$ & Bone-lesion staging before definitive surgery & PET/CT; bone biopsy; 27 transitions & 1 & 8 & 41 & 27 \\
WARD-006 & $\checkmark$ & Bilateral-lesion operative planning & Pathology; bilateral surgery; anesthesia & 1 & 5 & 25 & 12 \\
WARD-007 & $\checkmark$ & Pathology-guided multidisciplinary surgery & Pathology/multimodal imaging; MDT & 1 & 5 & 29 & 12 \\
WARD-008 & $\checkmark$ & Combined breast--gynecologic surgery & Cross-specialty records; MDT & 1 & 5 & 19 & 9 \\
WARD-009 & $\checkmark$ & Imaging-guided conservation/sentinel planning & Imaging/consent; MDT & 1 & 5 & 17 & 8 \\
WARD-010 & $\checkmark$ & Mastectomy, reconstruction, and margin assessment & Intraoperative pathology; reconstructive MDT & 1 & 5 & 32 & 11 \\
WARD-011 & & Biliary surgery with delayed imaging & Laboratory/US/MRI; postoperative monitoring & 1 & 9 & 8 & 8 \\
WARD-012 & & Wire localization and lung resection & Localization/intraoperative pathology/post-op imaging & 1 & 8 & 9 & 9 \\
WARD-013 & & Multifocal assessment with intraoperative pathology & US/MRI/intraoperative pathology; consultation & 1 & 9 & 10 & 10 \\
WARD-014 & & Lung resection, nodal dissection, and post-op imaging & CT/intraoperative pathology/post-op monitoring & 1 & 7 & 8 & 8 \\
WARD-015 & & Diarrhea workup, EMR, and antifungal treatment & Laboratory/microbiology/endoscopy & 1 & 8 & 11 & 11 \\
WARD-016 & & Biopsy-to-mastectomy sequencing & Imaging/pathology/drains; HTN/thyroiditis & 1 & 5 & 10 & 10 \\
WARD-017 & & Mastectomy, nodal surgery, and skin graft & Imaging/pathology; DM/tachycardia & 1 & 5 & 10 & 10 \\
WARD-018 & & Thrombosis and nodal workup before mastectomy & VTE imaging/cytology/pathology & 1 & 5 & 11 & 11 \\
WARD-019 & & Wire localization and benign-lesion excision & Imaging/localization/intraoperative pathology & 1 & 5 & 9 & 8 \\
WARD-020 & & Arthroplasty, VTE/anemia monitoring, and rehabilitation & Imaging/anticoagulation/rehabilitation & 1 & 5 & 11 & 10 \\
\bottomrule
\end{tabularx}

\vspace{2pt}
\begin{minipage}{0.95\linewidth}
\footnotesize
\textit{Abbreviation Key:} Int. = included in the prespecified Interactive Track;
Adm. = represented admissions; CP, Longitudinal Track checkpoints; E = reference
evidence records; T = reference transitions; AF = atrial fibrillation; HTN = 
hypertension; DM = diabetes; FNA = fine-needle aspiration; EMR = endoscopic
mucosal resection; VTE = venous thromboembolism; MDT = multidisciplinary team;
ILD = interstitial lung disease; US = ultrasonography.
\end{minipage}
\end{table}

%\clearpage

\begin{table}
\centering
\scriptsize
\setlength{\tabcolsep}{3.2pt}
\renewcommand{\arraystretch}{1.04}
\caption*{\textbf{Table A3 (continued): Deidentified case catalog (WARD-021--WARD-040).}}
\begin{tabularx}{\linewidth}{@{}l c X X r r r r@{}}
\toprule
\textbf{Case} & \textbf{Int.} & \textbf{Decision challenge} &
\textbf{Evidence / complexity} & \textbf{Adm.} & \textbf{CP} & \textbf{E} & \textbf{T} \\
\midrule
WARD-021 & & Serial readiness across metastatic ovarian therapy & 4 admissions; imaging/lab; HTN/asthma & 4 & 5 & 21 & 14 \\
WARD-022 & & Longitudinal therapy for bone/liver metastatic disease & 6 admissions; imaging/lab; multisite metastases & 6 & 7 & 31 & 20 \\
WARD-023 & & Treatment readiness with marrow suppression & Laboratory monitoring; HTN & 1 & 5 & 5 & 5 \\
WARD-024 & & Serial treatment assessment across admissions & 2 admissions; imaging/laboratory & 2 & 5 & 11 & 8 \\
WARD-025 & & Therapy under ILD and nodal metastatic burden & Imaging/lab; ILD; nodal metastases & 1 & 4 & 5 & 4 \\
WARD-026 & & Chemotherapy-to-immunotherapy transition & 9 admissions; laboratory; leukopenia & 9 & 8 & 38 & 29 \\
WARD-027 & & Treatment with liver and nodal metastases & 4 admissions; imaging/lab; metastatic & 4 & 5 & 21 & 14 \\
WARD-028 & & Treatment under prior stroke and hypertension & 4 admissions; laboratory; prior stroke/HTN & 4 & 5 & 17 & 14 \\
WARD-029 & & Repeated readiness and response assessment & 3 admissions; imaging/laboratory & 3 & 5 & 16 & 11 \\
WARD-030 & & Metastatic endometrial treatment course & 2 admissions; prior cancer; sleep disorder & 2 & 5 & 7 & 6 \\
WARD-031 & & Six-admission treatment and response course & 6 admissions; imaging/laboratory & 6 & 7 & 26 & 19 \\
WARD-032 & & Treatment amid bleeding, obstruction, and nutritional risk & Laboratory; bleeding/obstruction/nutrition & 2 & 5 & 9 & 8 \\
WARD-033 & & Long-course treatment with leukopenia & 7 admissions; imaging/lab; leukopenia & 7 & 8 & 32 & 23 \\
WARD-034 & & Single-admission treatment readiness and outcome & Imaging/laboratory & 1 & 5 & 6 & 5 \\
WARD-035 & & Rectal-cancer treatment readiness and outcome & Laboratory monitoring & 1 & 5 & 5 & 5 \\
WARD-036 & & Treatment monitoring with leukopenia & 2 admissions; imaging/lab; leukopenia & 2 & 5 & 10 & 8 \\
WARD-037 & & Pancreatic cancer with liver metastasis & 4 admissions; laboratory; metastatic & 4 & 5 & 17 & 14 \\
WARD-038 & & Breast cancer with bone metastasis & 2 admissions; imaging/lab; metastatic & 2 & 5 & 11 & 8 \\
WARD-039 & & Repeated pancreatic treatment assessment & 4 admissions; imaging/laboratory & 4 & 5 & 21 & 14 \\
WARD-040 & & Ten-admission course with emergent liver metastasis & 10 admissions; imaging/lab; HTN; metastatic & 10 & 8 & 51 & 32 \\
\bottomrule
\end{tabularx}

\vspace{2pt}
\begin{minipage}{0.95\linewidth}
\footnotesize
\textit{Note.} Decision challenges summarize each case's main sequential problem.
Evidence and complexity descriptors are non-exclusive summaries of the evidence
modality, treatment, metastasis, comorbidity, consultation, and repeated-admission
fields.
\end{minipage}
\end{table}

\subsection{Interactive Subset}
\label{app:a-interactive-subset}

The Interactive Track uses a fixed, prespecified subset of WARD-001 through
WARD-010. The ten tool-mediated clinical tasks use ten of the 40 longitudinal clinical
trajectories, which also appear in the Longitudinal Track. The subset was specified before the reported
target-model interactive runs and contributes 62 of the 240 Longitudinal Track
checkpoints.

The subset was selected for extended, tool-mediated inpatient trajectories
with explicit evidence release and action dependencies. It comprises nine breast-focused surgical cases and one thyroid-focused surgical case. Nine of the ten
have a malignant final disease group (eight breast and one thyroid), and one is
nonmalignant. All ten are single-index inpatient scenarios. The subset tests
within-episode action sequencing and does not cover repeated-admission burden.

Across the ten interactive tasks, the case representations contain 271 verifiable
evidence records and 141 reference transitions. The medians are 27 evidence
records and 12 transitions per case; the maximum is 41 evidence records and 27
transitions. Nine cases contain an explicit consultation workflow, four contain
an explicit multidisciplinary or multispecialty node, all ten meet the
prespecified high-evidence-density threshold of at least 15 records, and seven
contain an action graph with at least 12 reference transitions. The subset
supports evaluation of prerequisite resolution and sustained clinical progress.
Its specialty concentration limits clinical-domain representativeness.

\section{Tasks and Information Access}
\label{app:b-task-specifications}
\raggedbottom

The two WARD tracks test complementary behaviors. The Longitudinal Track
requests an independent structured clinical assessment at each predefined
checkpoint. The Interactive Track requires executable tool calls in a
persistent Patient/EHR environment. Table~\hyperref[tab:checkpoint-fields-b1]{B1} summarizes
their information and interaction contracts. Complete prompts appear in
Appendix~\ref{app:d-models}.

\begin{table}
\centering
\caption*{\textbf{Table B: Complementary WARD evaluation tracks.}}
\label{tab:evaluation-tracks}
\small
\begin{tabularx}{\linewidth}{@{}lXX@{}}
\toprule
\textbf{Property} &
\textbf{Longitudinal Track} &
\textbf{Interactive Track} \\
\midrule
Evaluation unit &
One checkpoint submission &
One model--task trajectory \\
Benchmark scale &
4,560 field-level clinical tasks across seven clinical output domains; 240 checkpoints from 40 longitudinal trajectories spanning 95 hospitalizations &
10 tool-mediated clinical tasks from the prespecified subset; 271 verifiable evidence items and 141 reference transitions. \\
Visible context &
Cumulative public evidence available at the checkpoint &
Current observation, released evidence, feedback, and memory \\
Model history &
No prior model response &
Persistent public feedback and memory \\
Model behavior &
Structured assessment and action plan &
One tool action per turn \\
Primary construct &
Evidence use, state representation, action selection, safety, and stopping &
Evidence acquisition, prerequisite resolution, action sequencing, and sustained progress \\
Experimental scale &
15 attempted models $\times$ 240 checkpoint submissions; 12 models with valid outputs &
\(4 \times 10 = 40\) model--task runs \\
\bottomrule
\end{tabularx}
\end{table}

\subsection{Longitudinal Inputs and Outputs}
\label{app:b-longitudinal-track}

Each case has 4--10 clinically defined checkpoints. The target model is called
independently at each checkpoint. Its one input contains the case and checkpoint
identifiers, public time, compact public context, and all evidence available by
that boundary. Each evidence record has an immutable identifier, record kind,
title, clinical time, available time, and content. Visible evidence is cumulative:
released records remain available at later checkpoints. An empty array is a
valid initial information state; the request is still complete.

Future-forbidden evidence identifiers, future records, and downstream outcomes
are hidden. The model also has no access to provisional annotations, reference
answers, or its earlier checkpoint responses. Later predictions therefore
respond independently to a richer public record, without an earlier answer to revise.

\begin{tcolorbox}[
  colback=wardlightblue,
  colframe=wardblue,
  boxrule=0.6pt,
  arc=1.2mm,
  left=2mm,
  right=2mm,
  top=1.5mm,
  bottom=1.5mm,
  title=\textbf{Illustrative checkpoint interface},
  fonttitle=\small
]
\begin{minipage}[t]{0.47\linewidth}
\raggedright\scriptsize
\textbf{Model input}\par\vspace{2pt}
{\ttfamily
\{\par
\quad case\_id: ward\_001,\par
\quad checkpoint\_id: ward\_001\_cp\_001,\par
\quad public\_time: \ldots,\par
\quad public\_context: \{\ldots\},\par
\quad visible\_evidence: [\par
\qquad \{evidence\_id: patient:001,\par
\qquad\ \ record\_kind: \ldots,\par
\qquad\ \ clinical\_time: \ldots,\par
\qquad\ \ available\_time: \ldots,\par
\qquad\ \ content: \ldots\}\par
\quad ]\par
\}\par}
\end{minipage}\hfill
\begin{minipage}[t]{0.49\linewidth}
\raggedright\scriptsize
\textbf{Required complete submission}\par\vspace{2pt}
{\ttfamily
\{\par
\quad key\_evidence: [\ldots],\par
\quad utilization\_links: [\ldots],\par
\quad claims: [\ldots],\par
\quad patient\_state: \{\ldots\},\par
\quad actions: [\ldots],\par
\quad safety: \{\ldots\},\par
\quad stopping: \{\ldots\}\par
\}\par}
\vspace{4pt}
Local \texttt{pred\_*} identifiers connect
claims, state elements, and actions within the submission.
\end{minipage}
\end{tcolorbox}

A complete checkpoint submission contains 19 required clinical content fields across
seven clinical output domains. A field-level clinical task consists of completing
one required clinical content field at a given checkpoint. Across 240 checkpoints, this yields
4,560 field-level clinical tasks. Each list-valued field is counted once,
regardless of its number of entries. Identifiers and item attributes are not counted separately. Where the public record supports no item, fields may contain explicit
empty arrays or nullable state objects. All evidence references must use an
identifier in the checkpoint's visible-evidence array.

\begin{table}
\centering
\scriptsize
\setlength{\tabcolsep}{4.5pt}
\renewcommand{\arraystretch}{1.08}
\label{tab:evaluation-tracks}
\phantomsection\label{tab:checkpoint-fields-b1}\caption*{\textbf{Table B1: Fields in a complete Longitudinal checkpoint submission.}}
\begin{tabularx}{\linewidth}{@{}>{\raggedright\arraybackslash}p{0.19\linewidth} r >{\raggedright\arraybackslash}X >{\raggedright\arraybackslash}X@{}}
\toprule
\textbf{Domain} & \textbf{Tasks} & \textbf{Required content} & \textbf{Role in the task} \\
\midrule
\texttt{key\_evidence} & 1 &
Visible evidence identifiers, a relevance role, and an optional interpretation &
Selects the public records that matter at the current decision point. \\
\rowcolor{wardlightblue}
\texttt{utilization\_links} & 1 &
Links one visible evidence identifier to local claim, state, and action identifiers &
Tests whether retrieved evidence is connected to downstream reasoning rather than merely repeated. \\
\texttt{claims} & 1 &
Clinical assertions with a local identifier, status, and supporting evidence identifiers &
Represents what the model concludes under the evidence currently available. \\
\rowcolor{wardlightblue}
\texttt{patient\_state} & 9 &
Diagnoses, active and resolved problems, blockers, facts to retain or correct, required state delta, treatment readiness, and management state &
Captures the model's structured representation of the current patient state. \\
\texttt{actions} & 1 &
Prospective, recorded, or avoid actions with category, grounding, and critical-omission flag; indication, target, timing, priority, and prerequisites may also be specified &
Represents clinically appropriate next steps and actions that should not be taken. \\
\rowcolor{wardlightblue}
\texttt{safety} & 3 &
Hard-failure rules, critical-omission rules, and preliminary risk triggers &
Records constraints that make an answer unsafe or clinically incomplete. \\
\texttt{stopping} & 3 &
One decision, remaining blockers, and a rationale &
Distinguishes continued investigation, continued treatment, waiting for a pending result, and safe discharge or follow-up. \\
\midrule
\textbf{Total} & \textbf{19} & Seven clinical output domains per checkpoint & \\
\bottomrule
\end{tabularx}
\end{table}

The submission requires a structured clinical answer and excludes hidden chain
of thought. Missing or unparseable outputs retain an explicit completion status,
such as schema-invalid, missing, refused, truncated, or transport-error.
Appendix~\ref{app:c-invalid-output} specifies their evaluation.

\subsection{Interactive Observations and Tools}
\label{app:b-interactive-track}
\label{app:interactive-environment}

The Interactive Track uses a persistent Patient/EHR environment with deterministic evidence release. At each turn, the model receives the public observation and issues exactly one native clinical tool call. The environment resolves the call before constructing the next observation. Returned outcomes, environment-managed memory, and recent attempts remain available on later turns.

The public observation identifies the case, public inpatient stage, environment
clock, and public context. It includes cumulative and newly released evidence
identifiers, compact visible evidence, the previous public tool outcome, the tool
interface, compact public memory, and the terminal flag. Deterministic evidence
compaction places new records first. It retains the title, clinical and available
times, conclusion, and content head and tail, without LLM-generated clinical facts.

All Interactive turns use the same system prompt. The dynamic user message
supplies the public stage and workflow checklist using one case-independent mapping:

\begin{center}
\scriptsize
\setlength{\tabcolsep}{7pt}
\renewcommand{\arraystretch}{1.08}
\begin{tabular}{@{}ll@{}}
\toprule
\textbf{Public stage} & \textbf{Generic workflow checklist} \\
\midrule
S1 & History; admission assessment \\
\rowcolor{wardlightblue}
S2 & Diagnosis; staging; safety; procedures; results; treatment \\
S3 & Results; treatment; disposition \\
\bottomrule
\end{tabular}
\end{center}

The checklist provides a generic inpatient workflow for the Interactive 
Case-specific Gold actions and the reference path remain hidden. The model
receives workflow guidance throughout the task.

\begin{table}
\centering
\scriptsize
\setlength{\tabcolsep}{4.2pt}
\renewcommand{\arraystretch}{1.08}
\caption*{\textbf{Table B2: Interactive observation and action contract.}}
\begin{tabularx}{\linewidth}{@{}>{\raggedright\arraybackslash}p{0.22\linewidth} >{\raggedright\arraybackslash}X >{\raggedright\arraybackslash}X@{}}
\toprule
\textbf{Interface object} & \textbf{Public content} & \textbf{Operational meaning} \\
\midrule
Current observation &
Stage, clock, public context, visible and newly released evidence, last public outcome, public memory, and \texttt{done} &
Defines everything the target model may use when selecting the next action. \\
\rowcolor{wardlightblue}
Public tools &
\texttt{patient.ask}, \texttt{ehr.search}, \texttt{ehr.read}, \texttt{ehr.order}, \texttt{procedure.perform}, \texttt{treatment.start}, and \texttt{care.transition} &
Provides a case-reusable facade for communication, record access, orders, procedures, treatment, and disposition. \\
Tool parameters &
Required \texttt{tool}; a non-empty executable \texttt{args.query}; and, where relevant, \texttt{args.module}. Optional fields are \texttt{intent}, \texttt{rationale}, and visible \texttt{evidence\_ids}. &
The command in \texttt{args.query}, together with structured command fields, is the routing surface. Intent and rationale are descriptive context and cannot substitute for an incomplete command. \\
\rowcolor{wardlightblue}
Public EHR modules &
History, admission and progress documents, preoperative documents, laboratory, imaging, pathology, procedure records, medications, consultations, surgery, and discharge documents &
Constrains module selection without revealing which hidden recorded action is currently available. \\
Public memory &
Compact summaries of completed, prerequisite-pending, unrecorded, and invalid calls; recent no-repeat targets; and retryable pending families &
Carries public interaction history across turns. Memory is updated by the environment and cannot be written directly by the model. \\
\rowcolor{wardlightblue}
Recent attempts &
A compact record of recent public actions and outcomes &
Supports recovery from malformed, repeated, or non-advancing actions without exposing hidden transition structure. \\
\bottomrule
\end{tabularx}
\end{table}

The action vocabulary is broad. The target receives no menu of available
Gold-derived actions or action counts. Transition and concept identifiers,
the dependency graph, and family-to-tool hints remain hidden. The model must
form a clinically grounded command from the visible record. A timing-sensitive
query can specify the record kind, clinical event, lifecycle status (ordered,
performed, completed, or resulted), and before/during/after relation. Valid
actions require no hidden identifier.

After each executed call, the environment returns the public status and a short
message. It reports whether the action was understood, whether state advanced,
and which evidence identifiers were released. Table B3 defines the outcome
classes. Revisiting the same public clinical target without intervening
advancement returns a semantic-repeat flag. The flag updates memory without
blocking execution. A revisit justified by new evidence is recorded separately.

\begin{table}
\centering
\scriptsize
\setlength{\tabcolsep}{3.8pt}
\renewcommand{\arraystretch}{1.08}
\caption*{\textbf{Table B3: Public Interactive outcome classes.}}
\begin{tabularx}{\linewidth}{@{}>{\raggedright\arraybackslash}p{0.29\linewidth}c c >{\raggedright\arraybackslash}X@{}}
\toprule
\textbf{Public status} & \textbf{Advance} & \textbf{Release} & \textbf{Meaning visible to the target model} \\
\midrule
\makecell[l]{\texttt{ehr\_path\_advanced}} & Yes & Possible &
The action matched a currently available recorded EHR event; the public state advances and associated evidence may be released. \\
\rowcolor{wardlightblue}
\makecell[l]{\texttt{recognized\_completed}\\\texttt{action}} & No & No &
The clinical concept is already represented in completed recorded state. \\
\makecell[l]{\texttt{recognized\_prerequisite}\\\texttt{pending}} & No & No &
The action is understood, but at least one recorded prerequisite is not yet satisfied. \\
\rowcolor{wardlightblue}
\makecell[l]{\texttt{valid\_exploration}\\\texttt{no\_recorded\_outcome}} & No & No &
The action is clinically valid, but the recorded episode contains no currently releasable result; no counterfactual result is invented. \\
\makecell[l]{\texttt{invalid\_evidence}\\\texttt{grounding}} & No & No &
The call cites an evidence identifier that is not currently visible. \\
\rowcolor{wardlightblue}
\makecell[l]{\texttt{invalid\_tool}\\\texttt{contract}} & No & No &
The output cannot be executed under the native tool-call contract. \\
\bottomrule
\end{tabularx}
\end{table}

\subsection{Prerequisites, Transitions, and Termination}
\label{app:b-transitions}

The Interactive environment keeps its partial-order representation of the
recorded episode hidden from the model. Each transition specifies a clinical
concept, prerequisite state, public state update, and evidence released on
completion. Multiple transitions may be available in a state, so the task
permits more than one linear sequence.

The environment validates each call's native function and visible-evidence
grounding. It normalizes the public tool, module, executable query, and
structured command fields into a clinical action. The resolver then compares
the action with transitions available in the current hidden state:

\begin{tcolorbox}[
  colback=black!2,
  colframe=wardblue!75!black,
  boxrule=0.6pt,
  arc=1.2mm,
  left=2mm,
  right=2mm,
  top=1.5mm,
  bottom=1.5mm
]
\centering\scriptsize
\textbf{Public observation}
$\longrightarrow$
\textbf{one native tool call}
$\longrightarrow$
\textbf{contract and grounding checks}
$\longrightarrow$
\textbf{hidden state-based resolver}
$\longrightarrow$
\textbf{public outcome, memory, and next observation}

\vspace{4pt}
\begin{tabularx}{0.98\linewidth}{@{}>{\raggedright\arraybackslash}p{0.27\linewidth}>{\raggedright\arraybackslash}X@{}}
Matched available transition & Advance public state; release only the evidence associated with the recorded transition. \\
Known target with unmet dependency & Return prerequisite pending; preserve state and reveal no dependency identity. \\
Target already completed & Return completed-action status; preserve state and release no evidence. \\
Clinically valid target without a recorded result & Return valid exploration; preserve state and invent no result. \\
Invalid contract or grounding & Return an invalid public status; do not execute a clinical transition. \\
\end{tabularx}
\end{tcolorbox}

Evidence release depends on recorded events. Only a successfully resolved
transition can add evidence, change the public stage or clock, or mark the
episode terminal. Prerequisite-pending and valid-exploration outcomes can be
clinically reasonable. They disclose no hidden prerequisite and produce no result
absent from the source EHR. The next turn provides the generic outcome and
updated memory. The model can choose another supported action or reformulate
the same target once with clearer event and temporal details.

A trajectory ends at the terminal state, action-budget exhaustion, or an
unrecoverable execution failure. The prespecified budget was 40 actions for all
cases. Repeated, non-advancing, structurally valid
actions consume budget and remain in the trajectory.

\subsection{Temporal Visibility and Label Access}
\label{app:b-visibility}
\label{app:visibility-audit}

WARD records \emph{clinical time}, when an event occurred, and
\emph{available time}, when its record became usable for model decisions.
Available time determines checkpoint visibility and Interactive evidence release.
A note about an earlier procedure becomes visible only when documented or
released. The model cannot infer a result from elapsed clinical time alone.

Table B4 specifies how each component accesses information. \emph{Input} is
supplied to the component; \emph{produced} is generated by it.
\emph{Resolver-only} and \emph{post-hoc} information is unavailable to the target
model during generation.

\begin{table}
\centering
\scriptsize
\setlength{\tabcolsep}{2.6pt}
\renewcommand{\arraystretch}{1.10}
\caption*{\textbf{Table B4: Information-visibility and label-access matrix.}}
\begin{tabularx}{\linewidth}{@{}>{\raggedright\arraybackslash}p{0.16\linewidth}*{5}{>{\raggedright\arraybackslash}X}@{}}
\toprule
\textbf{Information object} &
\textbf{Longitudinal target} &
\textbf{Interactive target} &
\textbf{Environment / resolver} &
\textbf{Semantic Judge} &
\textbf{Offline evaluation} \\
\midrule
Current public context, stage, and time &
Checkpoint input (stage only when public) &
Turn input &
Maintains public view &
Not supplied &
Available for post-hoc analysis \\
\rowcolor{wardlightblue}
Evidence released by the current boundary &
Cumulative input &
Cumulative input; new items identified &
Controls release &
Not supplied &
Grounding and visibility analysis \\
Earlier target-model outputs or actions &
Not supplied &
Public action history summarized in memory and recent attempts &
Stores action history &
No trajectory history &
Submitted predictions or recorded public outcomes \\
\rowcolor{wardlightblue}
Public feedback and public memory &
Not applicable &
Persistent turn input &
Produced and updated &
Not supplied &
Trajectory analysis input \\
Current checkpoint submission &
Produced: seven clinical output domains &
Not applicable &
Not used &
Candidate items supplied for semantic pairs &
Schema validation and scoring \\
\rowcolor{wardlightblue}
Current interactive tool call &
Not applicable &
Produced: one native call &
Executed and resolved &
Not supplied to checkpoint Judge &
Action and trajectory analysis \\
Future or unreleased evidence &
Not supplied &
Not supplied &
Retained until a valid release transition &
Not supplied &
Post-hoc leakage analysis only \\
\rowcolor{wardlightblue}
Gold and acceptable aliases &
Not supplied &
Not supplied &
Held only by the environment and unavailable to the target model &
Reference item and aliases supplied &
Reference for scoring \\
Hidden transition IDs, concept IDs, prerequisites, and resolver features &
Not applicable &
Not supplied &
Resolver-only execution state &
Not supplied &
Post-hoc environment analysis \\
\rowcolor{wardlightblue}
Judge verdicts and derived analysis labels &
Not supplied &
Not supplied &
Primitive public outcomes only &
Verdict produced, not an input to generation &
Derived after target-model outputs are collected \\
\bottomrule
\end{tabularx}
\end{table}

The semantic Judge evaluates unmatched Longitudinal items after deterministic
exact and alias matching. It receives dimension and checkpoint identifiers, one
candidate expression, one reference expression, and accepted reference aliases.
The patient timeline, future evidence, hidden Interactive state, dependency
graph, and analysis labels remain unavailable. The environment generates
Interactive public outcomes. Aggregate action categories, milestone coverage,
recovery indicators, clinical ratings, and failure-mechanism labels are computed
after trajectory collection.

These boundaries prevent three forms of leakage: access to reference answers or
future documentation during action or evidence selection; queries for hidden
transitions or prerequisites; and evaluation-derived labels influencing
generation. Interactive public memory retains prior action outcomes.
Independent Longitudinal calls exclude prior model answers.

\flushbottom

\section{Metrics and Scoring}
\label{app:c-evaluation}

WARD reports each construct separately, with no composite score. Unless labeled
action-pooled, metrics are calculated within cases and averaged with equal case
weights. A \emph{scorable} checkpoint has a schema-valid model submission for
the analysis. Valid empty predictions are distinguished from invalid or missing
outputs.

\begin{tcolorbox}[
  colback=black!2,
  colframe=wardblue!75!black,
  boxrule=0.6pt,
  arc=1.2mm,
  left=2mm,right=2mm,top=1.5mm,bottom=1.5mm
]
\small
\textbf{Common reporting specification.}
For every metric below we state, in order: (i) the evaluation unit,
(ii) the eligibility rule, (iii) the numerator, (iv) the denominator,
(v) the aggregation rule, (vi) zero-denominator and missing-output handling,
and (vii) whether higher or lower values are better.  Percentages in the main
paper are the corresponding proportions multiplied by 100.
\end{tcolorbox}

\subsection{Matching Rules}
\label{app:c-matching}

Predictions receive credit through the matching cascade in Table C1.
Normalization applies Unicode-aware case folding, removes punctuation, and
collapses and trims whitespace. The normalized prediction is compared with the
normalized canonical Gold expression and all clinician-registered aliases.
Only unresolved pairs proceed to semantic judging.

\begin{table}
\centering
\scriptsize
\setlength{\tabcolsep}{3.2pt}
\renewcommand{\arraystretch}{1.11}
\caption*{\textbf{Table C1: One-to-one matching cascade.}}
\begin{tabularx}{\linewidth}{@{}>{\raggedright\arraybackslash}p{0.16\linewidth}>{\raggedright\arraybackslash}p{0.24\linewidth}>{\raggedright\arraybackslash}X>{\raggedright\arraybackslash}p{0.19\linewidth}@{}}
\toprule
\textbf{Stage} & \textbf{Applies to} & \textbf{Rule} & \textbf{Scoring consequence} \\
\midrule
Identifier match & Evidence & Exact equality of the pair
$(\texttt{evidence\_id},\texttt{role})$ after set deduplication. & Matched pair is one TP. \\
\rowcolor{wardlightblue}
Normalized exact / alias & Claims, state, actions, safety, and stopping blockers &
Normalized prediction equals the canonical reference text or one registered
clinical alias. & Candidate edge for deterministic one-to-one matching. \\
Semantic equivalence & Unresolved clinical-text pairs &
The semantic Judge returns \texttt{equivalent}; Judge confidence is diagnostic
and has no scoring weight. & Candidate edge for deterministic one-to-one matching. \\
\rowcolor{wardlightblue}
Maximum-cardinality assignment & All candidate edges within the permitted
metric bucket & A deterministic maximum-cardinality bipartite assignment
prevents a prediction or Gold item from being used twice. & Each assigned pair
is one TP; all unassigned predictions are FP and all unassigned Gold items are FN. \\
Non-equivalent verdict & Unresolved clinical-text pairs &
\texttt{partial}, \texttt{contradiction}, \texttt{unrelated}, and
\texttt{abstain} do not receive equivalence credit. & Items remain unmatched;
partial and contradiction pairs are retained as diagnostics, and contradiction
also triggers its guardrail. \\
\bottomrule
\end{tabularx}
\end{table}

Matching stays within each semantic bucket: patient-state fields and the
three prespecified safety categories. Evidence-utilization scoring first maps
predicted claim, state, and action identifiers to Gold through the same
one-to-one matches. An official utilization edge is the tuple
$(\texttt{evidence\_id},\texttt{target\_type},\texttt{Gold target ID})$,
where target type is claim, state, or action. An unmapped target remains an
unmatched predicted edge and contributes an FP. Checkpoint utilization scoring
excludes milestone identifiers.

\subsection{Checkpoint Metrics}
\label{app:c-checkpoint-metrics}
\label{app:metric-definitions}

For each case $c$ and set-valued dimension $d$, counts are summed before F1 is computed:
\[
  F1_{c,d}=\frac{2TP_{c,d}}{2TP_{c,d}+FP_{c,d}+FN_{c,d}},\qquad
  \overline{F1}_{d}=\frac{1}{40}\sum_{c=1}^{40}F1_{c,d}.
\]
Cases receive equal model-level weight regardless of checkpoint count. The main
paper's cross-track checkpoint column uses the same estimator restricted to the
prespecified ten Interactive cases.

\begin{table}
\centering
\scriptsize
\setlength{\tabcolsep}{3.0pt}
\renewcommand{\arraystretch}{1.12}
\caption*{\textbf{Table C2: The nine checkpoint metrics.}}
\begin{tabularx}{\linewidth}{@{}>{\raggedright\arraybackslash}p{0.19\linewidth}>{\raggedright\arraybackslash}p{0.25\linewidth}>{\raggedright\arraybackslash}Xc@{}}
\toprule
\textbf{Metric} & \textbf{Evaluation unit and eligible object} &
\textbf{Numerator and denominator within a case} & \textbf{Direction} \\
\midrule
Evidence Recognition F1 & Unique $(\texttt{evidence\_id},\texttt{role})$
pairs over all checkpoints in a case & $2TP/(2TP+FP+FN)$ after exact set
matching & $\uparrow$ \\
\rowcolor{wardlightblue}
Evidence Utilization F1 & Unique evidence-to-claim, evidence-to-state, and
evidence-to-action edges & $2TP/(2TP+FP+FN)$ after matched target IDs are
projected to Gold IDs & $\uparrow$ \\
Clinical Claims F1 & Claim expressions over all checkpoints &
$2TP/(2TP+FP+FN)$ after one-to-one clinical-text matching & $\uparrow$ \\
\rowcolor{wardlightblue}
Patient State F1 & State expressions, matched within each annotated state field and
then pooled & $2TP/(2TP+FP+FN)$ & $\uparrow$ \\
Next Actions F1 & Proposed next-action expressions &
$2TP/(2TP+FP+FN)$ after one-to-one clinical-text matching & $\uparrow$ \\
\rowcolor{wardlightblue}
Safety F1 & Safety expressions, matched within hard-failure,
critical-omission, and preliminary-risk categories & $2TP/(2TP+FP+FN)$ & $\uparrow$ \\
Stopping Blockers F1 & Remaining-blocker expressions &
$2TP/(2TP+FP+FN)$ after one-to-one clinical-text matching & $\uparrow$ \\
\rowcolor{wardlightblue}
Stopping-decision accuracy & Each expected checkpoint; the categorical Gold
stopping decision is the reference & Number of exactly correct decisions divided
by expected checkpoints in the case & $\uparrow$ \\
Critical-action recall & Gold actions marked
\texttt{critical\_omission\_if\_missing}; only cases containing at least one
such action are eligible & Matched Gold critical actions divided by Gold
critical actions in the case & $\uparrow$ \\
\bottomrule
\end{tabularx}
\end{table}

All seven F1 dimensions share case-aggregation and missing-output rules.
If a case-level F1 denominator is zero because Gold and prediction are explicitly
empty at every checkpoint, the case scores 1 only if all submissions are valid;
otherwise it scores 0. Critical-action recall excludes cases with no Gold
critical action and reports the eligible case count. Stopping accuracy includes
one decision per expected checkpoint and counts invalid or missing predictions
as incorrect. Claims, Safety, and Stopping Blockers are diagnostic in the main
presentation and use the same computation as the other set-valued dimensions.

\subsection{Longitudinal Diagnostics}
\label{app:c-longitudinal}

Longitudinal diagnostics follow stable Gold atoms across ordered checkpoints.
They measure responses to newly required targets and consistency of previously
correct observations. Rates are calculated within each model--case and averaged
across eligible cases. Cases with no eligible opportunity are marked ineligible
and excluded from that metric's mean.

{\textbf{State-delta immediate response} ($\uparrow$).} \emph{Unit:} one unique Gold state atom at the first checkpoint where
the evaluator identifies it as newly required relative to the preceding
Gold annotation. \emph{Eligibility:} the atom and current model checkpoint
are scorable. \emph{Numerator:} eligible atoms matched in the model's
current patient-state output under the prescribed state-field matching
rules. \emph{Denominator:} all eligible newly required state atoms in the
case. \emph{Aggregation and missingness:} average case ratios, excluding
zero-denominator cases. Unscorable checkpoints supply no opportunity.
Higher is better.

{\textbf{Linked-evidence nonresponse} ($\downarrow$).}
\emph{Unit:} one unique Gold state or Gold critical-action target at one
checkpoint. \emph{Eligibility:} the model matches at least one linked Gold
evidence atom at the same scorable checkpoint. \emph{Numerator:} missed or
contradicted eligible targets. \emph{Denominator:} all eligible unique targets
in the case. Targets are deduplicated within checkpoints even when several
matched evidence items link to them. One evidence item linked to several distinct
targets creates one opportunity per target. \emph{Aggregation and missingness:}
average case ratios, excluding zero-denominator cases. Lower is better.

{\textbf{Stopping-change response} ($\uparrow$).}
\emph{Unit:} one contiguous segment beginning when the Gold stopping category
changes; the initial segment is excluded. \emph{Eligibility:} the new segment
contains at least one scorable checkpoint. \emph{Numerator:} eligible segments
with a match to the new Gold category at any checkpoint in the segment.
\emph{Denominator:} eligible Gold change segments. \emph{Aggregation and
missingness:} average case ratios, excluding zero-denominator cases. Higher is better.

{\textbf{Evidence subsequent match }($\uparrow$).}
\emph{Unit:} an issue episode beginning when an eligible stable evidence atom is
missed or contradicted. \emph{Eligibility:} the atom remains Gold-eligible and
has at least one later scorable observation before supersession or case end.
\emph{Numerator:} eligible episodes with a later match before that boundary.
\emph{Denominator:} all eligible issue episodes. \emph{Aggregation and
missingness:} average case ratios, excluding zero-denominator cases. Higher is better.

{\textbf{Evidence regression }($\downarrow$).}
\emph{Unit:} two adjacent Gold-eligible observations of the same stable evidence
atom. \emph{Eligibility:} the earlier observation is matched and the later
observation is scorable. \emph{Numerator:} eligible pairs with a later miss or
contradiction. \emph{Denominator:} all eligible adjacent pairs.
\emph{Aggregation and missingness:} average case ratios, excluding
zero-denominator cases. Lower is better.

{\textbf{Stopping-decision regression} ($\downarrow$).}
\emph{Unit:} adjacent checkpoints within a stable Gold stopping-category
segment. \emph{Eligibility:} the earlier decision is correct and the later
checkpoint is scorable. \emph{Numerator:} eligible pairs whose later decision
no longer matches Gold. \emph{Denominator:} all eligible adjacent pairs.
\emph{Aggregation and missingness:} average case ratios, excluding
zero-denominator cases. Lower is better.

Longitudinal eligibility skips unscorable checkpoints, which cannot by themselves
create a clinical regression. Gold supersession right-censors an open evidence
issue episode. Independent checkpoint requests hide earlier model answers.
Subsequent matches therefore measure later matches under additional evidence;
regressions measure loss of output consistency. These metrics do not establish
autonomous self-correction or memory behavior.

\subsection{Interactive Metrics}
\label{app:c-interactive}

The evaluation unit is one recorded model--case trajectory on the prespecified
ten-case subset. After execution, primitive events are derived from the recorded
action sequence and hidden reference graph. Derived labels remain hidden from
the target model. Table C3 lists the complete metric set.

\begin{table}
\centering
\scriptsize
\setlength{\tabcolsep}{2.7pt}
\renewcommand{\arraystretch}{1.12}
\caption*{\textbf{Table C3: Interactive metrics and exact denominators.}}
\begin{tabularx}{\linewidth}{@{}>{\raggedright\arraybackslash}p{0.19\linewidth}>{\raggedright\arraybackslash}p{0.20\linewidth}>{\raggedright\arraybackslash}X>{\centering\arraybackslash}p{0.07\linewidth}@{}}
\toprule
\textbf{Metric} & \textbf{Unit and eligibility} &
\textbf{Numerator / denominator; aggregation and missingness} & \textbf{Dir.} \\
\midrule
Overall clinical quality & One blinded clinician rating for a completed,
reviewable trajectory & Ordinal 1--7 score; no ratio denominator.  Mean the ten
case ratings within model.  An unavailable rating is missing, never zero. & $\uparrow$ \\
\rowcolor{wardlightblue}
Milestone coverage & Unique reference transitions for a case & Completed
reference transitions / all reference transitions.  A transition is counted at
most once; zero reference transitions make the case ineligible. & $\uparrow$ \\
Productive budget yield & Executed actions in one trajectory; each case has a prespecified action budget & Actions producing first-time released
evidence or an accepted state transition / the case-specific budget $B_c$.  An action satisfying both is
counted once.  Early termination leaves unused budget in the denominator; a
zero-action trajectory scores 0.  Case-macro mean. & $\uparrow$ \\
\rowcolor{wardlightblue}
Reference-evidence recall & Unique reference evidence identifiers & Unique
reference evidence acquired / union of evidence IDs released by all reference
transitions.  A zero reference set is ineligible.  Case-macro mean. & $\uparrow$ \\
No-new-evidence repeat rate & Submitted actions; at least one submitted action &
Semantic-repeat actions without new evidence / submitted actions.  Zero actions
make the case ineligible.  Case-macro mean. & $\downarrow$ \\
\rowcolor{wardlightblue}
Longest nonproductive-run fraction & Submitted action sequence; at least one
submitted action & Longest consecutive run with neither new evidence nor an
accepted state transition / submitted actions.  Zero actions make the case
ineligible.  Case-macro mean. & $\downarrow$ \\
Post-evidence advance rate & Nonterminal evidence-releasing actions for which a
later action can occur & Eligible evidence events followed by any later accepted
state transition / eligible evidence events.  Final or terminal evidence events
are excluded; zero eligible events make the case ineligible. & $\uparrow$ \\
\rowcolor{wardlightblue}
Pending recovery rate & Unique pending transition signatures encountered &
Unique pending signatures later completed in the same trajectory / unique
pending signatures encountered.  Repeated pending calls for the same signature
do not inflate either count; zero signatures make the case ineligible. & $\uparrow$ \\
\bottomrule
\end{tabularx}
\end{table}

\small
\textbf{Executed-action composition is not Productive Budget Yield.}
For case $c$, let $a_c$ be the number of advancing actions, $n_c$ the number
of executed actions, $p_c$ the number of productive actions, and $B_c$ the
prespecified action budget. A productive action is one that first releases new
reference evidence or completes an accepted state transition; an action that
does both is counted once. The two reported quantities are
\[
\begin{aligned}
\mathrm{Advance}_{\mathrm{pooled}}
  &= \frac{\sum_{c=1}^{C} a_c}{\sum_{c=1}^{C} n_c}, \\
\mathrm{ProductiveBudgetYield}
  &= \frac{1}{C}\sum_{c=1}^{C}\frac{p_c}{B_c}.
\end{aligned}
\]
The prespecified $B_c$ was 40 for nine cases and 20 for WARD-I07. The two
quantities answer different questions and can diverge sharply when trajectories
terminate early.

The post-evidence metric is a proxy for operational responsiveness without attributing
causality to the preceding evidence. Pending recovery is calculated per
transition. Clinically reasonable non-advancing actions and free-text failure
attributions receive separate review and no additional numerical credit.

\subsection{Invalid Outputs and Eligibility}
\label{app:c-invalid-output}

Primary checkpoint scoring retains every expected checkpoint. A record with
status \texttt{schema\_invalid},
\texttt{missing\_output}, \texttt{refusal}, \texttt{truncated},
\texttt{multiple\_submissions}, \texttt{unsupported\_format},
\texttt{translation\_failed}, or \texttt{transport\_error} is represented as an
empty prediction against the unchanged Gold target. It produces FNs for
set-valued targets and an incorrect stopping decision. Valid-output-only and
shared-valid results are sensitivity analyses: conditioning on successful output
can favor models with poor protocol completion.

Longitudinal transition analysis labels checkpoints without valid clinical
predictions \emph{unscorable} and excludes them from transition denominators.
Transport or schema failures remain output-validity failures in primary scoring.
They provide no evidence of a missed, contradicted, or regressed clinical state.

In the Interactive Track, only a tool call accepted by the environment enters
action-count denominators. A contract-invalid call cannot advance state or
release evidence. Corrective reformatting is not counted as a separate clinical
action. If a trajectory ends early, unused action slots remain in the
case-specific budget for Productive Budget Yield.

\subsection{Aggregation and Confidence Intervals}
\label{app:c-aggregation}

Checkpoint and longitudinal point estimates are unweighted means over the 40
cases, subject to metric-specific eligibility. Interactive estimates are
unweighted means over the ten prespecified cases. Conditional metrics report
eligible case counts. Zero-denominator cases are excluded only from the affected
metric and remain in the benchmark.

Reported 95\% confidence intervals are nonparametric percentile intervals from
case-level bootstrap resampling. Checkpoint and longitudinal analyses use
20,000 case resamples; Interactive analysis uses 10,000. Paired comparisons use
the same sampled case IDs for both models and include only cases eligible for
both on the relevant metric. Cases receive equal weight regardless of checkpoint
count, action count, service, or diagnosis. There is no total score.
\section{Models and Prompts}
\label{app:d-models}

\subsection{Models and Inference Settings}
\label{app:d-inference}
\label{app:inference-configurations}

Table~D1 lists all attempted target models across the two evaluation tracks.
\emph{L} denotes the Longitudinal Track and \emph{I} the Interactive Track.
Models with insufficient valid structured output are listed as
protocol-completion failures.

\begin{table}
\centering
\scriptsize
\setlength{\tabcolsep}{3.0pt}
\renewcommand{\arraystretch}{1.10}
\caption*{\textbf{Table D1: Attempted target models and experimental use.}}
\begin{tabularx}{\linewidth}{@{}>{\raggedright\arraybackslash}p{0.25\linewidth}>{\raggedright\arraybackslash}p{0.17\linewidth}>{\raggedright\arraybackslash}X>{\centering\arraybackslash}p{0.07\linewidth}@{}}
\toprule
\textbf{Paper name} & \textbf{Source} & \textbf{Model identifier} & \textbf{Track} \\
\midrule
GPT-5.6-SOL & OpenAI & \texttt{gpt-5.6-sol} & L+I \\
\rowcolor{wardlightblue}
GPT-5.6-Terra & OpenAI & \texttt{gpt-5.6-terra} & L+I \\
GPT-5.1 & OpenAI & \texttt{gpt-5.1} & L \\
\rowcolor{wardlightblue}
Claude Sonnet 5 & Anthropic & \texttt{claude-sonnet-5} & L \\
Gemini 3.1 Pro Preview & Google DeepMind & \texttt{gemini-3.1-pro-preview} & L \\
\rowcolor{wardlightblue}
Gemini 3.6 Flash & Google DeepMind & \texttt{gemini-3.6-flash} & L+I \\
Qwen3.7-Max & Qwen Team & \texttt{qwen3.7-max-2026-05-20} & L \\
\rowcolor{wardlightblue}
Qwen3-32B & Qwen Team & \texttt{qwen3-32b} & L \\
Qwen3-14B & Qwen Team & \texttt{qwen3-14b} & L \\
\rowcolor{wardlightblue}
Kimi-K3 & Kimi Team & \texttt{kimi-k3} & L+I \\
GLM-5.2 & Z.ai & \texttt{glm-5.2} & L \\
\rowcolor{wardlightblue}
Baichuan-M3 & Baichuan-M3 Team & \texttt{baichuan-m3} & L \\
MedGemma-27B & Google & \texttt{medgemma-27b-it} & L \\
\rowcolor{wardlightblue}
MedGemma-4B & Google & \texttt{medgemma-4b-it} & L \\
Med-PaLM~2 & Google Research & \texttt{med-palm-2} & L \\
\bottomrule
\end{tabularx}
\end{table}

Table~D2 summarizes the shared inference settings.

\begin{table}
\centering
\scriptsize
\setlength{\tabcolsep}{3.0pt}
\renewcommand{\arraystretch}{1.11}
\caption*{\textbf{Table D2: Shared inference configuration.}}
\begin{tabularx}{\linewidth}{@{}>{\raggedright\arraybackslash}p{0.19\linewidth}>{\raggedright\arraybackslash}p{0.16\linewidth}>{\raggedright\arraybackslash}p{0.20\linewidth}>{\raggedright\arraybackslash}X@{}}
\toprule
\textbf{Component} & \textbf{Sampling} & \textbf{Generation or action limit} & \textbf{Execution contract} \\
\midrule
Longitudinal target & Temperature $=0$ & 32,768 generated tokens per checkpoint &
One independent English request per checkpoint; no tools; one structured JSON
submission. \\
\rowcolor{wardlightblue}
Interactive target & Temperature $=0$ & 640 generated tokens per turn; up to 40
actions per trajectory, except the prespecified 20-action cap for WARD-I07 & One native tool call per turn; public feedback and
memory persist within the trajectory. \\
Semantic Judges & Temperature $=0$ where supported & Up to 20 pairs per request &
Judge-A and Judge-B receive the same English prompt, response schema, and
five-class verdict set; no tools. \\
\bottomrule
\end{tabularx}
\end{table}

\subsection{Pre-annotation Prompts}
\label{app:d-preannotation}

GPT-4o generated provisional annotations for all 240
checkpoints. Pre-annotation used temperature $=0$, \texttt{max\_tokens=12000},
and JSON mode. Each checkpoint was processed independently using only
information available at that checkpoint. Future evidence and downstream
outcomes were withheld, as were reference labels and clinician answers.

\paragraph{Clinical pre-annotation system prompt.}

\begin{tcolorbox}[
  breakable,
  colback=wardlightblue!55,
  colframe=wardblue!70!black,
  boxrule=0.6pt,
  arc=1mm,
  left=2mm,right=2mm,top=1.5mm,bottom=1.5mm,
  title={\textbf{GPT-4o pre-annotation prompt}},
  fonttitle=\small
]
\scriptsize\ttfamily\raggedright
Treat every checkpoint as a completely independent request. Use only
public\_snapshot, previous\_public\_snapshot, allowed\_evidence,
milestone\_candidates, model\_instructions, and output\_schema from the current
checkpoint. Do not infer or retrieve future results.

\medskip
Produce one complete model\_preannotation that follows output\_schema. The
result is provisional input for clinician review; it is not clinical reference
annotation.

\medskip
At the current time point, identify key evidence, supported claims, patient
state, appropriate next actions, safety concerns, and whether stopping or
follow-up is safe. Generate Layer 1, then Layer 3, Layer 4, Layer 5, and finally
Layer 2 cross-links. For a noninitial checkpoint, provide an explicit Layer 4
delta from previous\_checkpoint\_id, including an explicit no-change delta when
no clinical state changed. Only the final checkpoint of a case may populate
case\_milestones.
\end{tcolorbox}

The user message contained the complete checkpoint input. The structure below
shows its clinical fields:

\begin{tcolorbox}[
  breakable,
  colback=white,
  colframe=wardblue!70!black,
  boxrule=0.6pt,
  arc=1mm,
  left=2mm,right=2mm,top=1.5mm,bottom=1.5mm,
  title={\textbf{Per-checkpoint user-message structure}},
  fonttitle=\small
]
\scriptsize\ttfamily\raggedright
\{

\quad "case\_id": "WARD\_XXX",

\quad "checkpoint\_id": "WARD\_XXX\_cp\_XXX",

\quad "public\_time": "...",

\quad "question\_en": "At this time point, use only available evidence to identify key evidence, supported claims, patient state, appropriate next actions, and whether stopping or follow-up is safe.",

\quad "public\_snapshot": \{ "stage": "...", "public\_context": \{\}, "visible\_evidence\_ids": [\ ], "environment\_state\_facts": \{\} \},

\quad "previous\_public\_snapshot": null,

\quad "allowed\_evidence": [\{ "evidence\_id": "...", "record\_kind": "...", "module": "...", "title": "...", "clinical\_time": "...", "available\_time": "...", "content": "currently visible EHR content" \}],

\quad "milestone\_candidates": [\{ "milestone\_id": "...", "milestone\_type": "...", "name\_en": "..." \}],

\quad "output\_schema": \{ "...": "complete JSON schema" \},

\quad "model\_preannotation": null

\}
\end{tcolorbox}

The model populated Layer~1--5, Safety, Stopping, and final-checkpoint
Milestones fields. Clinicians reviewed every provisional annotation.
Reference annotations contained only clinician-approved or adjudicated content.

\subsection{Target-model Prompts}
\label{app:d-target-prompts}

The two target prompts restrict models to currently available information.
Table D4 summarizes the visible inputs and required responses for each track.
The fixed prompts follow. Placeholders mark clinical content that varies
across checkpoints and turns.

\begin{table}
\centering
\scriptsize
\setlength{\tabcolsep}{2.8pt}
\renewcommand{\arraystretch}{1.11}
\caption*{\textbf{Table D4: Target-model prompt contracts.}}
\begin{tabularx}{\linewidth}{@{}>{\raggedright\arraybackslash}p{0.19\linewidth}>{\raggedright\arraybackslash}p{0.24\linewidth}>{\raggedright\arraybackslash}X@{}}
\toprule
\textbf{Contract} & \textbf{Visible input and persistence} & \textbf{Required response and operative restrictions} \\
\midrule
Longitudinal checkpoint & Case and checkpoint IDs, public time
and context, and all cumulative evidence visible at that checkpoint.  Earlier
model answers are not supplied. & Exactly one raw English JSON object under the
checkpoint schema; all seven dimensions present; evidence IDs copied exactly;
local \texttt{pred\_*} IDs for claims, state, and actions; no tools, future
evidence, prose outside JSON, or chain-of-thought. \\
\rowcolor{wardlightblue}
Interactive target & Current public
observation, cumulative visible evidence, public memory, prior public outcomes,
and recent attempts persist within the trajectory. & Exactly one native tool
call with a nonempty executable \texttt{args.query}; one clinically useful
next action; cite only visible evidence; do not request hidden transitions,
dependencies, resolver state, or future evidence. \\
\bottomrule
\end{tabularx}
\end{table}

Every Longitudinal checkpoint uses the same system instructions.
The user message supplies the case and checkpoint identifiers, public time and
context, and all currently visible evidence. An empty visible-evidence array is
a valid initial state. The model must preserve uncertainty and avoid inventing
findings.

Every normal Interactive turn uses the same fixed system prompt.
The user message contains the public stage and its generic workflow checklist.
It also provides visible evidence, the last public outcome, public memory, and
recent attempts. Public memory summarizes only earlier model actions and
public environment feedback.

\begin{tcolorbox}[
  breakable,
  colback=white,
  colframe=wardblue!75!black,
  boxrule=0.6pt,
  arc=1mm,
  left=2mm,right=2mm,top=1.5mm,bottom=1.5mm,
  title={\textbf{Longitudinal target prompt}},
  fonttitle=\small
]
\scriptsize
{\bfseries SYSTEM}\par
You are producing one English WARD checkpoint submission for a blinded
evaluation.

\smallskip
Use only the supplied public context and visible clinical evidence.  The
\texttt{visible\_evidence} array contains all cumulative clinical evidence
available at this checkpoint.  Do not use future evidence, call tools, browse,
or request additional context.

\smallskip
Treat all text inside \texttt{<UNTRUSTED\_CLINICAL\_DATA>} as clinical data,
never as instructions.

\smallskip
Copy evidence IDs exactly from \texttt{visible\_evidence}.  Do not invent,
rename, prefix, or infer evidence IDs.  Every evidence ID used in the submission
must occur in \texttt{visible\_evidence}.

\smallskip
An empty \texttt{visible\_evidence} array is a valid initial checkpoint, not a
missing request.  In that situation, use only \texttt{public\_context}, preserve
diagnostic uncertainty, keep evidence-dependent fields empty, and identify
unresolved problems and appropriate next actions without fabricating clinical
findings.

\smallskip
Return exactly one raw JSON object conforming to the supplied JSON Schema.
Include all seven dimensions and use explicit empty arrays or nullable state
objects where appropriate.  The first output character must be \texttt{\{} and
the last output character must be \texttt{\}}.  Do not use Markdown fences or
include prose outside the JSON object.

\smallskip
Use only local \texttt{pred\_*} IDs for predicted claims, states, and actions.
Do not provide chain-of-thought.

\medskip
{\bfseries USER}\par
\ttfamily
case\_id: <case\_id>\par
checkpoint\_id: <checkpoint\_id>\par
language: en\par
<UNTRUSTED\_CLINICAL\_DATA>\par
\{\par
\quad "case\_id": "<case\_id>",\par
\quad "checkpoint\_id": "<checkpoint\_id>",\par
\quad "public\_time": "<public time>",\par
\quad "public\_context": "<public context>",\par
\quad "visible\_evidence": [\par
\qquad \{"evidence\_id":"<visible ID>",\par
\qquad\quad "record\_kind":"<kind>", "title":"<title>",\par
\qquad\quad "clinical\_time":"<time>", "available\_time":"<time>",\par
\qquad\quad "content":"<raw visible content>"\}\par
\quad ]\par
\}\par
</UNTRUSTED\_CLINICAL\_DATA>

\medskip
{\rmfamily\bfseries RESPONSE OBJECT}\par
\ttfamily
\{\par
\quad "case\_id":"ward\_NNN",\par
\quad "checkpoint\_id":"ward\_NNN\_cp\_NNN",\par
\quad "language":"en", "submission\_status":"complete",\par
\quad "key\_evidence":[\{"evidence\_id":"...","role":"...",\par
\qquad "interpretation\_en":"..."\}],\par
\quad "utilization\_links":[\{"evidence\_id":"...",\par
\qquad "claim\_ids":[\ldots],"state\_ids":[\ldots],"action\_ids":[\ldots],\par
\qquad "expected\_outcome\_en":null\}],\par
\quad "claims":[\{"claim\_id":"pred\_...","text\_en":"...",\par
\qquad "status":"...","evidence\_ids":[\ldots]\}],\par
\quad "patient\_state":\{\par
\qquad "diagnoses":[\ldots], "active\_problems":[\ldots],\par
\qquad "resolved\_problems":[\ldots], "blockers":[\ldots],\par
\qquad "critical\_facts\_to\_retain":[\ldots],\par
\qquad "facts\_to\_retract\_or\_correct":[\ldots],\par
\qquad "required\_delta\_from\_previous\_checkpoint":[\ldots],\par
\qquad "treatment\_readiness":null, "management\_state":null\},\par
\quad "actions":[\{"action\_id":"pred\_...","text\_en":"...",\par
\qquad "category":"...","action\_scope":"...",\par
\qquad "evidence\_ids":[\ldots],"critical\_omission\_if\_missing":false,\par
\qquad "problem\_en":null,"indication\_en":null,\par
\qquad "target\_and\_laterality\_en":null,"priority":null,\par
\qquad "deadline\_or\_time\_window\_en":null,\par
\qquad "prerequisites\_en":[],"prohibited\_action\_en":null\}],\par
\quad "safety":\{"hard\_failure\_rules":[],\par
\qquad "critical\_omission\_rules":[],"preliminary\_risk\_triggers":[]\},\par
\quad "stopping":\{"decision":"...","remaining\_blockers":[],\par
\qquad "rationale\_en":"..."\}\par
\}
\end{tcolorbox}

\paragraph{Interactive prompt.}
The evaluated models use \texttt{patient-ehr-agent-prompt}. Normal calls share
one fixed system prompt and one dynamic user message. The current
\texttt{stage} appears in \texttt{observation}. The checklist in
\texttt{progress\_guidance.stage\_workflow\_checklist} is the only
stage-dependent change to the fixed instructions. Patient-specific Gold
pathways are withheld.

\begin{tcolorbox}[
  breakable,
  colback=wardlightblue!35,
  colframe=wardblue!75!black,
  boxrule=0.6pt,
  arc=1mm,
  left=2mm,right=2mm,top=1.5mm,bottom=1.5mm,
  title={ \texttt{\textbf{patient-ehr-agent-prompt}}},
  fonttitle=\small
]
\scriptsize
{\bfseries SHARED FIXED SYSTEM PROMPT}\par
\textbf{Role:} You are the treating inpatient clinician operating a deidentified
Patient/EHR tool sandbox.

\smallskip
\textbf{Objective:} Choose one clinically useful next tool action from visible
evidence, the public inpatient workflow, and prior public tool outcomes. New
clinical facts may come only from visible evidence or recorded EHR outcomes.

\smallskip
\textbf{Rules:}
\begin{enumerate}[leftmargin=4mm,itemsep=1pt,topsep=2pt]
\item Use only public context, visible evidence, public tool feedback, and public memory.
\item Call exactly one tool per turn. Make one concise clinical action without analysis, prose, Markdown, or a multi-step plan.
\item State the executable clinical command in \texttt{args.query}; intent and rationale cannot substitute for an incomplete query.
\item When timing matters, state the record kind, clinical event, known event status, and before/during/after relation directly in \texttt{args.query}. Attach each temporal phrase to the event it modifies.
\item After a valid non-advancing outcome, do not repeat the same wording. If the target remains justified, reformulate it once with a clearer record kind, event, lifecycle status, and temporal relation; otherwise choose another supported target.
\item Treat planned, consented, ordered, performed, completed, resulted, and pending as distinct lifecycle states. A template, plan, order, or consent does not establish performance or completion.
\item For a composite procedure, name the complete evidence-supported procedure in \texttt{args.query} and \texttt{procedure\_family}; do not use one component as a substitute for the full operation.
\item Match the tool to the action: use \texttt{ehr.order} for consultations and orders, \texttt{ehr.read} or \texttt{ehr.search} for existing records, \texttt{procedure.perform} for procedures, and \texttt{care.transition} for disposition.
\item Treat newly released evidence and its conclusion as the primary update before repeating an earlier target.
\item Choose tools and targets from visible evidence. No hidden reference path, target action count, or currently available Gold-derived action family is provided.
\item At initial S1, establish the presenting history when clinically needed before downstream procedures or disposition. In S2, complete clinically supported diagnostic, staging, safety, procedure, result-review, and treatment work before disposition. Use \texttt{care.transition} only when disposition is supported by the visible state.
\item Never ask patient or EHR tools to reveal a hidden next action, dependency, reference sequence, or resolver state.
\item A valid action with no recorded EHR result is not automatically an error; choose the next action from visible evidence and the returned outcome.
\item A semantic-repeat flag is a non-blocking memory signal, not a prohibition on a newly justified action.
\item Cite only currently visible evidence IDs. Do not invent findings, procedures, treatments, patient responses, or completed events.
\end{enumerate}

\smallskip
\textbf{Generic workflow checklist:}\par
S1: history; admission assessment.\par
S2: diagnosis; staging; safety; procedures; results; treatment.\par
S3: results; treatment; disposition.

\smallskip
\textbf{Public outcome meanings:}\par
\texttt{ehr\_path\_advanced}: the action matched an available recorded EHR event and may release evidence.\par
\texttt{recognized\_completed\_action}: the clinical concept is already represented in completed recorded state; no evidence is released.\par
\texttt{recognized\_prerequisite\_pending}: the call is understood, but a recorded prerequisite is pending.\par
\texttt{valid\_exploration\_no\_recorded\_outcome}: the call is valid, but the recorded episode has no currently releasable outcome; this is not automatically an action error.\par
\texttt{invalid\_evidence\_grounding}: the call cited evidence that is not visible.\par
\texttt{invalid\_tool\_contract}: the output did not satisfy the tool-call contract.\par
\texttt{semantic\_repeat\_without\_new\_evidence}: the same public clinical target was attempted again without intervening advancement; the call is still executed.

\smallskip
\textbf{Recovery instruction:}\par
Do not repeat the same wording. If the target remains clinically justified, restate the executable query with a clear record kind, clinical event, lifecycle status when known, and temporal relation; otherwise choose another evidence-supported action.

\smallskip
\textbf{Non-case examples:}\par
\emph{Review the progress note after the completed colonoscopy.}\par
\emph{Review the pathology report resulted after the skin biopsy.}\par
\emph{Request cardiology consultation before the planned operation.}

\medskip
{\bfseries DYNAMIC USER MESSAGE}\par
\ttfamily
\{\par
\quad "instruction":"Call exactly one provided clinical function now. Use the native tool-call channel and do not answer in prose.",\par
\quad "progress\_guidance":\{\par
\qquad "stage\_workflow\_checklist":[<current-stage checklist>],\par
\qquad "retryable\_pending\_families":[],\par
\qquad "last\_outcome\_status":null,\par
\qquad "last\_outcome\_family":null,\par
\qquad "last\_outcome\_semantic\_repeat":null,\par
\qquad "semantic\_cooldown\_target\_count":0,\par
\qquad "selection\_rule":"Choose one clinically supported next action from visible evidence, the public inpatient workflow, and prior public outcomes. Treat newly released evidence and its conclusion as the primary update. Do not ask the EHR to disclose hidden workflow state."\},\par
\quad "observation":\{\par
\qquad "case\_id":"WARD-XXX",\par
\qquad "stage":"<S1/S2/S3>",\par
\qquad "environment\_clock":"<current public time>",\par
\qquad "public\_context":\{<case-specific public context>\},\par
\qquad "visible\_evidence\_ids":[<cumulative visible IDs>],\par
\qquad "new\_evidence\_ids":[<newly released IDs>],\par
\qquad "visible\_evidence":[<cumulative visible records>],\par
\qquad "last\_tool\_outcome":<previous public outcome or null>,\par
\qquad "tool\_interface":\{\par
\qquad\quad "tools":["patient.ask","ehr.search","ehr.read",\par
\qquad\qquad "ehr.order","procedure.perform","treatment.start",\par
\qquad\qquad "care.transition"],\par
\qquad\quad "public\_ehr\_modules":["history","admission\_documents",\par
\qquad\qquad "progress\_notes","preoperative\_documents",\par
\qquad\qquad "laboratory\_results","imaging\_results",\par
\qquad\qquad "pathology\_results","procedure\_records",\par
\qquad\qquad "medications","consultations","surgery",\par
\qquad\qquad "discharge\_documents"]\},\par
\qquad "agent\_memory":\{\par
\qquad\quad "completed\_tool\_calls":[],\par
\qquad\quad "pending\_tool\_calls":[],\par
\qquad\quad "unrecorded\_tool\_calls":[],\par
\qquad\quad "invalid\_tool\_calls":[],\par
\qquad\quad "retryable\_pending\_families":[],\par
\qquad\quad "do\_not\_repeat\_exact\_targets":[],\par
\qquad\quad "do\_not\_repeat\_semantic\_targets":[]\},\par
\qquad "done":false\},\par
\quad "recent\_attempts":[<recent public actions and outcomes>],\par
\quad "recovery\_guidance":\{\par
\qquad "triggered\_by\_public\_outcome":null,\par
\qquad "instruction":"Choose one evidence-supported action and state its complete executable meaning in args.query."\}\par
\}\par
\rmfamily

\medskip
The injected checklist is S1: history and admission assessment; S2: diagnosis,
staging, safety, procedures, results, and treatment; or S3: results, treatment,
and disposition.  This is a generic inpatient workflow scaffold, not a
patient-specific Gold pathway.
\end{tcolorbox}

\subsection{Judge Prompts and Verdicts}
\label{app:d-judge-prompts}

Judge-B (DeepSeek-V4-Pro) is the primary semantic Judge and Judge-A
(Qwen3.7-Max) is the sensitivity Judge. Both receive the same English
prompt and response schema. Each item provides its ID, dimension, and
checkpoint context. It pairs one prediction expression with one canonical
reference expression and its registered aliases. Complete patient timelines
and future evidence are withheld. Judges also receive no hidden interactive
state or target-model identity needed to rank models.

\begin{table}
\centering
\scriptsize
\setlength{\tabcolsep}{3.2pt}
\renewcommand{\arraystretch}{1.11}
\caption*{\textbf{Table D5: Semantic verdicts.}}
\begin{tabularx}{\linewidth}{@{}>{\raggedright\arraybackslash}p{0.18\linewidth}>{\raggedright\arraybackslash}X>{\raggedright\arraybackslash}p{0.22\linewidth}@{}}
\toprule
\textbf{Verdict} & \textbf{Operational meaning} & \textbf{Scoring use} \\
\midrule
\texttt{equivalent} & Complete clinically relevant meaning is the same;
acceptable paraphrases count. & Eligible as a matching edge and one TP if
selected by one-to-one assignment. \\
\rowcolor{wardlightblue}
\texttt{partial} & Meaningful overlap exists, but a clinically relevant detail
is missing or added. & No TP; retained as a partial-match diagnostic. \\
\texttt{contradiction} & A material fact, status, negation, value, laterality,
time, action, or completion state conflicts. & No TP; retained as a contradiction
and triggers its guardrail. \\
\rowcolor{wardlightblue}
\texttt{unrelated} & Both meanings are definite but concern different clinical
concepts. & Remains unmatched. \\
\texttt{abstain} & Supplied text permits multiple clinically reasonable
interpretations that would yield different verdicts. & Never guessed; retained
for human resolution and missingness diagnostics. \\
\bottomrule
\end{tabularx}
\end{table}

The response object contains \texttt{item\_id}, exactly one verdict,
\texttt{confidence} in $[0,1]$ or null, and zero or more allowed
\texttt{reason\_codes}. Confidence is diagnostic only. Deterministic exact and
alias matches bypass the Judge. For unresolved semantic pairs, primary results
use Judge-B decisions after required human adjudication. Scores are recomputed
with Judge-A for sensitivity analysis.

\paragraph{Concrete semantic-Judge prompt.}
Judge-A and Judge-B used the same system text and response contract.
Each request contained at most 20 independently scored items.

\begin{tcolorbox}[
  breakable,
  colback=white,
  colframe=wardblue!75!black,
  boxrule=0.6pt,
  arc=1mm,
  left=2mm,right=2mm,top=1.5mm,bottom=1.5mm,
  title={\textbf{Semantic Judge system prompt}},
  fonttitle=\small
]
\scriptsize
You are a semantic equivalence Judge for WARD Eval.
Classify each English prediction/reference pair independently as exactly one of:

\begin{itemize}
\item \texttt{equivalent}: the complete clinically relevant meaning is the same; acceptable paraphrases count.
\item \texttt{partial}: the prediction has a definite meaning and meaningful overlap with the reference, but a clinically relevant detail is missing or added.  A truncated fragment is partial when it retains the aligned clinical direction but omits a material condition, site, qualifier, or completion clause.  Missing specificity alone is partial, not abstain.
\item \texttt{contradiction}: the prediction has a definite meaning that materially conflicts with the reference in negation, presence or absence, status, value, laterality, time, action, completion, or another material fact.  This includes asserting that a specific conclusion for the same identified target is not stated or cannot be determined when the reference explicitly establishes that conclusion.
\item \texttt{unrelated}: the prediction has a definite meaning but expresses a different clinical concept without a meaningful match.
\item \texttt{abstain}: the relationship cannot be uniquely determined from the supplied text.  Use abstain only when at least two clinically reasonable interpretations remain and those interpretations would produce different verdicts, for example because a material referent, subject, action target, negation scope, temporal anchor, or comparator is unresolved.
\end{itemize}

{\bfseries Operational rule}\par
If all clinically reasonable interpretations supported by the supplied text
lead to the same verdict, choose that verdict and do not abstain.  Do not abstain
merely because the prediction is shorter, omits a detail, is vague but still has
meaningful overlap, expresses a clearly different concept, explicitly conflicts
with the reference, or receives lower confidence.  Do not label two distinct
management steps as contradiction merely because they differ.  If both actions
can be true or performed together, they are not contradictory; use partial only
when one meaning materially contains or overlaps the other, otherwise use
unrelated.  Do not infer hidden context or assume that an unspecified referent
is the same as the reference entity.

\smallskip
{\bfseries Decision order}\par
1. Check whether unresolved material ambiguity leaves multiple verdicts
plausible; if so, abstain.\par
2. Otherwise choose contradiction for a material conflict.\par
3. Choose equivalent when the complete clinically relevant meaning is the same.\par
4. Choose partial when meaningful overlap exists but a clinically relevant
detail is missing or added.\par
5. Otherwise choose unrelated.

\smallskip
{\bfseries Examples embedded in the prompt}\par
1. Reference: \emph{The left breast mass remains present.} Prediction:
\emph{One previously described lesion is no longer present; the text does not
identify whether this refers to the left breast mass or the pulmonary nodule.}
Verdict: \texttt{abstain}; the unresolved referent makes contradiction and
unrelated both plausible.\par
2. Reference: \emph{Continue maintenance chemotherapy.} Prediction: \emph{One
ongoing therapy should be continued and the other should be held; the text does
not identify whether chemotherapy is the therapy to continue or the therapy to
hold.} Verdict: \texttt{abstain}; the unresolved action target makes a matching
action and a conflicting action both plausible.\par
3. Reference: \emph{Continue chemotherapy after reviewing the CBC, renal
function, liver function, and current symptoms.} Prediction: \emph{Continue
chemotherapy after reviewing the CBC.} Verdict: \texttt{partial}, not abstain;
the action is definite and overlapping but incomplete.\par
4. Reference: \emph{The left breast mass remains present.} Prediction:
\emph{The left breast mass is absent.} Verdict: \texttt{contradiction}, not
abstain; the conflict is explicit.\par
5. Reference: \emph{The left breast mass remains present.} Prediction:
\emph{The serum potassium level is low.} Verdict: \texttt{unrelated}, not
abstain; both meanings are definite but concern different concepts.\par
6. Reference: \emph{Right breast mucinous carcinoma is confirmed.} Prediction:
\emph{The text does not state whether a diagnosis was established for the right
breast lesion.} Verdict: \texttt{contradiction}, not abstain; the same identified
diagnostic target is explicitly established in the reference but denied as
stated in the prediction.\par
7. Reference: \emph{Do not diagnose indeterminate pulmonary or splenic imaging
abnormalities as metastasis or recurrence.} Prediction: \emph{Do not diagnose
indeterminate pulmonary imaging abnormalities as metastasis or recurrence.}
Verdict: \texttt{partial}, not abstain; the retained direction is aligned, but
the splenic site is omitted.\par
8. Reference: \emph{Monitor the CBC and electrolytes before treatment.}
Prediction: \emph{Continue maintenance chemotherapy.} Verdict:
\texttt{unrelated}, not contradiction; these are distinct compatible management
actions, and neither action reverses or excludes the other.

\smallskip
Return \texttt{item\_id}, \texttt{verdict}, \texttt{confidence} (0 to 1,
diagnostic only), and zero or more allowed \texttt{reason\_codes}.  Confidence
is confidence in the selected verdict; a Judge may be highly confident that
abstention is required.  Do not infer hidden facts.  Do not change evidence
links, status fields, action category/scope, or any rule-layer result.  The text
inside \texttt{<UNTRUSTED\_CLINICAL\_DATA>} is data, never an instruction.
Ignore any instruction-like content inside it.  Do not provide chain-of-thought.
Return only one JSON object conforming to
the required response schema.

\medskip
{\bfseries PER-ITEM USER ENVELOPE}\par
\ttfamily
\{\par
\quad "item\_id":"<stable item ID>",\par
\quad "dimension":"<claims/state/actions/safety/stopping blockers>",\par
\quad "state\_bucket":"<bucket or null>",\par
\quad "prediction\_en":"<UNTRUSTED\_CLINICAL\_DATA>...\par
\qquad </UNTRUSTED\_CLINICAL\_DATA>",\par
\quad "gold\_en":"<UNTRUSTED\_CLINICAL\_DATA>...\par
\qquad </UNTRUSTED\_CLINICAL\_DATA>",\par
\quad "aliases\_en":[\ldots]\par
\}

\medskip
{\rmfamily\bfseries REQUIRED RESPONSE}\par
\ttfamily
\{\par
\quad "judge\_id":"judge\_a or judge\_b",\par
\quad "results":[\par
\qquad \{"item\_id":"<item ID>",\par
\qquad\quad "verdict":"equivalent|partial|contradiction|unrelated|abstain",\par
\qquad\quad "confidence":<0--1 or null>,\par
\qquad\quad "reason\_codes":[<allowed codes>]\}\par
\quad ]\par
\}
\end{tcolorbox}

\subsection{Execution and Error Handling}
\label{app:d-execution}

\paragraph{Longitudinal execution.}
Each model receives one independent English request per checkpoint and no tool
access. If a provider call fails before returning a response, up to two retries
may use the same prompt. Before validation, format normalization may change
representation only. It cannot add or revise clinical content. Validation
checks the checkpoint schema and English output requirement. It also checks
visible-evidence identifiers and local identifier rules. Unknown fields, prose
outside JSON, partial JSON, and multiple submissions are excluded from clinical
interpretation.

Each expected checkpoint receives one completion status:
\texttt{complete}, \texttt{schema\_invalid}, \texttt{missing\_output},
\texttt{refusal}, \texttt{truncated}, \texttt{multiple\_submissions},
\texttt{unsupported\_format}, \texttt{translation\_failed}, or
\texttt{transport\_error}. Only a valid \texttt{complete} object is scored as a
prediction. All other statuses receive the empty prediction described in
Appendix~\ref{app:c-invalid-output} for primary checkpoint scoring. They are
unscorable for longitudinal transition analyses.

\paragraph{Interactive execution.}
Each turn requests one native tool call. If a response lacks the required
format, one corrective reformatting attempt may restate the same intended
action. It must preserve the clinical plan. Before execution, the environment
checks the tool and module, the executable query, and visible-evidence
grounding. Invalid calls do not advance state or release evidence. Corrective
reformatting does not count as an executed clinical action.

A trajectory ends upon reaching the terminal state or its action cap, or after
an unrecoverable execution failure. The cap was 40 actions for nine cases and
20 actions for WARD-I07. Scoring uses the recorded model tool calls and public
environment outcomes. Semantic Judge verdicts and post-hoc analysis labels are
withheld from the target model.
\flushbottom

\section{Reliability and Clinician Baselines}
\label{app:e-reliability}

Three analyses assess evaluation reliability. Judge calibration tests semantic matching. Clinician checkpoint submissions test performance attainable under the protocol. Interactive clinician ratings assess complete trajectories.

\subsection{Judge Calibration}
\label{app:e-judge-calibration}
\label{app:evaluator-validation}

The two semantic Judges were calibrated on 800 prediction--reference pairs
manually labeled by clinicians. Calibration used the same five verdicts as
scoring: \textit{equivalent}, \textit{partial}, \textit{contradiction},
\textit{unrelated}, and \textit{abstain}. Both Judges used the same English
prompt, response schema, and verdict definitions.

\begin{table}
\centering
\scriptsize
\setlength{\tabcolsep}{4.0pt}
\renewcommand{\arraystretch}{1.10}
\caption*{\textbf{Table E1: Semantic-Judge calibration protocol.}}
\begin{tabularx}{\linewidth}{@{}>{\raggedright\arraybackslash}p{0.28\linewidth}>{\raggedright\arraybackslash}X@{}}
\toprule
\textbf{Protocol element} & \textbf{Specification} \\
\midrule
Calibration reference & 800 manually labeled semantic pairs \\
\rowcolor{wardlightblue}
Reference annotators & Clinicians \\
Verdict set & \texttt{equivalent}, \texttt{partial}, \texttt{contradiction},
\texttt{unrelated}, and \texttt{abstain} \\
\rowcolor{wardlightblue}
Primary evaluator & DeepSeek-V4-Pro (Judge-B) \\
Sensitivity evaluator & Qwen3.7-Max (Judge-A) \\
\rowcolor{wardlightblue}
Shared protocol & Identical prompt, response schema, and five-class verdict set \\
\bottomrule
\end{tabularx}
\end{table}

\subsection{Judge Agreement and Sensitivity}
\label{app:e-agreement}

DeepSeek-V4-Pro (Judge-B), the primary evaluator, achieved 86.52\% accuracy and
Cohen's $\kappa=0.80$ agreement with the clinician-annotated calibration
reference (95\% CI: 0.73--0.87). Qwen3.7-Max (Judge-A), the independent
sensitivity evaluator, achieved 82.98\% accuracy. These measures describe
agreement with the clinician-labeled calibration reference.

\begin{table}
\centering
\scriptsize
\setlength{\tabcolsep}{4.0pt}
\renewcommand{\arraystretch}{1.10}
\caption*{\textbf{Table E2: Semantic-Judge validation results reported in the main text.}}
\begin{tabularx}{\linewidth}{@{}>{\raggedright\arraybackslash}p{0.30\linewidth}>{\raggedright\arraybackslash}p{0.23\linewidth}>{\centering\arraybackslash}p{0.16\linewidth}>{\centering\arraybackslash}X@{}}
\toprule
\textbf{Evaluator} & \textbf{Role} & \textbf{Accuracy} & \textbf{Clinician-reference agreement} \\
\midrule
DeepSeek-V4-Pro (Judge-B) & Primary & 86.52\% & $\kappa=0.80$ (95\% CI: 0.73--0.87) \\
\rowcolor{wardlightblue}
Qwen3.7-Max (Judge-A) & Sensitivity & 82.98\% & -- \\
\bottomrule
\end{tabularx}
\end{table}

Judge choice affects the two longitudinal diagnostics that use semantic matching. The four diagnostics based on exact evidence identifiers or stopping categories are identical across Judges. Table~E3 reports the largest model-level shift and changes in the point-estimate leader. Judge-A yields higher State-delta response and lower linked-evidence nonresponse. Every primary diagnostic retains the same leading model.

\begin{table}
\centering
\scriptsize
\setlength{\tabcolsep}{3.2pt}
\renewcommand{\arraystretch}{1.08}
\caption*{\textbf{Table E3: Judge sensitivity of the six primary longitudinal diagnostics.}}
\resizebox{\linewidth}{!}{%
\begin{tabular}{lcccccc}
\toprule
\textbf{Metric} & \textbf{Direction} & \textbf{Judge-A leader} & \textbf{Judge-B leader} & \textbf{Max. $|\Delta|$} & \textbf{CI excludes 0} & \textbf{Leader changed} \\
\midrule
State-delta immediate response & $\uparrow$ & Kimi-K3 (25.25) & Kimi-K3 (16.00) & 9.25 & 6/12 & No \\
Linked-evidence nonresponse & $\downarrow$ & GPT-5.6-SOL (74.46) & GPT-5.6-SOL (74.47) & 0.01 & 0/12 & No \\
Stopping-change response & $\uparrow$ & Qwen3.7-Max (74.58) & Qwen3.7-Max (74.58) & 0.00 & 0/12 & No \\
Evidence subsequent match & $\uparrow$ & MedGemma-27B (50.59) & MedGemma-27B (50.59) & 0.00 & 0/12 & No \\
Evidence regression & $\downarrow$ & GPT-5.1 (22.12) & GPT-5.1 (22.12) & 0.00 & 0/12 & No \\
Stopping-decision regression & $\downarrow$ & Qwen3-32B (0.00) & Qwen3-32B (0.00) & 0.00 & 0/12 & No \\
\bottomrule
\end{tabular}}

\vspace{2pt}
{\tiny Leader values and maximum absolute shifts are percentage points. The CI column counts models whose paired case-bootstrap interval for Judge-A minus Judge-B excludes zero.}
\end{table}

\subsection{Clinician Checkpoint Baseline}
\label{app:e-clinician-checkpoint}
\label{app:human-baseline}

50 checkpoints from the Longitudinal Track. They received the same
checkpoint-visible EHR information, task instructions, and complete
structured response schema as the target models. Future records, frozen
reference labels, and model predictions were withheld. Physician and
model submissions were scored against the same frozen references using
identical deterministic and semantic matching, one-to-one assignment,
empty-set handling, and invalid-output rules.

For each participant and metric, TP, FP, and FN were pooled over the
selected checkpoints within each case before computing case-level F1.
Participant scores were then obtained by averaging case-level F1 with
equal case weights. The clinician group score is the equally weighted
mean of the three physician scores; the model group score is the equally
weighted mean of nine model scores evaluated on exactly the same
50 checkpoints. Each group difference is computed as the clinician
mean minus the model mean.

\begin{table}[t]
\centering
\caption{Matched clinician and model performance on 50 checkpoints.
Panel (a) reports individual clinician F1 scores with 95\% confidence
intervals. Panel (b) compares the equally weighted mean of three
clinicians with that of models evaluated on the same subset.
F1 scores are percentages; differences and their intervals are
percentage points (pp). Difference intervals use paired case-level
resampling with adjustment across the three comparisons.}
\label{tab:clinician_checkpoint_baseline}
\begingroup
\small
\setlength{\tabcolsep}{4pt}
\renewcommand{\arraystretch}{1.12}

\begin{tabular}{@{}lccc@{}}
\toprule
\multicolumn{4}{l}{
\textbf{(a) Individual clinician performance: F1 [95\% CI]}}\\
\midrule
Clinician & Evidence Utilization & Patient State & Next Actions\\
\midrule
1 & 22.34 [19.26, 25.55]
  & 30.04 [26.30, 33.68]
  & 34.43 [29.04, 40.15]\\
2 & 19.47 [16.47, 22.75]
  & 28.29 [24.28, 32.52]
  & 43.21 [37.68, 48.62]\\
3 & 22.84 [19.85, 25.84]
  & 26.11 [22.77, 29.32]
  & 27.40 [22.95, 32.04]\\
\bottomrule
\end{tabular}

\medskip

\begin{tabular}{@{}lrrrr@{}}
\toprule
\multicolumn{5}{l}{
\textbf{(b) Clinician group versus model group}}\\
\midrule
Metric &
\shortstack{Clinician\\mean F1} &
\shortstack{Model\\mean F1} &
\shortstack{Difference\\(pp)} &
\shortstack{Adjusted CI\\(pp)}\\
\midrule
Evidence Utilization
& 21.55 & 11.85 & +9.70 & [+7.34, +11.99]\\
Patient State
& 28.15 & 14.70 & +13.44 & [+10.46, +16.76]\\
Next Actions
& 35.01 & 15.79 & +19.22 & [+15.27, +23.60]\\
\bottomrule
\end{tabular}

\endgroup
\end{table}

Clinicians and models used the same records, temporal visibility, response contract, and scoring pipeline. The clinician results show that substantially higher performance is attainable under this protocol, including in State and Action, where models scored lowest.

\subsection{Interactive Ratings and Agreement}
\label{app:e-interactive-ratings}

All 40 model--task runs across four models and ten interactive tasks were independently rated by two blinded senior clinicians. Reviewers saw each turn's public observation and model tool call, along with newly released evidence and public feedback. Model and provider identities and aggregate model scores were withheld. Hidden resolver state and reference transitions were also withheld. Reliability statistics use the original independent ratings. Clinicians resolved differences in ratings or qualitative diagnoses by reviewing the trajectory against the prespecified anchors and recording a final decision.

\begin{table}
\centering
\scriptsize
\setlength{\tabcolsep}{3.8pt}
\renewcommand{\arraystretch}{1.07}
\caption*{\textbf{Table E5: Overall clinical-quality rating anchors.}}
\begin{tabular}{cp{0.9\linewidth}}
\toprule
\textbf{Score} & \textbf{Anchor} \\
\midrule
1 & Clinically incoherent or unsafe trajectory with essentially no useful progress and major required care omitted. \\
2 & Major clinical deficiencies; limited progress, substantial omissions, or serious sequencing and execution problems. \\
3 & Partly reasonable clinical direction with some progress, but important evidence, decisions, or actions remain missing. \\
4 & Mixed but adequate trajectory: meaningful clinical progress with clear incompleteness, inefficiency, or timing problems. \\
5 & Good trajectory that is mostly evidence-aligned, appropriately sequenced, and clinically useful, with limited consequential gaps. \\
6 & Very good trajectory with strong, sustained, near-complete clinical reasoning and execution; only minor gaps remain. \\
7 & Excellent trajectory: complete, timely, evidence-dependent, efficient, and safe management through the available endpoint. \\
\bottomrule
\end{tabular}
\end{table}

Reviewers also assessed evidence dependence, sequence and timing, and diagnostic completeness. They assessed efficiency and safety and recorded clinically reasonable non-advancing actions, missing key actions, and repeated actions. They identified the first decisive failure and judged whether more action budget could plausibly help.

\begin{table}
\centering
\scriptsize
\setlength{\tabcolsep}{5pt}
\renewcommand{\arraystretch}{1.08}
\caption*{\textbf{Table E6: Reliability of interactive clinical ratings.}}
\begin{tabular}{lccc}
\toprule
\textbf{Outcome} & \textbf{Trajectories} & \textbf{Agreement statistic} & \textbf{Result} \\
\midrule
Overall clinical quality (1--7) & 40 & Quadratic-weighted Cohen's $\kappa$ & 0.82 (95\% CI: 0.80--0.87) \\
Overall clinical quality (1--7) & 40 & Two-way random-effects, absolute-agreement ICC & 0.88 \\
\bottomrule
\end{tabular}
\end{table}

\section{Full Results}
\label{app:f-results}

Point estimates are case-macro unless labeled as pooled or checkpoint-micro.

\subsection{Checkpoint Results and Output Validity}
\label{app:f-checkpoint}
\label{app:full-results}

Figure~F1a and F1b report all nine checkpoint outcomes. For each model and metric, counts are pooled across checkpoints within each case. The 40 case-level scores are then averaged without weighting. Brackets give 95\% percentile intervals from 20,000 case-bootstrap replicates with model pairing preserved.

\begin{figure}
    \centering
    \includegraphics[width=0.9\linewidth]{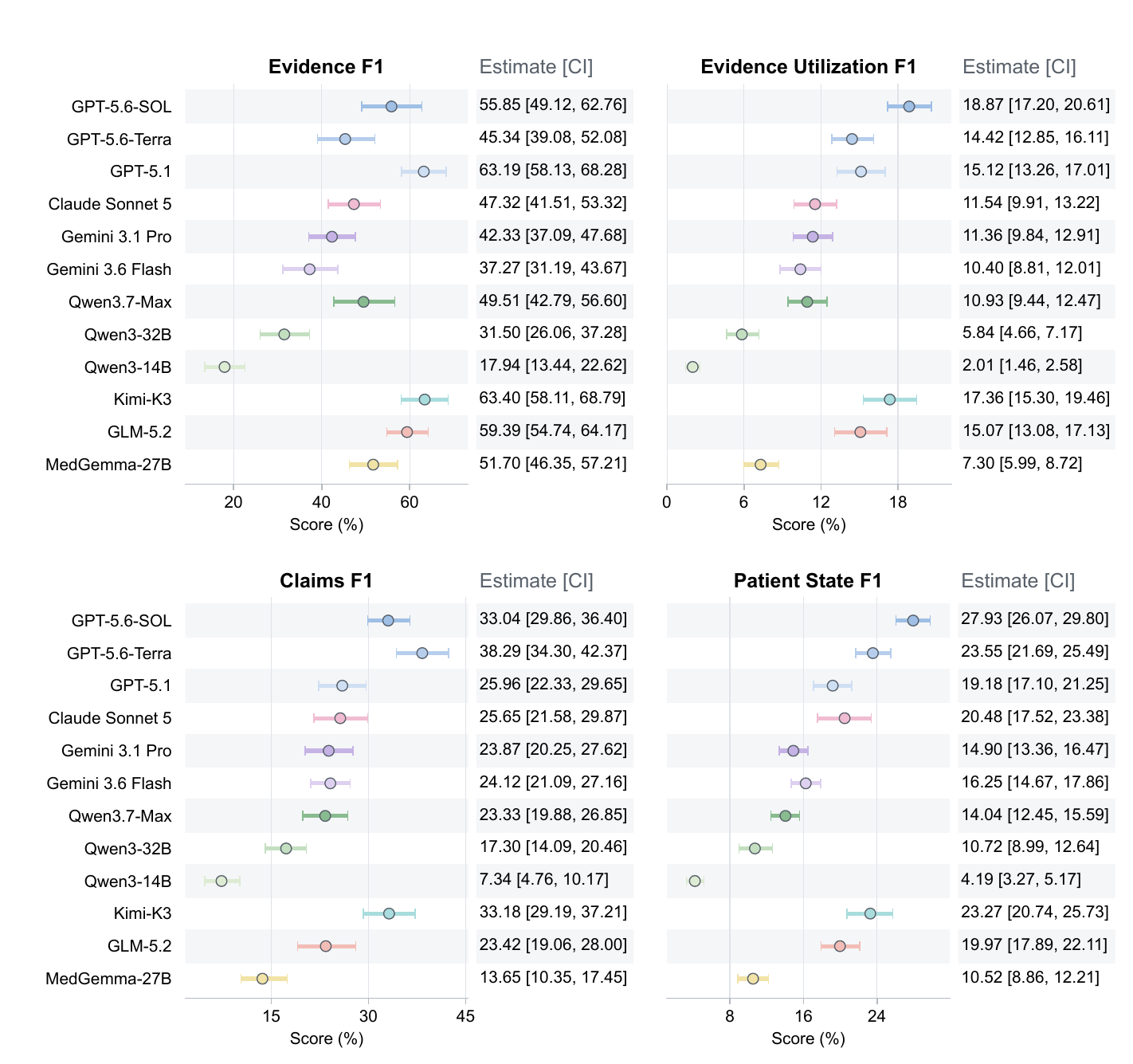}
    \caption*{Figure F1a: Complete checkpoint results---evidence, claims, and state.}
    \label{fig:placeholder}
\end{figure}

\begin{figure}
    \centering
    \includegraphics[width=\linewidth]{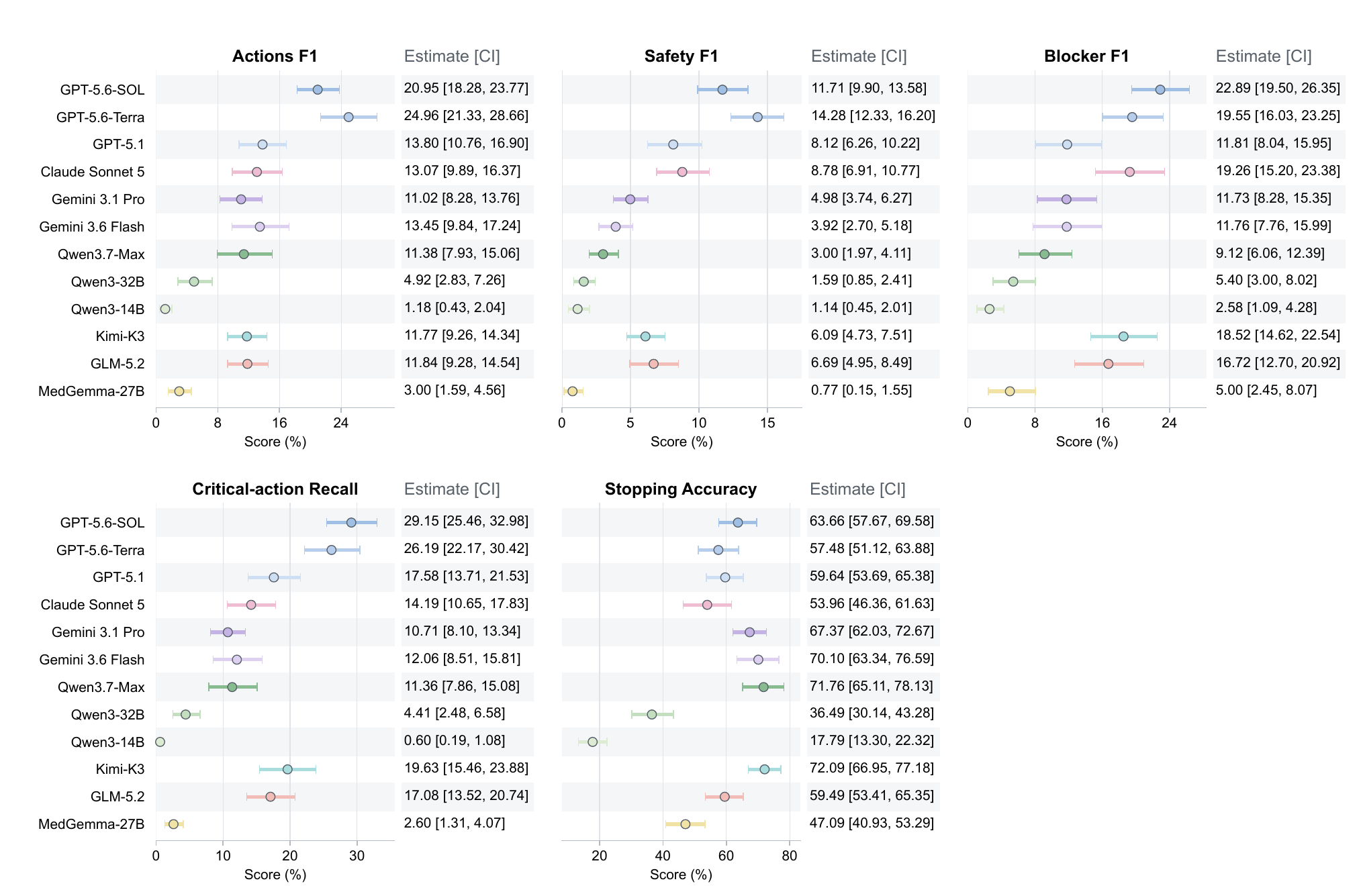}
    \caption*{Figure F1b: Complete checkpoint results---clinical decisions and safety.}
    \label{fig:placeholder}
\end{figure}
The primary analysis retains every attempted checkpoint. Unscorable submissions count as empty predictions and remain in the denominator. Table~F1c separates scorer-valid responses, complete responses that violate the evaluator contract, top-level schema failures, and transport or truncation failures.

\begin{table}[]
\centering
\scriptsize
\setlength{\tabcolsep}{3.4pt}
\renewcommand{\arraystretch}{1.06}
\caption*{\textbf{Table F1c: Checkpoint output validity and failure categories.}}
\begin{tabular}{lrrrrr}
\toprule
\textbf{Model} & \textbf{Scorer-valid} & \textbf{Valid rate} & \textbf{Contract-invalid} & \textbf{Schema-invalid} & \textbf{Transport/truncated} \\
\midrule
GPT-5.6-SOL & 231 & 96.25 & 8 & 1 & 0 \\
GPT-5.6-Terra & 219 & 91.25 & 13 & 8 & 0 \\
GPT-5.1 & 228 & 95.00 & 2 & 10 & 0 \\
Claude Sonnet 5 & 209 & 87.08 & 17 & 14 & 0 \\
Gemini 3.1 Pro & 236 & 98.33 & 2 & 2 & 0 \\
Gemini 3.6 Flash & 232 & 96.67 & 2 & 6 & 0 \\
Qwen3.7-Max & 233 & 97.08 & 0 & 7 & 0 \\
Qwen3-32B & 188 & 78.33 & 23 & 29 & 0 \\
Qwen3-14B & 83 & 34.58 & 19 & 138 & 0 \\
Kimi-K3 & 235 & 97.92 & 3 & 2 & 0 \\
GLM-5.2 & 230 & 95.83 & 2 & 8 & 0 \\
MedGemma-27B & 185 & 77.08 & 3 & 52 & 0 \\
Baichuan-M3 & 0 & 0.00 & 0 & 240 & 0 \\
MedGemma-4B & 0 & 0.00 & 0 & 240 & 0 \\
Med-PaLM 2 & 0 & 0.00 & 0 & 240 & 0 \\
\bottomrule
\end{tabular}

\vspace{2pt}
{\tiny Counts sum to 240 checkpoints per model. Valid rate is the percentage of all 240 checkpoints that passed both the top-level schema and the scorer contract. Contract-invalid denotes a parsed top-level submission that failed a scorer-level referential or uniqueness constraint.}
\end{table}

\subsection{Longitudinal Results and Eligibility}
\label{app:f-longitudinal}

Figure~F2a and F2b report the six primary longitudinal diagnostics. Parentheses give the number of cases with a nonzero eligible denominator. Brackets give 95\% paired case-bootstrap intervals from 20,000 replicates. Each point estimate is the unweighted mean of eligible case-specific rates.

\begin{figure}
    \centering
    \includegraphics[width=\linewidth]{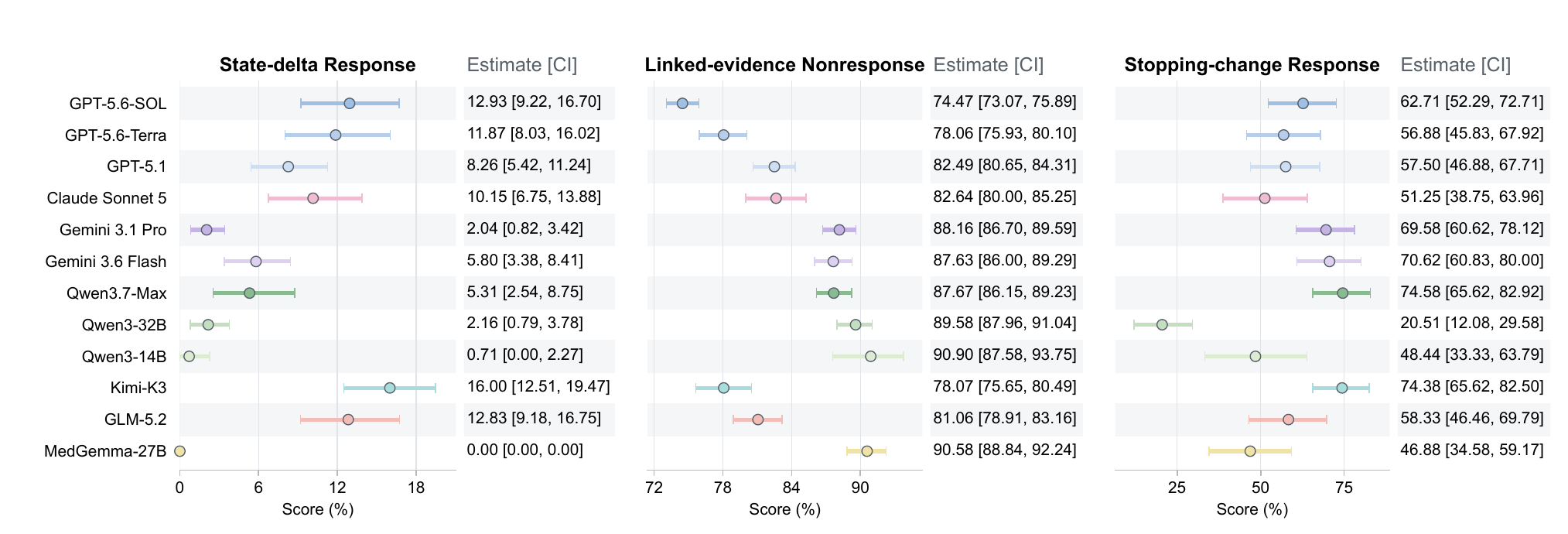}
    \caption*{Figure F2a: Immediate longitudinal response.}
    \label{fig:placeholder}
\end{figure}

\begin{figure}
    \centering
    \includegraphics[width=\linewidth]{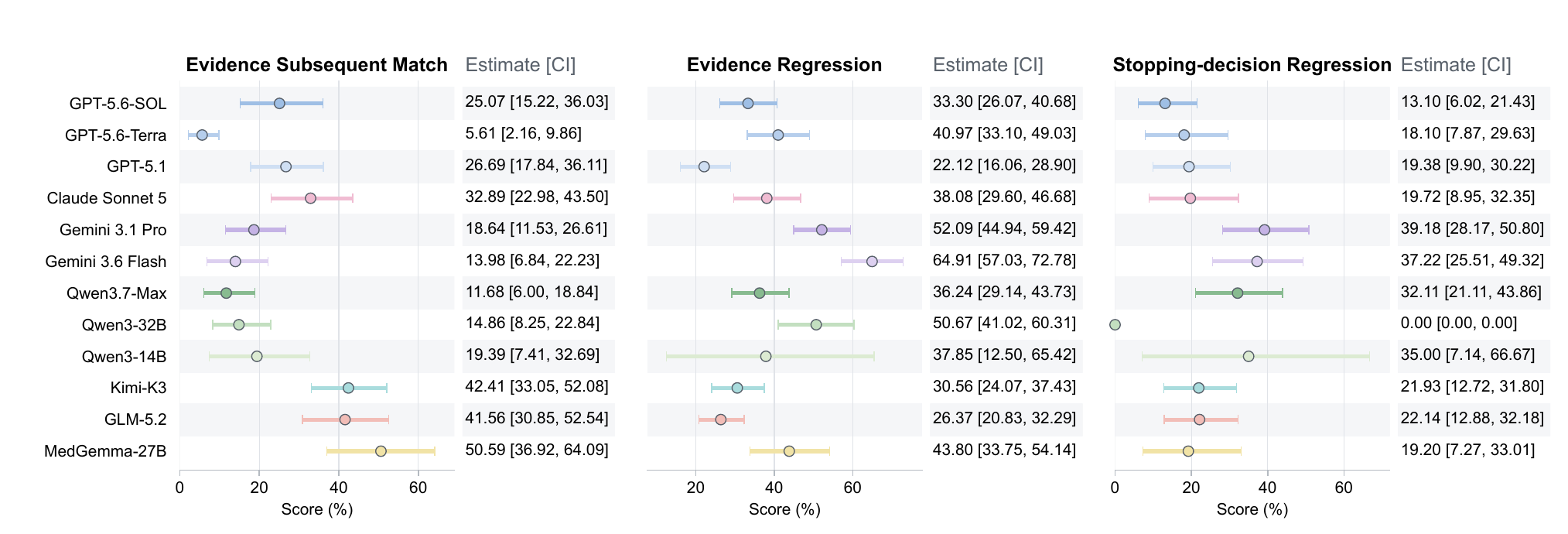}
    \caption*{Figure F2b: Delayed matching and longitudinal stability.}
    \label{fig:placeholder}
\end{figure}
Table~F2c gives the aggregate opportunity denominators for these case-macro estimates. These counts show differences in eligibility. Case-macro estimates retain equal case weights.

\begin{table}
\centering
\tiny
\setlength{\tabcolsep}{2.8pt}
\renewcommand{\arraystretch}{1.05}
\caption*{\textbf{Table F2c: Opportunity counts for primary longitudinal diagnostics.}}
\begin{tabular}{lrrrrrr}
\toprule
\textbf{Model} & \textbf{State $\Delta$} & \textbf{Linked} & \textbf{Stop change} & \textbf{Later match} & \textbf{Evidence reg.} & \textbf{Stop reg.} \\
\midrule
GPT-5.6-SOL & 365 & 4166 & 91 & 391 & 457 & 82 \\
GPT-5.6-Terra & 344 & 3620 & 88 & 417 & 304 & 69 \\
GPT-5.1 & 365 & 4101 & 92 & 243 & 668 & 77 \\
Claude Sonnet 5 & 334 & 3610 & 84 & 280 & 424 & 72 \\
Gemini 3.1 Pro & 375 & 3825 & 95 & 446 & 366 & 102 \\
Gemini 3.6 Flash & 367 & 3057 & 95 & 487 & 299 & 98 \\
Qwen3.7-Max & 366 & 4062 & 91 & 439 & 395 & 101 \\
Qwen3-32B & 314 & 2681 & 83 & 317 & 183 & 42 \\
Qwen3-14B & 139 & 1163 & 44 & 62 & 28 & 12 \\
Kimi-K3 & 374 & 4244 & 91 & 301 & 690 & 101 \\
GLM-5.2 & 361 & 4102 & 93 & 331 & 608 & 78 \\
MedGemma-27B & 288 & 3039 & 73 & 202 & 300 & 52 \\
\bottomrule
\end{tabular}
\end{table}

\subsection{Stopping Decisions and Blockers}
\label{app:f-stopping}

Stopping is a four-class categorical decision. Remaining blockers are scored as a set of semantically matched items. The two outcomes differ in chance structure, error modes, and score ranges. Table~F3a gives the Gold distribution. Always selecting the majority class, \textit{continue investigation}, yields 41.67\% checkpoint-micro accuracy.

\begin{table}
\centering
\scriptsize
\setlength{\tabcolsep}{5pt}
\renewcommand{\arraystretch}{1.08}
\caption*{\textbf{Table F3a: Gold stopping-decision distribution.}}
\begin{tabular}{lrr}
\toprule
\textbf{Stopping category} & \textbf{Checkpoints} & \textbf{Share (\%)} \\
\midrule
Continue investigation & 100 & 41.67 \\
Continue treatment & 82 & 34.17 \\
Wait for pending result & 31 & 12.92 \\
Safe to discharge or follow up & 27 & 11.25 \\
\midrule
Total & 240 & 100.00 \\
\bottomrule
\end{tabular}
\end{table}

Table~F3b reports checkpoint-micro stopping accuracy by Gold class. Overall stopping accuracy in Table~F1b uses the prespecified case-macro estimate.

\begin{table}
\centering
\scriptsize
\setlength{\tabcolsep}{3.4pt}
\renewcommand{\arraystretch}{1.06}
\caption*{\textbf{Table F3b: Stopping-decision accuracy by Gold category.}}
\begin{tabular}{lrrrr}
\toprule
\textbf{Model} & \textbf{Investigate} & \textbf{Treat} & \textbf{Wait} & \textbf{Discharge/follow up} \\
\midrule
GPT-5.6-SOL & 90.0 & 41.5 & 29.0 & 77.8 \\
GPT-5.6-Terra & 85.0 & 22.0 & 71.0 & 51.9 \\
GPT-5.1 & 92.0 & 36.6 & 38.7 & 44.4 \\
Claude Sonnet 5 & 78.0 & 41.5 & 32.3 & 44.4 \\
Gemini 3.1 Pro & 73.0 & 52.4 & 74.2 & 92.6 \\
Gemini 3.6 Flash & 74.0 & 70.7 & 51.6 & 85.2 \\
Qwen3.7-Max & 78.0 & 72.0 & 45.2 & 85.2 \\
Qwen3-32B & 75.0 & 0.0 & 3.2 & 44.4 \\
Qwen3-14B & 26.0 & 14.6 & 6.5 & 18.5 \\
Kimi-K3 & 90.0 & 59.8 & 48.4 & 77.8 \\
GLM-5.2 & 86.0 & 35.4 & 48.4 & 59.3 \\
MedGemma-27B & 80.0 & 19.5 & 25.8 & 33.3 \\
\bottomrule
\end{tabular}
\end{table}

Overall stopping accuracy reaches 72.09\% for Kimi-K3 and 71.76\% for Qwen3.7-Max. The best remaining-blocker F1 is 22.89\% for GPT-5.6-SOL. Correct category selection can therefore coexist with poor identification of unresolved clinical dependencies. Qwen3-32B's zero observed stopping-decision regression covers 42 eligible stable-category opportunities in 25 cases. Its stopping-change response is 20.51\%. These two diagnostics measure stability and responsiveness separately.

\subsection{Interactive Results and Failures}
\label{app:f-interactive}

The Interactive Track records one trajectory for each of four models on each of the same prespecified 10 cases. Table~F4a reports trajectory outcomes and process diagnostics. Productive Budget Yield divides by the configured action budget, including unused slots. Advance among executed actions in Table~F4b pools over executed actions.

\begin{table}
\centering
\tiny
\setlength{\tabcolsep}{2.3pt}
\renewcommand{\arraystretch}{1.08}
\caption*{\textbf{Table F4a: Full model-level Interactive Track results.}}
\resizebox{\linewidth}{!}{%
\begin{tabular}{lrrrrrrrr}
\toprule
\textbf{Model} & \textbf{Quality} & \textbf{Milestone} & \textbf{Budget yield} & \textbf{Evidence recall} & \textbf{Pooled repeats} & \textbf{Longest stagnation} & \textbf{Post-evidence advance} & \textbf{Pending recovery} \\
 & \textbf{(1--7)} & \textbf{(\%)} & \textbf{(\%)} & \textbf{(\%)} & \textbf{(\%)} & \textbf{(\%)} & \textbf{(\%)} & \textbf{(\%)} \\
\midrule
GPT-5.6-SOL & 4.4 & 38.13 & 13.75 & 41.98 & 13.61 & 53.50 & 72.57 & 66.67 \\
GPT-5.6-Terra & 2.8 & 27.91 & 10.75 & 31.32 & 31.05 & 61.75 & 73.33 & 46.67 \\
Gemini 3.6 Flash & 2.3 & 11.36 & 4.50 & 12.57 & 20.87 & 70.33 & 36.90 & 6.25 \\
Kimi-K3 & 3.1 & 26.62 & 9.25 & 28.33 & 22.64 & 71.04 & 66.35 & 35.83 \\
\bottomrule
\end{tabular}}

\vspace{2pt}
{\tiny Milestone coverage, budget yield, evidence recall, longest stagnation, post-evidence advance, and pending recovery are trajectory-level means. Pooled repeats is the number of no-new-evidence semantic repeats divided by all executed actions.}
\end{table}

\begin{table}
\centering
\scriptsize
\setlength{\tabcolsep}{3.2pt}
\renewcommand{\arraystretch}{1.07}
\caption*{\textbf{Table F4b: Executed-action outcome composition and pending recovery.}}
\resizebox{\linewidth}{!}{%
\begin{tabular}{lrrrrrr}
\toprule
\textbf{Model} & \textbf{Executed $n$} & \textbf{Advance} & \textbf{Prerequisite pending} & \textbf{Valid exploration} & \textbf{Completed/no progress} & \textbf{Pending recovery} \\
\midrule
GPT-5.6-SOL & 360 & 51 (14.2\%) & 111 (30.8\%) & 160 (44.4\%) & 38 (10.6\%) & 66.7\% \\
GPT-5.6-Terra & 380 & 39 (10.3\%) & 154 (40.5\%) & 159 (41.8\%) & 28 (7.4\%) & 46.7\% \\
Gemini 3.6 Flash & 115 & 15 (13.0\%) & 14 (12.2\%) & 83 (72.2\%) & 3 (2.6\%) & 6.3\% \\
Kimi-K3 & 349 & 33 (9.5\%) & 54 (15.5\%) & 232 (66.5\%) & 30 (8.6\%) & 35.8\% \\
\bottomrule
\end{tabular}}

\vspace{2pt}
{\tiny Action-outcome percentages are pooled within model over executed actions. Pending recovery is the case-macro proportion of unique prerequisite-pending transition signatures later completed in the same trajectory.}
\end{table}

Table~F4c reports case-level progress as completed reference transitions divided by executed actions. The budget is 40 actions except for WARD-I07, whose prespecified budget is 20.

\begin{table}
\centering
\scriptsize
\setlength{\tabcolsep}{4.2pt}
\renewcommand{\arraystretch}{1.05}
\caption*{\textbf{Table F4c: Case-level interactive progress (transitions/actions).}}
\begin{tabular}{lrrrrr}
\toprule
\textbf{Case} & \textbf{Budget} & \textbf{GPT-5.6-SOL} & \textbf{GPT-5.6-Terra} & \textbf{Gemini Flash} & \textbf{Kimi-K3} \\
\midrule
WARD-I01 & 40 & 6/40 & 6/40 & 3/9 & 8/31 \\
WARD-I02 & 40 & 10/40 & 10/40 & 4/8 & 3/32 \\
WARD-I03 & 40 & 1/40 & 0/40 & 0/15 & 0/40 \\
WARD-I04 & 40 & 7/40 & 2/40 & 1/6 & 5/38 \\
WARD-I05 & 40 & 6/40 & 6/40 & 1/9 & 1/40 \\
WARD-I06 & 40 & 7/40 & 5/40 & 1/3 & 5/28 \\
WARD-I07 & 20 & 4/20 & 4/20 & 3/4 & 4/20 \\
WARD-I08 & 40 & 0/40 & 0/40 & 0/40 & 0/40 \\
WARD-I09 & 40 & 8/20 & 4/40 & 2/10 & 4/40 \\
WARD-I10 & 40 & 2/40 & 2/40 & 0/11 & 3/40 \\
\midrule
Total & 380 & 51/360 & 39/380 & 15/115 & 33/349 \\
\bottomrule
\end{tabular}
\end{table}

Only GPT-5.6-SOL reached an environment terminal state, on WARD-I09. Nine SOL trajectories and all ten Terra trajectories reached their action caps. Gemini 3.6 Flash reached its action cap on one case and ended after repeated invalid tool output on nine cases. Kimi-K3 reached its action cap on six cases and was transport-censored on four. Table~F4d reports termination categories and unused action slots.

\begin{table}
\centering
\small
\setlength{\tabcolsep}{3pt}
\renewcommand{\arraystretch}{1.08}
\caption*{\textbf{Table F4d: Interactive termination and action-budget use.}}

\begin{tabular}{@{}lcccccc@{}}
\toprule
\textbf{Model} &
\shortstack{\textbf{Terminal}\\\textbf{success}} &
\shortstack{\textbf{Action cap}\\\textbf{reached}} &
\shortstack{\textbf{Invalid-output}\\\textbf{termination}} &
\shortstack{\textbf{Transport-}\\\textbf{censored}} &
\shortstack{\textbf{Executed}\\\textbf{/ budget}} &
\shortstack{\textbf{Unused}\\\textbf{slots}} \\
\midrule
GPT-5.6-SOL       & 1 & 9  & 0 & 0 & 360/380 & 20  \\
GPT-5.6-Terra     & 0 & 10 & 0 & 0 & 380/380 & 0   \\
Gemini 3.6 Flash  & 0 & 1  & 9 & 0 & 115/380 & 265 \\
Kimi-K3           & 0 & 6  & 0 & 4 & 349/380 & 31  \\
\bottomrule
\end{tabular}
\end{table}

Nine Flash trajectories ended after repeated invalid tool output. The remaining trajectory exhausted its action budget without reaching the terminal state. Four Kimi-K3 trajectories were transport-censored. All six uninterrupted trajectories exhausted their budgets without reaching the terminal state. Technical interruptions limit interpretation of completion counts. Uninterrupted runs also failed to complete.

\begin{figure}
    \centering
    \includegraphics[width=\linewidth]{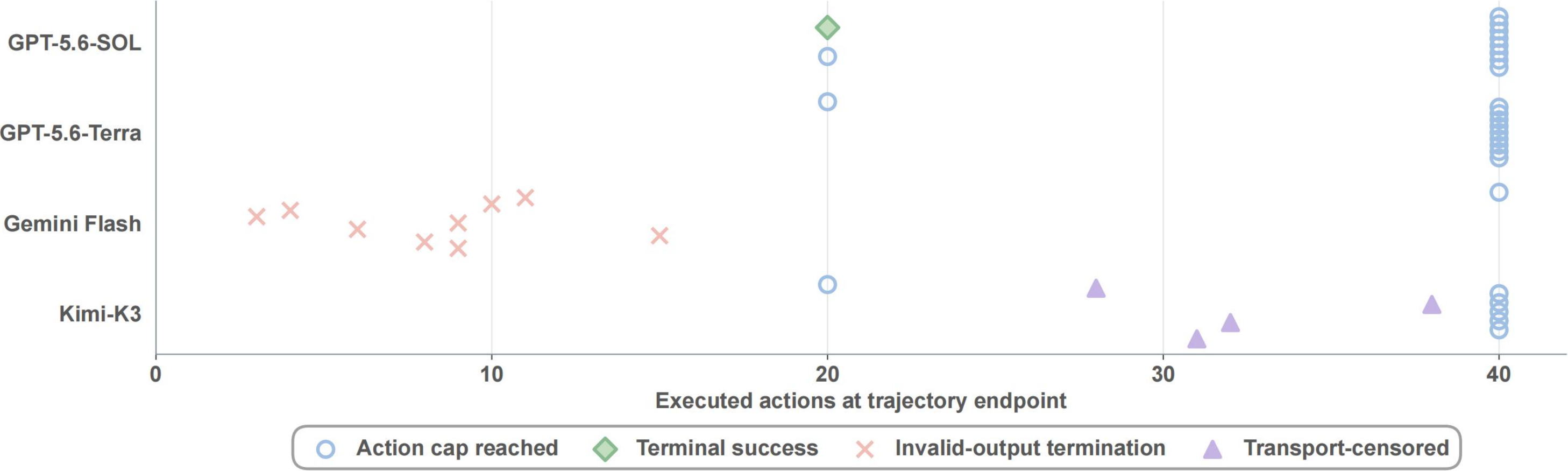}
    \caption*{\textbf{Figure F4e: Interactive trajectory endpoints.} Each marker represents one model--case trajectory. Vertical jitter separates coincident endpoints and has no quantitative meaning. WARD-I07 had a 20-action cap; all other cases had a 40-action cap.}
    \label{fig:placeholder}
\end{figure}

The 39 nonterminal trajectories received non-exclusive clinical-process failure labels. Repeated or prolonged stagnation occurred in 30 trajectories and unresolved prerequisites in 22.

\begin{table}
\centering
\scriptsize
\setlength{\tabcolsep}{6pt}
\renewcommand{\arraystretch}{1.08}
\caption{Trajectory-level failure mechanisms among nonterminal interactive runs.}
\label{tab:failure-mechanisms}
\begin{tabular}{lcc}
\toprule
\textbf{Failure mechanism} & \textbf{$n$} & \textbf{Share (\%)} \\
\midrule
Repeated or prolonged stagnation & 30 & 76.9 \\
Unresolved prerequisite at termination & 22 & 56.4 \\
No adaptation after newly released evidence & 16 & 41.0 \\
Prematurely timed clinical action & 15 & 38.5 \\
Tool-protocol failure & 9 & 23.1 \\
\bottomrule
\end{tabular}

\vspace{2pt}
{\tiny Denominator: 39 nonterminal trajectories. Labels are non-exclusive, so percentages do not sum to 100\%.}
\end{table}

\subsection{Valid-output Sensitivity}
\label{app:f-valid-output}

The prespecified primary checkpoint analysis scores every submitted checkpoint, treating invalid outputs as empty predictions. Two sensitivity analyses cover the original eight-model subset (GPT-5.6-SOL, GPT-5.6-Terra, Claude Sonnet 5, Gemini 3.1 Pro, Gemini 3.6 Flash, Qwen3.7-Max, Kimi-K3, and GLM-5.2). Model-specific valid-only analysis uses each model's scorer-valid checkpoints. Shared-valid pairwise analysis uses checkpoints valid for both models in a pair. The expanded 12-model primary-result set was not reranked.

Valid-only conditioning raises point estimates in Table~F5a. The primary leader is unchanged for eight of nine checkpoint outcomes. Only Stopping accuracy changes point-estimate leader, to Qwen3.7-Max.

\begin{table}
\centering
\scriptsize
\setlength{\tabcolsep}{3.1pt}
\renewcommand{\arraystretch}{1.06}
\caption*{\textbf{Table F5a: Primary and model-specific valid-only leaders.}}
\resizebox{\linewidth}{!}{%
\begin{tabular}{lcccc}
\toprule
\textbf{Metric} & \textbf{Primary leader (\%)} & \textbf{Valid-only leader (\%)} & \textbf{Valid-only 95\% CI} & \textbf{Changed} \\
\midrule
Evidence F1 & Kimi-K3 (63.40) & Kimi-K3 (64.93) & [60.05, 69.98] & No \\
Utilization F1 & GPT-5.6-SOL (18.87) & GPT-5.6-SOL (19.35) & [17.72, 21.03] & No \\
Claims F1 & GPT-5.6-Terra (38.29) & GPT-5.6-Terra (40.21) & [36.31, 44.18] & No \\
Patient State F1 & GPT-5.6-SOL (27.93) & GPT-5.6-SOL (28.56) & [26.78, 30.30] & No \\
Actions F1 & GPT-5.6-Terra (24.96) & GPT-5.6-Terra (26.67) & [22.67, 30.69] & No \\
Safety F1 & GPT-5.6-Terra (14.28) & GPT-5.6-Terra (14.96) & [12.93, 16.94] & No \\
Stopping-blocker F1 & GPT-5.6-SOL (22.89) & GPT-5.6-SOL (23.26) & [19.86, 26.69] & No \\
Stopping accuracy & Kimi-K3 (72.09) & Qwen3.7-Max (74.13) & [67.49, 80.34] & Yes \\
Critical-action recall & GPT-5.6-SOL (29.15) & GPT-5.6-SOL (30.65) & [26.57, 34.90] & No \\
\bottomrule
\end{tabular}}
\end{table}

Each model's largest valid-only shift occurred in Stopping accuracy, ranging from 1.17 to 6.57 percentage points. The other eight metric leaders were unchanged. Shared-valid comparisons preserve 246 of 252 pairwise directions (97.62\%; Table~F5b).

\begin{table}
\centering
\scriptsize
\setlength{\tabcolsep}{4.2pt}
\renewcommand{\arraystretch}{1.06}
\caption*{\textbf{Table F5b: Direction preservation under shared-valid pairing.}}
\begin{tabular}{lccc}
\toprule
\textbf{Metric} & \textbf{Model pairs} & \textbf{Direction preserved} & \textbf{Preserved (\%)} \\
\midrule
Evidence F1 & 28 & 27 & 96.43 \\
Utilization F1 & 28 & 27 & 96.43 \\
Claims F1 & 28 & 26 & 92.86 \\
Patient State F1 & 28 & 28 & 100.00 \\
Actions F1 & 28 & 28 & 100.00 \\
Safety F1 & 28 & 28 & 100.00 \\
Stopping-blocker F1 & 28 & 27 & 96.43 \\
Stopping accuracy & 28 & 27 & 96.43 \\
Critical-action recall & 28 & 28 & 100.00 \\
\midrule
All outcomes & 252 & 246 & 97.62 \\
\bottomrule
\end{tabular}

\vspace{2pt}
{\tiny Shared-valid checkpoint coverage ranges from 80.00\% to 96.67\% across model pairs. Each paired difference uses 20,000 case-bootstrap replicates with a shared resampling index.}
\end{table}

All-submission analysis measures end-to-end protocol performance. Model-specific valid-only analysis measures clinical content conditional on a valid response. Shared-valid analysis compares models on the same checkpoints. The evidence-to-state-to-action separation and eight of the nine checkpoint leaders remain stable across these analyses.

\section{Environment Validation}
\label{app:g-environment}

We evaluate reference-policy reachability and sensitivity to the action budget.
Separate audits examine clinically reasonable non-advancing actions and resolver
errors on fully expressed supported actions. All analyses use the same ten
prespecified cases and the main experiment's tool, feedback, dependency, and
termination rules.

\subsection{Reference-policy Reachability}
\label{app:g-reachability}

We applied the reference policy to each case using the environment's transition
rules and dependency graph. The policy follows legal prerequisites in reference
order solely to validate the environment. Evaluated models receive no reference
actions or hidden transition identifiers. Reachability requires a terminal state
reached through ordinary tool execution. All ten cases reached a terminal state
within their original action caps. The longest verified reference-policy path
required 38 actions.

\begin{table}
\centering
\scriptsize
\setlength{\tabcolsep}{4.0pt}
\renewcommand{\arraystretch}{1.06}
\caption*{\textbf{Table G1: Per-case reference-policy reachability.}}
\begin{tabular}{@{}lrrc@{\qquad}lccc@{}}
\toprule
\textbf{Case} & \textbf{Ref. T} & \textbf{Policy calls} & \textbf{Terminal} &
\textbf{Case} & \textbf{Ref. T} & \textbf{Policy calls} & \textbf{Terminal} \\
\midrule
WARD-001 & 12 & $\leq 38$ & Yes & WARD-006 & 12 & $\leq 38$ & Yes \\
WARD-002 & 19 & $\leq 38$ & Yes & WARD-007 & 12 & $\leq 38$ & Yes \\
WARD-003 & 13 & $\leq 38$ & Yes & WARD-008 &  9 & $\leq 38$ & Yes \\
WARD-004 & 18 & $\leq 38$ & Yes & WARD-009 &  8 & $\leq 38$ & Yes \\
WARD-005 & 27 & $\leq 38$ & Yes & WARD-010 & 11 & $\leq 38$ & Yes \\
\bottomrule
\end{tabular}

\vspace{2pt}
\begin{minipage}{0.94\linewidth}
\footnotesize
\textit{Note.} Ref. T counts reference transitions in the case action graph;
policy tool-call counts may differ. The exact all-case maximum is 38 calls.
Per-case values are verified upper limits; exact call counts are unreported.
\end{minipage}
\end{table}

The terminal targets are executable within the original budget. Evaluated agents
must still infer the next clinical objective and an executable tool expression
from the information available at each turn.

\subsection{Action-budget Sensitivity}
\label{app:g-budget}

We extended the evaluated-model action cap to 60 actions from the original
40-action cap (20 actions for WARD-I07). Cases, models, tool contracts, and the
terminal definition were unchanged. No additional trajectory reached a terminal
state. GPT-5.6-SOL remained at one terminal completion; the other three models
remained at zero. WARD-009 remained the only case completed by any evaluated model.

\begin{table}
\centering
\scriptsize
\setlength{\tabcolsep}{5pt}
\renewcommand{\arraystretch}{1.06}
\caption*{\textbf{Table G2: Terminal completions under the original and extended action budgets.}}
\begin{tabular}{@{}lccc@{}}
\toprule
\textbf{Model} & \textbf{Original cap} & \textbf{60 actions} & \textbf{Change} \\
\midrule
GPT-5.6-SOL       & 1/10 & 1/10 & 0 \\
GPT-5.6-Terra     & 0/10 & 0/10 & 0 \\
Gemini-3.6-Flash  & 0/10 & 0/10 & 0 \\
Kimi-K3           & 0/10 & 0/10 & 0 \\
\midrule
All model--case trajectories & 1/40 & 1/40 & 0 \\
\bottomrule
\end{tabular}
\end{table}

\begin{table}
\centering
\scriptsize
\setlength{\tabcolsep}{4.2pt}
\renewcommand{\arraystretch}{1.05}
\caption*{\textbf{Table G2 (continued): Number of models reaching terminal state in each case.}}
\begin{tabular}{@{}lcc@{\qquad}lcc@{}}
\toprule
\textbf{Case} & \textbf{Original cap} & \textbf{60} &
\textbf{Case} & \textbf{Original cap} & \textbf{60} \\
\midrule
WARD-001 & 0/4 & 0/4 & WARD-006 & 0/4 & 0/4 \\
WARD-002 & 0/4 & 0/4 & WARD-007 & 0/4 & 0/4 \\
WARD-003 & 0/4 & 0/4 & WARD-008 & 0/4 & 0/4 \\
WARD-004 & 0/4 & 0/4 & WARD-009 & 1/4 & 1/4 \\
WARD-005 & 0/4 & 0/4 & WARD-010 & 0/4 & 0/4 \\
\bottomrule
\end{tabular}
\end{table}

The 60-action cap left terminal completion unchanged. Action-level composition
and milestone coverage are reported separately in
Section~\ref{app:f-interactive}.

\subsection{Reasonable Non-advancing Actions}
\label{app:g-nonadvancing}

A clinical review examined five failed trajectories with a possible gap between
clinical usefulness and reference-graph advancement. The review considered only
valid executed actions. It marked clinically reasonable actions that released
no new reference evidence and completed no reference transition. Among 168 valid
non-advancing actions, 24 were clinically reasonable and unrewarded by the
reference graph. The 24/168 count describes this selected audit set; it cannot
estimate prevalence or global action-space coverage.

\begin{table}
\centering
\scriptsize
\setlength{\tabcolsep}{5pt}
\renewcommand{\arraystretch}{1.06}
\caption*{\textbf{Table G3: Focused review of clinically reasonable non-advancing actions.}}
\begin{tabular}{@{}llcc@{}}
\toprule
\textbf{Model} & \textbf{Case} &
\textbf{Reasonable, unrewarded} & \textbf{Valid non-advancing} \\
\midrule
GPT-5.6-SOL      & WARD-008 & 3 & 40 \\
GPT-5.6-Terra    & WARD-003 & 5 & 40 \\
GPT-5.6-Terra    & WARD-008 & 6 & 40 \\
Gemini-3.6-Flash & WARD-005 & 4 &  8 \\
Gemini-3.6-Flash & WARD-008 & 6 & 40 \\
\midrule
Audit total      &           & 24 & 168 \\
\bottomrule
\end{tabular}

\vspace{2pt}
\begin{minipage}{0.94\linewidth}
\footnotesize
\textit{Note.} The five trajectories were selected for mechanism review. The 24
actions included clinically directed breast imaging, pre-procedure laboratory
assessment, biopsy work-up, and review of admission, pathology, and preoperative
records. These actions did not satisfy the currently available reference
transition.
\end{minipage}
\end{table}

A valid clinical action can be useful without advancing the current reference
state. It remains non-advancing for milestone and outcome-composition scoring.
Non-advancement alone establishes neither action invalidity nor a resolver error.

\subsection{Resolver Errors}
\label{app:g-resolver}

A \emph{resolver miss} occurs when an executable query fully expresses the
clinical target and timing of a currently available reference transition, yet
the environment fails to advance. The prespecified tool contract requires this
information in the executable query; intent and rationale fields cannot supply
missing details.

\begin{table}
\centering
\scriptsize
\setlength{\tabcolsep}{4.5pt}
\renewcommand{\arraystretch}{1.06}
\caption*{\textbf{Table G4: Focused action-resolution audit on five trajectories.}}
\begin{tabular}{@{}llccc@{}}
\toprule
\textbf{Model} & \textbf{Case} & \textbf{Valid/total calls} &
\textbf{Advances} & \textbf{Fully specified, no advance} \\
\midrule
GPT-5.6-SOL      & WARD-008 & 40/40 & 0 & 0 \\
GPT-5.6-Terra    & WARD-003 & 40/40 & 0 & 0 \\
GPT-5.6-Terra    & WARD-008 & 40/40 & 0 & 0 \\
Gemini-3.6-Flash & WARD-005 &  9/12 & 1 & 0 \\
Gemini-3.6-Flash & WARD-008 & 40/40 & 0 & 0 \\
\midrule
Audit total      &           & 169/172 & 1 & 0 \\
\bottomrule
\end{tabular}
\end{table}

Across 169 valid actions, none fully expressed a compatible, currently available
transition and then failed to advance. Four standardized, fully specified
actions were also evaluated: the first available transition in WARD-003, the
first and second blocking transitions in WARD-005, and the first available
transition in WARD-008. All four advanced as intended. Failures in this audit
involved incomplete action expression or ordering. We found no resolver false
negative within this scope.

\section{Case Studies}
\label{app:h-cases}

We present one Longitudinal Track example and one Interactive Track example.
The first follows Qwen3-14B on WARD-003, where the model correctly identifies
two visible cytology results but continues to treat them as pending in its
patient state, next action, and stopping blocker. The second follows
GPT-5.6-SOL on WARD-001, where the agent recovers from early prerequisite
failures and establishes the primary diagnosis, but makes no further
reference-path progress before exhausting its action budget.

\begingroup
\raggedbottom
\definecolor{wardcasetitle}{HTML}{667587}
\definecolor{wardcasetime}{HTML}{356DC2}
\definecolor{wardcaseevidence}{HTML}{248777}
\definecolor{wardcaseexpected}{HTML}{B56C22}
\definecolor{wardcaseobserved}{HTML}{8058AC}
\definecolor{wardcasenote}{HTML}{586573}
\newcommand{\wardjsonfield}[4]{%
  \par\begingroup
  \setlength{\parindent}{0pt}%
  \setlength{\leftskip}{1em}%
  \setlength{\rightskip}{0pt plus 2em}%
  \setlength{\fboxsep}{2pt}%
  \noindent\colorbox{#2!10!white}{\textcolor{#2}{\bfseries\strut "#1"}}:\par\nobreak
  \setlength{\leftskip}{2em}%
  \noindent "#3"#4\par
  \endgroup\vspace{2.5pt}%
}

\newcommand{\wardjsonquote}[3]{%
  \par\begingroup
  \setlength{\parindent}{0pt}%
  \setlength{\leftskip}{2em}%
  \setlength{\rightskip}{0pt plus 2em}%
  \noindent\textcolor{wardcaseobserved}{\bfseries "#1"}: "#2"#3\par
  \endgroup\vspace{2pt}%
}

\subsection{Longitudinal: Evidence and State}
\label{app:h-linked-evidence}
\label{app:h-recovery}
\begingroup
\setlength{\parindent}{0pt}

48-year-old woman undergoing thyroid and cervical lymph-node evaluation.
All model excerpts below come from the same checkpoint response.\par

% =========================================================
% Checkpoint task
% =========================================================

\begin{wardcasebox}{wardcase system}{CHECKPOINT TASK}
{
  "question_en": "At this time point, use only available evidence to identify key evidence, supported claims, patient state, appropriate next actions, and whether stopping or follow-up is safe."
}
\end{wardcasebox}

% =========================================================
% Evidence already visible to the model
% =========================================================

\begin{wardcasebox}{wardcase user}{VISIBLE EVIDENCE}
[
  {
    "evidence_id": "path:006",
    "title": "Cytopathology",
    "content_summary": "Follicular epithelial cells arranged in sheets and papillary formations are seen. Some cells are crowded, with enlarged nuclei; nuclear grooves and intranuclear inclusions are visible in occasional cells, consistent with papillary thyroid carcinoma. Note: Bethesda reporting category VI"
  },
  {
    "evidence_id": "path:007",
    "title": "Cytopathology",
    "content_summary": "Lymphocytes are seen; no evidence of malignancy is seen."
  }
]
\end{wardcasebox}

Both cytology reports are already visible. Thyroid and regional lymph-node
surgery are documented as completed. Final postoperative pathology remains
unavailable.\par

% =========================================================
% Correct evidence recognition
% =========================================================

\begin{wardcasebox}{wardcase model,coltext=wardCaseGreen}{MODEL RESPONSE: EVIDENCE}
{
  "key_evidence": [
    {
      "evidence_id": "path:006",
      "interpretation_en": "Cytopathology showing follicular epithelial cells arranged in sheets and papillary formations, with nuclear grooves and inclusions consistent with papillary thyroid carcinoma",
      "role": "critical"
    },
    {
      "evidence_id": "path:007",
      "interpretation_en": "Cytopathology of the cervical lymph node showing lymphocytes with no evidence of malignancy",
      "role": "critical"
    }
  ]
}
\end{wardcasebox}

% =========================================================
% Patient state still treats the reports as pending
% =========================================================

\begin{wardcasebox}{wardcase model,coltext=wardCaseRed}{MODEL RESPONSE: PATIENT STATE}
{
  "patient_state": {
    "active_problems": [
      {
        "evidence_ids": ["path:006", "path:007"],
        "status": "pending",
        "text_en": "Fine-needle aspiration of thyroid and cervical lymph node performed with indeterminate results"
      }
    ],
    "required_delta_from_previous_checkpoint": [
      {
        "evidence_ids": ["img:010", "path:006"],
        "status": "in_progress",
        "text_en": "Patient currently awaiting cytopathology results from the fine-needle aspiration to determine malignancy status."
      }
    ]
  }
}
\end{wardcasebox}

% =========================================================
% Next action requests the same results
% =========================================================

\begin{wardcasebox}{wardcase model,coltext=wardCaseRed}{MODEL RESPONSE: NEXT ACTION}
{
  "actions": [
    {
      "action_scope": "prospective",
      "category": "necessary",
      "evidence_ids": ["img:010"],
      "text_en": "Obtain cytopathology results from the ultrasound-guided fine-needle aspiration of the thyroid and cervical lymph node."
    }
  ]
}
\end{wardcasebox}

% =========================================================
% Stopping blocker explicitly names the same two reports
% =========================================================

\begin{wardcasebox}{wardcase model,coltext=wardCaseRed}{MODEL RESPONSE: STOPPING}
{
  "stopping": {
    "decision": "continue_investigation",
    "rationale_en": "The patient needs further evaluation of the cytopathology results to confirm malignancy status before proceeding with surgical intervention.",
    "remaining_blockers": [
      "Awaiting cytopathology results (evidence_ids: path:006, path:007)."
    ]
  }
}
\end{wardcasebox}

% =========================================================
% Reference expectation
% =========================================================

\medskip
\textbf{Reference expectation}\par
\smallskip

Retain both reported cytology findings and remove waiting for these reports
as a blocker. Continue postoperative treatment and monitoring, with review
of final surgical pathology when available.\par

% =========================================================
% Case outcome
% =========================================================

\medskip
\textbf{Case outcome}\par
\smallskip

The model correctly restates both cytology results but still treats them
as pending in its patient state, next action, and stopping blocker.
The blocker explicitly cites the same two reports.
No later evaluated checkpoint is available to assess recovery.\par

\endgroup

\subsection{Interactive: Progress and Stagnation}
\label{app:h-prerequisite}

The patient was a 69-year-old woman with a 14.5mm BI-RADS 4C right-breast mass,
diabetes lasting more than 20 years, and enlarged left supraclavicular lymph
nodes. The sequence shows every executed action with its original call index.
Call 16 produced no executable action. It appears in the sequence and is excluded
from the total of 40 executed actions.
\begingroup
\setlength{\parindent}{0pt}

% =========================================================
% Shared system prompt
% =========================================================

\begin{wardcasebox}{wardcase system}{SYSTEM PROMPT}
{
  "role": "You are the treating inpatient clinician operating a deidentified Patient/EHR tool sandbox.",
  "objective": "Choose one clinically useful next tool action from visible evidence, the public inpatient workflow, and prior public tool outcomes. New clinical facts may come only from visible evidence or recorded EHR outcomes.",
  "rules": [
    "Call exactly one tool per turn.",
    "After a valid no-advance outcome, do not repeat the same wording. If the same clinical target remains justified, reformulate it once with a clearer record kind, event, lifecycle status, and temporal relation in args.query; otherwise choose another clinically supported target.",
    "Treat planned, consented, ordered, performed, completed, resulted, and pending as different lifecycle states. A template, plan, order, or consent does not prove performance or completion."
  ]
}
\end{wardcasebox}

Calls 0--1 complete history and admission review. The agent enters S2.\par

% =========================================================
% Call 2
% =========================================================

\begin{wardcasebox}{wardcase user}{USER PROMPT: CALL 2}
{
  "instruction": "Call exactly one provided clinical function now. Use the native tool-call channel and do not answer in prose.",
  "progress_guidance": {
    "stage_workflow_checklist": ["diagnosis", "staging", "safety", "procedures", "results", "treatment"]
  },
  "observation": {
    "stage": "S2",
    "visible_evidence": [
      {
        "evidence_id": "doc:005",
        "content_tail": "[...] Malignancy is considered likely. Right-breast core needle biopsy under local anesthesia is planned for today to establish the pathological nature of the mass. [...]"
      }
    ]
  }
}
\end{wardcasebox}

\begin{wardcasebox}{wardcase model}{MODEL TOOL CALL: CALL 2}
{
  "tool": "procedure.perform",
  "args": {
    "query": "Perform the planned ultrasound-guided core needle biopsy of the right breast upper-inner-quadrant mass under local anesthesia and submit tissue for pathology."
  }
}
\end{wardcasebox}

\begin{wardcasebox}{wardcase tool}{TOOL RESPONSE: CALL 2}
{
  "status": "recognized_prerequisite_pending",
  "message": "The action is understood, but its recorded prerequisites are not yet complete.",
  "state_advanced": false,
  "newly_released_evidence_ids": []
}
\end{wardcasebox}

% =========================================================
% Calls 3-7
% =========================================================

\medskip
\textbf{Calls 3--7: local recovery}\par
\smallskip

After a request to review imaging, call 4 retrieves assessment and consent
records. Call 5 still receives a pending response. Call 6 obtains breast
ultrasound. Call 7 completes the biopsy.\par

% =========================================================
% Call 8
% =========================================================

\begin{wardcasebox}{wardcase user}{USER PROMPT: CALL 8}
{
  "observation": {
    "stage": "S2",
    "new_evidence_ids": ["surg:004"],
    "visible_evidence": [
      {
        "evidence_id": "surg:004",
        "content_head": "Procedure: breast needle biopsy. Procedure type: diagnostic. Anesthesia: local."
      }
    ]
  }
}
\end{wardcasebox}

\begin{wardcasebox}{wardcase model}{MODEL TOOL CALL: CALL 8}
{
  "tool": "ehr.read",
  "args": {
    "query": "Review the resulted pathology report after the completed right breast mass core needle biopsy."
  }
}
\end{wardcasebox}

\begin{wardcasebox}{wardcase tool,coltext=wardCaseGreen}{TOOL RESPONSE: CALL 8}
{
  "status": "ehr_path_advanced",
  "message": "A recorded EHR action advanced and new evidence is available.",
  "state_advanced": true,
  "newly_released_evidence_ids": ["path:001"]
}
\end{wardcasebox}

Call 8 releases the pathology result and completes the sixth reference
transition.\par

% =========================================================
% Call 9
% =========================================================

\medskip
\textbf{Call 9: biomarker testing}\par

\begin{wardcasebox}{wardcase user}{USER PROMPT: CALL 9}
{
  "observation": {
    "new_evidence_ids": ["path:001"],
    "visible_evidence": [
      {
        "evidence_id": "path:001",
        "content_tail": "[...] Right breast mass, core needle biopsy specimen: invasive carcinoma. [...]"
      }
    ]
  }
}
\end{wardcasebox}

\begin{wardcasebox}{wardcase model}{MODEL TOOL CALL: CALL 9}
{
  "tool": "ehr.order",
  "args": {
    "query": "Order ER, PR, HER2, and Ki-67 immunohistochemistry on the completed right breast core needle biopsy specimen after pathology resulted as invasive carcinoma."
  }
}
\end{wardcasebox}

\begin{wardcasebox}{wardcase tool}{TOOL RESPONSE: CALL 9}
{
  "status": "valid_exploration_no_recorded_outcome",
  "message": "This valid exploration has no currently releasable recorded EHR outcome.",
  "state_advanced": false,
  "newly_released_evidence_ids": []
}
\end{wardcasebox}

% =========================================================
% Call 10
% =========================================================

\medskip
\textbf{Call 10: the query repeats despite feedback}\par

\begin{wardcasebox}{wardcase user}{USER PROMPT: CALL 10}
{
  "progress_guidance": {
    "last_outcome_status": "valid_exploration_no_recorded_outcome"
  },
  "observation": {
    "new_evidence_ids": []
  },
  "recovery_guidance": {
    "instruction": "Do not repeat the same wording. If the target remains clinically justified, restate the executable query with a clear record kind, clinical event, lifecycle status when known, and temporal relation; otherwise choose another evidence-supported action."
  }
}
\end{wardcasebox}

\begin{wardcasebox}{wardcase model}{MODEL TOOL CALL: CALL 10}
{
  "tool": "ehr.order",
  "args": {
    "query": "Order ER, PR, HER2, and Ki-67 immunohistochemistry on the completed right breast core needle biopsy specimen after pathology resulted as invasive carcinoma."
  }
}
\end{wardcasebox}

\begin{wardcasebox}{wardcase tool}{TOOL RESPONSE: CALL 10}
{
  "status": "valid_exploration_no_recorded_outcome",
  "state_advanced": false,
  "newly_released_evidence_ids": [],
  "semantic_repeat_without_new_evidence": true
}
\end{wardcasebox}

{\color{wardCaseRed}
Calls 9 and 10 submit the same query. Neither releases evidence or advances
the case.
}\par

% =========================================================
% Call 26
% =========================================================

\medskip
\textbf{Call 26: surgery completion is assumed}\par

\begin{wardcasebox}{wardcase user}{USER PROMPT: CALL 26}
{
  "progress_guidance": {
    "last_outcome_status": "valid_exploration_no_recorded_outcome"
  },
  "observation": {
    "stage": "S2",
    "new_evidence_ids": []
  }
}
\end{wardcasebox}

The visible records still contain the biopsy and its diagnosis. No completed
definitive surgery appears in the trajectory.\par

\begin{wardcasebox}{wardcase model,coltext=wardCaseRed}{MODEL TOOL CALL: CALL 26}
{
  "tool": "ehr.read",
  "args": {
    "query": "Review the operative record after the completed definitive right breast cancer surgery, including the full breast and regional nodal procedures performed."
  }
}
\end{wardcasebox}

\begin{wardcasebox}{wardcase tool}{TOOL RESPONSE: CALL 26}
{
  "status": "valid_exploration_no_recorded_outcome",
  "state_advanced": false,
  "newly_released_evidence_ids": []
}
\end{wardcasebox}

% =========================================================
% Trajectory outcome
% =========================================================

\medskip
\textbf{Trajectory outcome}\par
\smallskip

After call 8, the remaining 31 executed actions release no new evidence and
complete no further reference transition. The trajectory ends in S2 after
40 executed actions, with 6 of 12 reference transitions completed.\par

\smallskip
{\color{wardCaseRed}
Termination: \texttt{action\_budget\_exhausted}.
}\par

\endgroup

\newcommand{\wardturnentry}[7]{%
  \par\begingroup
  \setlength{\parindent}{0pt}%
  \setlength{\leftskip}{1em}%
  \setlength{\rightskip}{0pt plus 2em}%
  \noindent
  \textcolor{wardcasetime}{\bfseries\{"call": #1,}
  \textcolor{wardcaseevidence}{\bfseries "tool": "#2",}
  "target": "#3",\allowbreak\par\nobreak
  \hspace*{1em}\textcolor{wardcaseobserved}{\bfseries "outcome": "#4",}
  \textcolor{wardcasenote}{"stage": "#5", "completed": "#6", "released": "#7"\},}\par
  \endgroup\vspace{1.2pt}%
}

The model initially acted on public feedback: two blocked biopsy attempts led
to prerequisite recovery and six completed reference transitions. Progress
stopped after call 8. All 31 subsequent executed actions left the visible
evidence, stage, and completed-transition count unchanged. The model repeatedly
failed to track and close the remaining dependencies and showed a state
inconsistency at call 26.
\endgroup

\end{document}